\documentclass{article}

\usepackage{microtype}
\usepackage{graphicx}
\usepackage{subfigure}
\usepackage{booktabs} 

\usepackage{hyperref}

\usepackage[accepted]{icml2024}

\makeatletter
\def\newmaketag{%
  \def\maketag@@@##1{\hbox{\m@th\normalfont\normalsize##1}}%
  }
\makeatother

\usepackage[T1]{fontenc}    
\usepackage{hyperref}       
\usepackage{url}            
\usepackage{booktabs}       
\usepackage{amsfonts}       
\usepackage{amssymb,amsmath,amsthm}
\usepackage{nicefrac}       
\usepackage{bm}
\usepackage{braket}
\usepackage{microtype}      
\usepackage{xcolor}
\usepackage{graphicx}
\usepackage{thmtools} 
\usepackage{dsfont} 
\usepackage{thm-restate}
\usepackage{cancel}
\usepackage{enumerate}   

\usepackage{multicol}
\usepackage{multirow}
\hypersetup{
    colorlinks,
    linkcolor={red!50!black},
    citecolor={green!50!black},
    urlcolor={blue!80!black}
}

\newtheorem{theorem}{Theorem}
\newtheorem{proposition}{Proposition}

\begin{document}

\twocolumn[
    \icmltitle{Variational Linearized Laplace Approximation for Bayesian Deep Learning}



\icmlsetsymbol{equal}{*}

\begin{icmlauthorlist}
\icmlauthor{Luis A. Ortega}{uam}
\icmlauthor{Simón Rodríguez Santana}{comillas}
\icmlauthor{Daniel Hernández-Lobato}{uam}
\end{icmlauthorlist}

\icmlaffiliation{uam}{Universidad Autónoma de Madrid}
\icmlaffiliation{comillas}{Institute for Research in Technology (IIT), ICAI Engineering School, Universidad Pontificia Comillas}

\icmlcorrespondingauthor{Luis A. Ortega}{luis.ortega@uam.es}

\icmlkeywords{Variational Inference, Linearized Laplace Approximation, Bayesian Deep Learning, Pretrained Uncertainty}

\vskip 0.3in
]
\printAffiliationsAndNotice{} 
\begin{abstract}
The Linearized Laplace Approximation (LLA) has been recently used to perform uncertainty estimation on the predictions of pre-trained deep neural networks (DNNs). However, its widespread application is hindered by significant computational costs, particularly in scenarios with a large number of training points or DNN parameters. Consequently, additional approximations of LLA, such as Kronecker-factored or diagonal approximate GGN matrices, are utilized, potentially compromising the model's performance. To address these challenges, we propose a new method for approximating LLA using a variational sparse Gaussian Process (GP). Our method is based on the dual RKHS formulation of GPs and retains, as the predictive mean, the output of the original DNN. Furthermore, it  allows for efficient stochastic optimization, which results in sub-linear training time in the size of the training dataset. Specifically, its training cost is independent of the number of training points. We compare our proposed method against accelerated LLA (ELLA), which relies on the Nystr\"om  approximation, as well as other LLA variants employing the sample-then-optimize principle. Experimental results, both on regression and classification datasets, show that our method outperforms these already existing efficient variants of LLA, both in terms of the quality of the predictive distribution and in terms of total computational time. 
\end{abstract}

\section{Introduction}

Deep neural networks (DNNs) have gained widespread popularity for addressing pattern recognition problems due to their state-of-the-art performance in predicting target values from a set of input attributes \citep{he2016deep,Vaswani2017}. Despite this great success, DNNs exhibit limitations when computing a predictive distribution that accounts for the confidence in the predictions. Specifically, DNNs result in weak calibration \citep{guo2017calibration} and in poor reasoning regarding model uncertainty \citep{blundell2015weight}. These issues become particularly critical in risk-sensitive situations like autonomous driving \citep{kendall2017uncertainties} and healthcare systems \citep{leibig2017leveraging} among others.

A Bayesian approach, where probabilities describe degrees of belief in the potential values of the DNN parameters, has proven useful for treating these pathologies \citep{mackay1992practical, neal2012bayesian, graves2011practical}. Bayes' rule is used here to get a posterior distribution in the high-dimensional space of DNN parameters. Nevertheless, due to the intractability of the calculations, the exact posterior is often approximated with a simpler distribution. Different techniques can be used for this, including (but not limited to) variational inference (VI) \citep{blundell2015weight}, Markov chain Monte Carlo (MCMC) \citep{chen2014stochastic} and the Laplace approximation (LA) \citep{mackay1992bayesian, ritter2018scalable}. 

\begin{figure*}[tb!]
	\begin{center}
	\begin{tabular}{ccc}
    {\scriptsize LLA }& {\scriptsize VaLLA }& {\scriptsize ELLA} \\
	\includegraphics[width=0.30\textwidth]{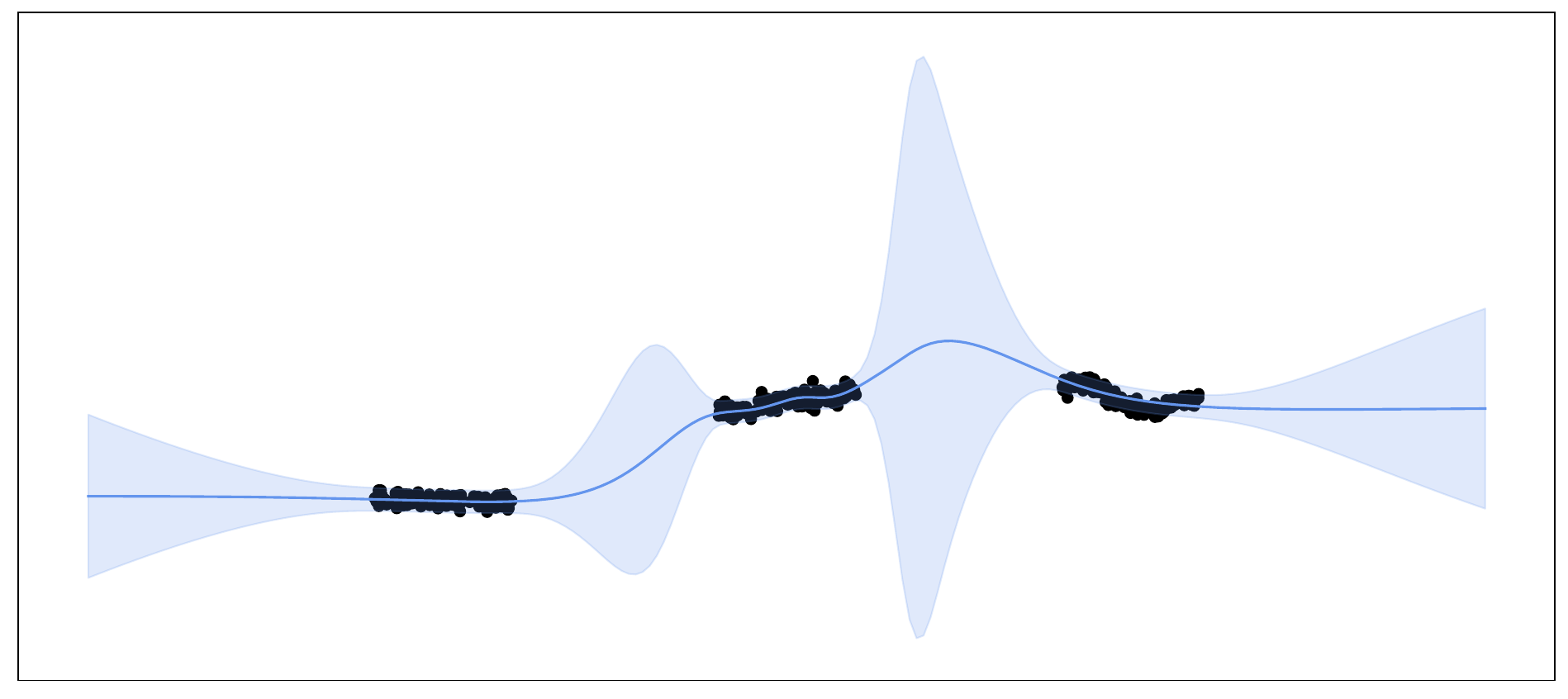} & \includegraphics[width=0.30\textwidth]{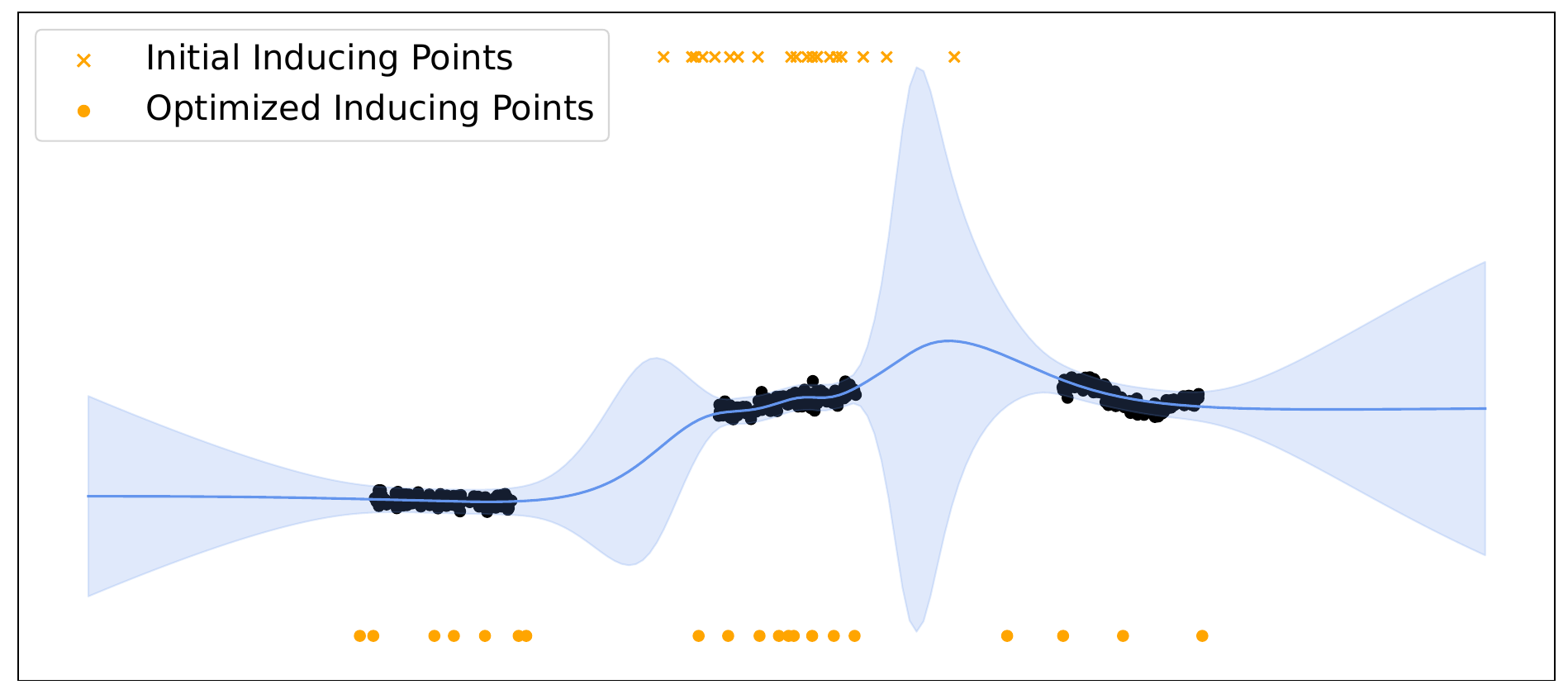} &\includegraphics[width=0.30\textwidth]{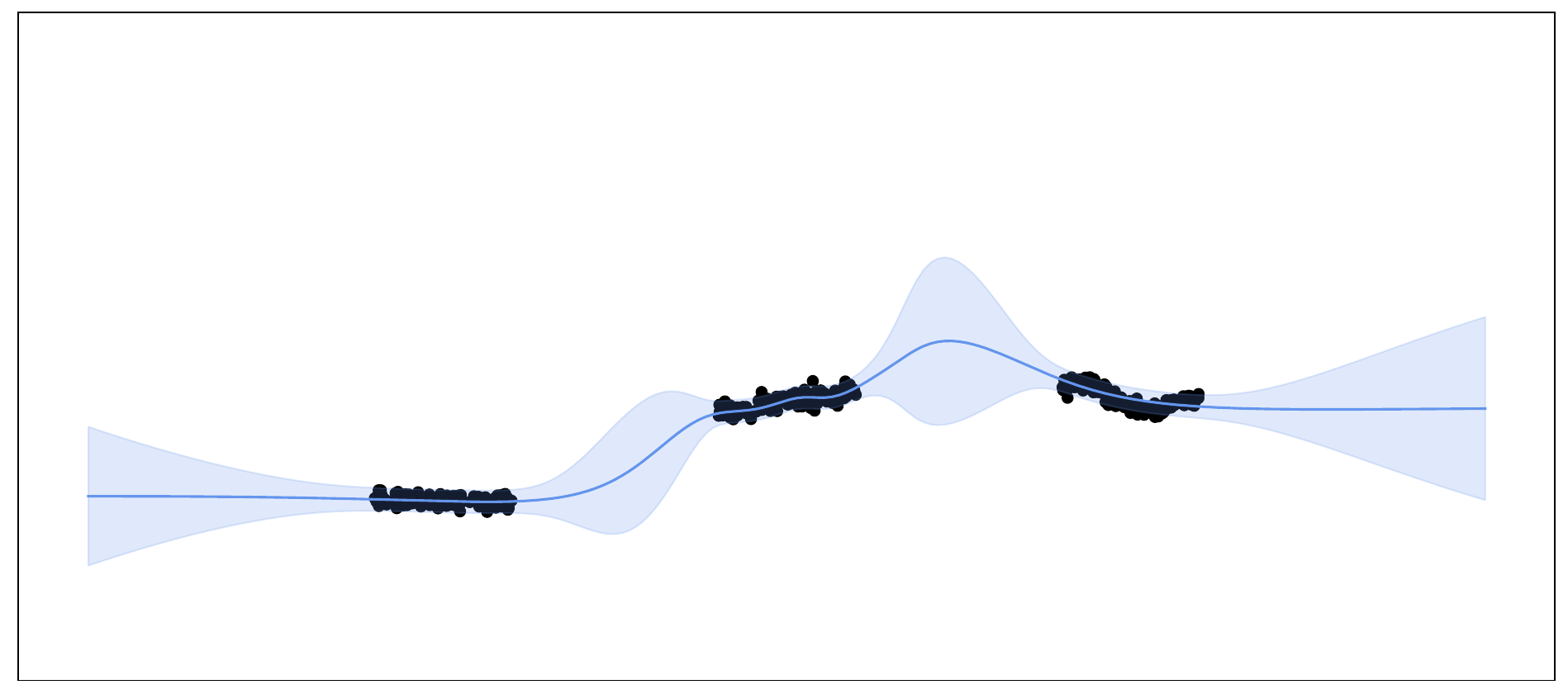} \\
    {\scriptsize Last-Layer LLA} & {\scriptsize MoE LLA} & {\scriptsize Kronecker LLA}\\
	\includegraphics[width=0.30\textwidth]{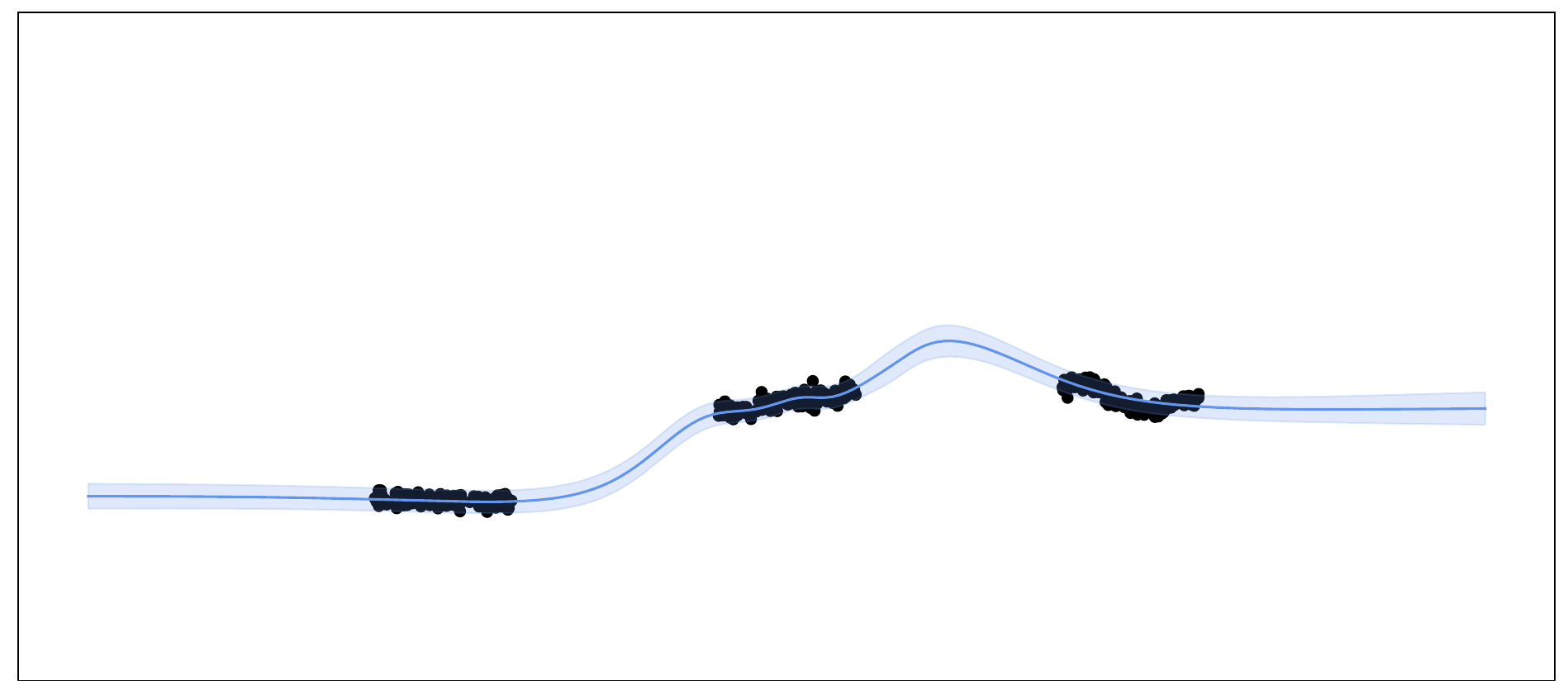} & \includegraphics[width=0.28\textwidth]{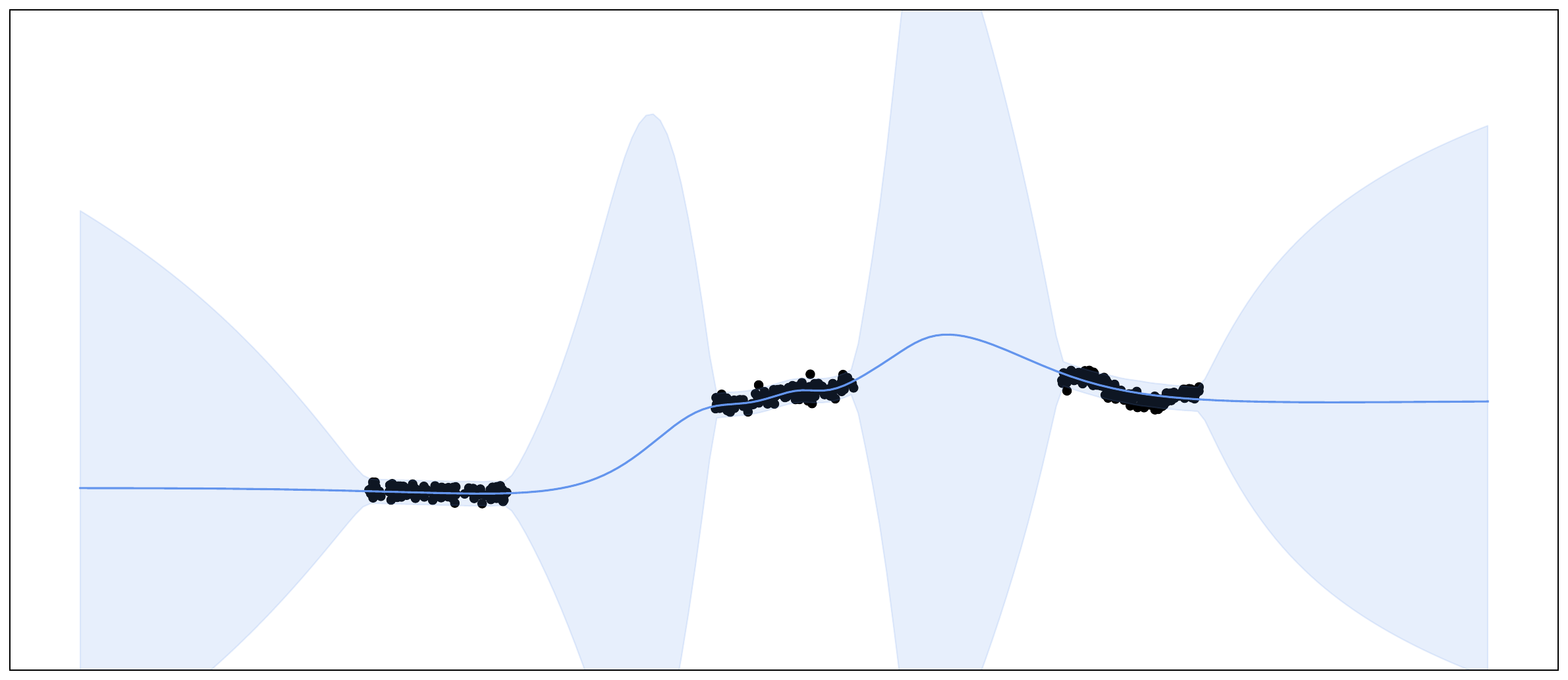} & \includegraphics[width=0.30\textwidth]{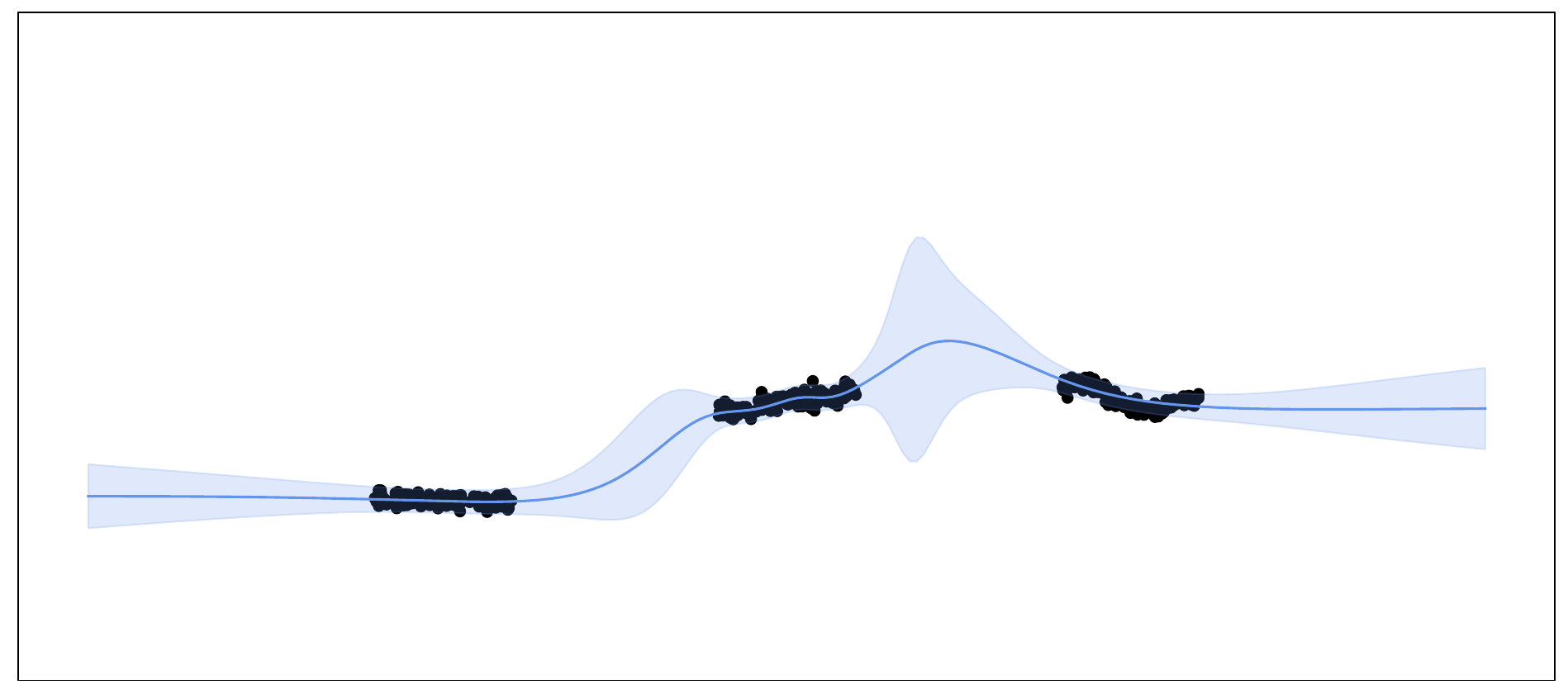} \\
	\end{tabular}
	\end{center}
    \caption{Predictive distribution (mean in blue and shaded two times the standard deviation) on a toy 1D regression dataset with 
	a 2 hidden layer MLP with \(50\) units trained using back-propagation. The predictive distribution of VaLLA is on par with or 
	better than other approximations (last layer and Kronecker factorization), MoE LLA with \(200\) clusters and other methods (ELLA). The optimal 
	values for the noise and prior are optimized by maximizing the marginal log likelihood estimate of LLA. VaLLA and ELLA use 
	the optimal values found by LLA. VaLLA uses \(20\) inducing points for the predictive variances. ELLA uses \(20\) random locations and \(20\) features.}
    \label{fig:intro}
\end{figure*}

LA offers several advantages over alternative methods. It leverages the maximum a posteriori (MAP) solution, attainable via back-propagation, along with the inverse of the Hessian of the DNN parameters there. The negative inverse  Hessian provides the covariances of a Gaussian posterior approximation and the MAP estimate furnishes the mean. Consequently, LA yields a Gaussian posterior approximation following standard DNN training with back-propagation. This resembles a pre-training step followed by fine-tuning, a common practice in deep learning \citep{daxberger2021laplace}. 

The main drawback of LA is the necessity of computing the Hessian at the MAP estimate, which becomes prohibitive for large DNNs. To simplify this, the Hessian is often approximated using the generalized Gauss-Newton (GGN) matrix \citep{martens2015optimizing}. While this is known to produce underfitting in LA \citep{lawrence2001variational}, using a linearized DNN for prediction alleviates it \citep{foong2019between}. This approach aligns with the fact that the GGN Hessian estimate is exact when applying LA to a linearized DNN. This method is referred to as Linearized LA (LLA) \citep{immer2021improving} and it simply consist in applying LA on a first-order Taylor approximation of the DNN output. 

As a consequence of the Taylor approximation, in regression problems, the mean of LLA's predictive distribution coincides with the DNN's prediction at the MAP estimate. This means that LLA simply provides error bars on the pre-trained DNN's predictions. Additionally, the Hessian estimate is always positive definite in LLA. This enables the usage of LLA at any point, not limited to the MAP estimate. This enhances the post-hoc nature of LLA w.r.t. that of LA.

LLA has shown competitiveness with other approaches on a variety of uncertainty quantification tasks \citep{immer2021improving, foong2019between}. Nevertheless, for the practical usage of LLA in the context of large DNNs, further approximations on top of GGN are required. For instance, diagonal or Kronecker-factored (KFAC) approximations of the complete GGN matrix are used. This is motivated by the cubic cost of computing posterior variances in LLA w.r.t. the training set size or the number of parameters in the DNN.

Here, we broaden LLA's usage for DNNs to cases with a substantial number of parameters and training instances. We reinterpret the LLA predictive distribution as that of a Gaussian Process (GP) \citep{khan2019approximate}, and use a sparse variational GP based on inducing points to approximate it \citep{titsias2009variational}. Standard sparse GPs often alter the predictive mean and hence may deteriorate the accuracy of the pre-trained DNN, which is assumed to have minimal generalization error. To avoid this, we use a dual representation of the GP in the Reproducing Kernel Hilbert Space (RKHS). This enables the sparse approximation to be used only in the computation of the predictive variances of the GP. The result is a sparse GP approximation where the predictive mean coincides with the predictions of the pre-trained DNN, without introducing additional prediction error.

We call our method Variational LLA (VaLLA) and conduct a series of regression and classification experiments to compare its performance and computational cost with that of related methods from the literature. Our comparisons include (i) a method that uses Kronecker and diagonal approximations of the GGN matrix, (ii) a method that uses a Nystr\"om approximation of that matrix (ELLA) \citep{deng2022accelerated}, and (iii) a method that relies on generating samples from LLA's posterior distribution using the sample-then-optimize principle \citep{antoran2023sampling}. VaLLA's performance is comparable to or better than that of these other methods and is obtained at a smaller cost. Specifically, VaLLA adopts mini-batch training, resulting in sub-linear cost w.r.t. the training set size. Furthermore, VaLLA often generates predictive distributions that closely resemble those of LLA. Figure~\ref{fig:intro} shows this in a simple 1-D toy regression problem. A Mixture of Experts approach (MoE LLA) based on dividing the input space into different regions \citep{lee2022trust} using \(200\) clusters is also included in the figure for comparison. This method is further discussed in Section \ref{sec:related_work}.

\section{Background}

Consider the task of inferring an unknown function \(f:\mathbb{R}^D \to \mathbb{R}\) based on noisy observations \(\mathbf y = ( y_1, \dots,  y_N)^\text{T}\) at corresponding locations \(\mathbf X = (\mathbf x_1, \dots, \mathbf x_N)\). Deep learning (DL) defines a Neural Network \(g:\mathbb{R}^D \times \mathbb{R}^P \to \mathbb{R}\) with $P$ parameters so that \(\exists \bm{\theta}^\star\in \mathbb{R}^P\) \emph{s.t.} \(f(\cdot) \approx g(\cdot, \bm{\theta}^\star)\). Thus, $f$ is fully specified by $\bm{\theta}^\star$. DL often estimates \(\bm{\theta}^{\star}\) via back-propagation. Nonetheless, despite the remarkable performance achieved \citep{Vaswani2017}, DL methods lack proper estimation of output uncertainty, which can cause overconfident predictions in regions without training data.

In the context of Bayesian inference, the observations are related to \(\mathbf g = (g(\mathbf {x}_1, \bm{\theta}),
\ldots,g(\mathbf{x}_N,\bm{\theta}))^\text{T}\) via the 
likelihood, \(p( \mathbf y |\bm \theta)\). In regression problems, $y_i \in \mathbb{R}$ and 
the likelihood is typically Gaussian. By contrast, in classification problems, $y_i \in \{1,\ldots,C\}$, with $C$ 
the number of classes and the likelihood is categorical with probabilities given by a softmax link function.
In this case, \(g:\mathbb{R}^D \times \mathbb{R}^P\to \mathbb{R}^C\) is a multi-output function with $C$ outputs, one per class label.

Bayesian neural networks (BNNs) use a probabilistic framework \citep{mackay1992bayesian}, 
establishing a prior over the network parameters \(p(\bm{\theta})\) and computing the 
Bayesian posterior, \(p(\bm{\theta}|\mathbf{y}) \propto p(\mathbf{y}|\bm{\theta})p(\bm{\theta})\), 
for predictions\footnote{We ignore the dependence on $\mathbf{X}$ to simplify the notation.}. 
The analytic computation of the posterior is often intractable due to the strong
non-linearities of the NN. Most methods resort to an approximate posterior 
\(q(\bm{\theta}) \approx p(\bm{\theta}|\mathbf{y})\), later used for
prediction \(p(y^\star|\mathbf{x}^\star, \mathbf{y}) = \mathbb{E}_{p(\bm{\theta}|\mathbf{y})}
\left[p(y^\star|\mathbf{x}^\star, \bm{\theta}) \right] \approx \mathbb{E}_{q(\bm{\theta})}\left[p(y^\star|\mathbf{x}^\star, \bm{\theta}) \right]
\approx S^{-1}\sum_{s=1}^S p(y^\star|\mathbf{x}^\star, \bm{\theta}_s)\) with $\bm{\theta}_s \sim q(\cdot)$ and $S$ Monte Carlo samples.
This approximate distribution captures prediction uncertainty \citep{bishop2006}.

The Laplace approximation (LA) builds a Gaussian approximate posterior \citep{mackay1992bayesian}. 
This takes the form of \(q(\bm{\theta}) = \mathcal{N}(\bm{\theta}|\hat{\bm{\theta}}, \bm{\Sigma})\), 
where \(\hat{\bm{\theta}}\) denotes the MAP solution, \emph{i.e.}, \(\hat{\bm{\theta}} = \arg \max_{\bm{\theta}} 
\log p(\mathbf{y}|\bm{\theta}) + \log p(\bm{\theta})\), and \(\bm{\Sigma}\) is the inverse of 
the negative Hessian of the log posterior, \emph{i.e.}, 
\begin{equation}
	\bm{\Sigma}^{-1} = -\nabla^2_{\bm{\theta} \bm{\theta}} \left[\left.\log p(\mathbf{y}|\bm{\theta}) + \log p(\bm{\theta})\right]\right|_{\bm{\theta} = \hat{\bm{\theta}}}\,.
\end{equation}
Often, an isotropic Gaussian prior \(p(\bm{\theta}) = \mathcal{N}(\bm{\theta}| \bm{0}, \sigma^2_0 \bm{I}_P)\) is considered
so that \(-\nabla^2_{\bm{\theta} \bm{\theta}} \left. \log p(\bm{\theta})\right|_{\bm{\theta} = \hat{\bm{\theta}}} = \bm{I}_P/\sigma_0^2\). Hence
\begin{equation}
    \bm{\Sigma}^{-1} = -\nabla^2_{\bm{\theta} \bm{\theta}} \left.\log p(\mathbf{y}|\bm{\theta})\right|_{\bm{\theta} = \hat{\bm{\theta}}} + \tfrac{1}{\sigma_0^2}\bm{I}_P\,.
\end{equation}
Given the intractability of the Hessian in large DNNs and its non-guaranteed positive-definiteness, it is common to approximate 
it with the generalized Gauss-Newton (GGN) matrix \citep{immer2021improving}:
\begin{equation}
	\bm{\Sigma}^{-1} \approx \textstyle \sum_{n=1}^{N} J_{\hat{\bm{\theta}}}(\mathbf{x}_n)^\text{T} \Lambda(\mathbf{x}_n, y_n) 
	J_{\hat{\bm{\theta}}}(\mathbf{x}_n)+ \frac{1}{\sigma_0^2}\bm{I}_P\,,
	\label{eq:ggn}
\end{equation}
where \(J_{\hat{\bm{\theta}}}(\mathbf{x}_n) = \nabla_{\bm{\theta}} 
\left. g(\mathbf{x}_n, \bm{\theta})\right|_{\bm{\theta} = \hat{\bm{\theta}}}\) and 
\(\Lambda(\mathbf{x}_n, y_n) = -\nabla^2_{\mathbf g \mathbf g} \left. \log p(y_n|\mathbf g)
\right|_{\mathbf g = g_{\hat{\bm{\theta}}}(\mathbf x_n, \bm{\theta})}\). 
The GGN matrix is guaranteed to be positive semi-definite, which means that $\hat{\bm{\theta}}$ need not be 
a maximum of the log posterior. It can be, \emph{e.g.}, any solution found by early-stopping back-propagation.

The earlier formulation of LA suffers from underfitting 
\citep{lawrence2001variational}. This is attributed to the fact that the GGN approximation is the true Hessian matrix of the 
linearized DNN \(g_{\hat{\bm{\theta}}}^\text{lin}(\mathbf{x}, \bm{\theta}) := g(\mathbf{x}, \hat{\bm{\theta}}) + 
J_{\hat{\bm{\theta}}}(\mathbf{x}_n) (\bm{\theta} - \hat{\bm{\theta}})\) \citep{immer2021improving}. This implies a shift between posterior inference and predictions that can be mitigated by also predicting using the linearized model:
\begin{equation}
	p_{\text{LLA}}(y^\star|\mathbf{x}^\star, \mathbf{y}) = 
	\mathbb{E}_{q(\bm{\theta})}\big[p(y^\star|g_{\hat{\bm{\theta}}}^\text{lin}(\mathbf{x}^\star, \bm{\theta})) \big]\,.
\end{equation}
This method is known as the linearized LA (LLA). However, despite these approximations, LLA still 
requires the inversion of \(\bm{\Sigma}^{-1}\) which scales cubically with the number of 
parameters of the DNN. A dual formulation of LLA as a Gaussian Process (GP), described in 
the next section, scales with cubic cost w.r.t the number of training points.

\subsection{Gaussian Process (GP) Interpretation of LLA}

A linear model with a Gaussian distribution on the model parameters generates a GP 
with specific mean and covariance functions \citep{williams2006gaussian}. 
Consider the prior \(p(\bm{\theta}) = \mathcal{N}(\bm{\theta}| \bm{0}, \sigma_0^2 \mathbf{I}_P)\). The linearized BNN generates random functions that follow a GP with mean and covariance functions defined as
\begin{equation}
	m(\mathbf x) = g_{\hat{\bm{\theta}}}^\text{lin}(\mathbf{x},\bm{0}), \quad K(\mathbf{x}, \mathbf{x}') = \sigma_0^2 J_{\hat{\bm{\theta}}}(\mathbf{x})^\text{T} J_{\hat{\bm{\theta}}}(\mathbf{x}' )\,.
\end{equation}
Using the approximate posterior from LLA, \emph{i.e.}, \(q(\bm{\theta}) = \mathcal{N}(\bm{\theta}|\hat{\bm{\theta}}, \bm{\Sigma})\), we
also obtain a GP for prediction. The mean and covariance functions, providing the predictive mean and variances of the LLA approximation, are in this case:
\begin{equation}
\begin{aligned}
	m(\mathbf x) &=  g_{\hat{\bm{\theta}}}^\text{lin}(\mathbf{x},\hat{\bm{\theta}}) = g(\mathbf{x},\hat{\bm{\theta}})\,,\\
    K(\mathbf{x}, \mathbf{x}') &= J_{\hat{\bm{\theta}}}(\mathbf{x})^\text{T} \bm{\Sigma} J_{\hat{\bm{\theta}}}(\mathbf{x}' )\,.
	\label{eq:posterior_gp}
\end{aligned}
\end{equation}
The Woodbury formula on $\bm{\Sigma}$, as defined in (\ref{eq:ggn}), gives
\begin{equation}
    \bm{\Sigma} =\sigma_0^2 \big( \bm{I}_P - \mathbf{J}^T \big( \tfrac{1}{\sigma_0^2} \bm{\Lambda}_{\mathbf{X}, \mathbf{y}}^{-1} + \mathbf{J}\mathbf{J}^T\big)^{-1} \mathbf{J} \big)\,.
\end{equation}
Defining \(\kappa(\mathbf{x}, \mathbf{x}') = \sigma_0^2 J_{\hat{\bm{\theta}}}(\mathbf{x})^\text{T} J_{\hat{\bm{\theta}}}(\mathbf{x}' ) \) 
as the (scaled) Neural Tangent Kernel (\emph{i.e.}, prior covariance function) of 
the GP \citep{immer2021improving}, and \(\mathbf{Q} = \bm{\Lambda}_{\mathbf{X}, \mathbf{y}}^{-1}+  \kappa(\mathbf{X}, \mathbf{X})\),
the covariance function of the GP in (\ref{eq:posterior_gp}) takes the expression
\begin{equation}\label{eq:kernel}
	K(\mathbf{x}, \mathbf{x}') = \kappa(\mathbf{x}, \mathbf{x}') - \kappa(\mathbf{x}, \mathbf{X})\mathbf{Q}^{-1} \kappa(\mathbf{X}, \mathbf{x}' )\,.
\end{equation}
This allows us to interpret the LLA approximate predictive distribution as a posterior GP in function space, with prior covariance function given by $\kappa(\cdot,\cdot)$. The bottleneck of this interpretation is the evaluation of \(\mathbf{Q}^{-1}\), with  $\mathcal{O}(N^3 + N^2 P)$ cost. In the case of classification problems with $C$ classes, $\kappa(\mathbf{X},\mathbf{X})$ and $\mathbf{Q}$ have size $NC\times NC$, thus the cost is 
also cubic w.r.t. the number of classes $C$ or DNN outputs.

\subsection{Dual formulation of Gaussian Processes in RKHS}

A Reproducing Kernel Hilbert Space (RKHS) \(\mathcal{H}\) is a Hilbert space of 
functions with the reproducing property: \(\forall \mathbf{x} \in \mathcal{X} \ \exists \phi_{\mathbf{x}} \in \mathcal{H}\) such 
that \(\forall f \in \mathcal{H}, f(\mathbf{x}) = \braket{\phi_{\mathbf{x}}, f} \). 
In general, \(\mathcal{H}\) can be infinite-dimensional and uniformly approximate continuous functions on a compact set.

A zero-mean prior GP with posterior mean and covariance functions $m(\cdot)$ and $K(\cdot,\cdot)$, respectively, has a dual representation in a RKHS \citep{cheng2016incremental}. There exists \(\mu \in \mathcal{H}\) and a linear 
semi-definite positive operator \(\Sigma : \mathcal{H} \to \mathcal{H}\) such that \( \forall \mathbf{x}, \mathbf{x}' \in \mathcal{X}\), \(\exists \phi_{\mathbf{x}}, \phi_{\mathbf{x}' }\) verifying
\begin{equation}
        m(\mathbf{x}) = \braket{\phi_{\mathbf{x}},\mu} \quad \text{and} \quad  K(\mathbf{x}, \mathbf{x}' ) = \braket{\phi_{\mathbf{x}},\Sigma (\phi_{\mathbf{x}'})}\,.
\end{equation}
As a shorthand and an abuse of notation, we write that \(p(f) = \mathcal{N}(f|\mu, \Sigma)\), where this 
refers to a Gaussian measure in an infinite dimensional space, not a Gaussian density.  

\section{Variational LLA (VaLLA)}

Here, we present our proposed method, \textit{Variational LLA} (VaLLA).
VaLLA leverages the use of variational sparse GPs. 
Thus, we proceed to describe these methods first.

\subsection{Variational Sparse GPs}

Variational sparse GPs approximate the GP posterior 
using a GP parameterized by $M$ inducing points $\mathbf{Z}$,
each in $\mathbb{R}^D$, and associated process 
values \(\mathbf{u} = f(\mathbf{Z})\) \citep{titsias2009variational},
\begin{equation}
    p(\mathbf{f}, \mathbf{u} | \mathbf{y}) \approx q(\mathbf{f}, \mathbf{u}) = p(\mathbf{f} | \mathbf{u})q(\mathbf{u})\,,
\end{equation}
where  \(q(\mathbf{u}) = \mathcal{N}(\mathbf{u}|\hat{\bm{m}}, \hat{\bm{S}})\), $\mathbf{f}=f(\mathbf{X})$ and $p(\mathbf{f} | \mathbf{u})$ is fixed. The approximate distribution $q(\mathbf{u})$ is obtained by
minimizing the KL-divergence \(\text{KL}\left(q(\mathbf{f}, \mathbf{u}) \mid  p(\mathbf{f}, \mathbf{u}| \mathbf{y}) \right) \). In practice, the minimization problem is transformed into the maximization of the lower bound of the log-marginal likelihood
\begin{equation}\label{eq:elbo}
    \mbox{\small$\displaystyle \log p(\mathbf{y}) \geq \max_{\mathbf{Z}, \hat{\bm{m}}, \hat{\bm{S}}} \int_{\mathbf{f}, \mathbf{u}} q(\mathbf{f}, \mathbf{u}) \log \frac{p(\mathbf{y}|\mathbf{f}) p(\mathbf{f}| \mathbf{u})p(\mathbf{u})}{q(\mathbf{f}, \mathbf{u})}\,$}\,,
\end{equation}
which has cost $\mathcal{O}(NM^2+M^3)$ due to 
the cancellation of the factor $p(\mathbf{f}|\mathbf{u})$ 
since $q(\mathbf{f}, \mathbf{u}) = p(\mathbf{f} | \mathbf{u})q(\mathbf{u})$.

\begin{theorem}[{\citet{cheng2016incremental}}]\label{thm:sparse_rkhs}
    Using a sparse GP approximation with \(q(\mathbf{f}, \mathbf{u}) = p(\mathbf{f}|\mathbf{u})q(\mathbf{u})\) is 
	equivalent to restricting the mean and covariance functions of the dual representation in the RKHS to
    \begin{equation}
        \tilde{\mu} = \Phi_{\mathbf{Z}}(\bm{a}) \quad \text{and} \quad \tilde{\Sigma} = I + \Phi_{\mathbf{Z}}\bm{A}\Phi_{\mathbf{Z}}^T\,,
    \end{equation}
    where the functional \(\Phi_{\mathbf{Z}}: \mathbb{R}^M \to \mathcal{H} \) defines a linear combination of basis functions as \(\Phi_{\mathbf{Z}}(\bm{a}) = \sum_{m=1}^{M} a_m \phi_{\mathbf{z}_m}\), with \(\bm{a} = (a_1,\dots,a_M) \in \mathbb{R}^M\) and the functional \(\Phi_{\mathbf{Z}}\bm{A}\Phi_{\mathbf{Z}}^T= \sum_{i=1}^{M}\sum_{j=1}^M\phi_{\mathbf{z}_i} A_{i,j} \phi_{\mathbf{z}_j}^T\), defines a quadratic expression where \(\bm{A} \in \mathbb{R}^{M\times M}\) such that \(\tilde{\Sigma} \geq 0\).
\end{theorem}


Theorem~\ref{thm:sparse_rkhs} indicates that the algorithm of
\citet{titsias2009variational} optimizes a variational Gaussian measure where 
\(\tilde{\mu}\) and \(\tilde{\Sigma}\) are parameterized by a function basis 
\(\{\phi_{\mathbf{z}} \in \mathcal{H}\ | \ \mathbf{z} \in \mathbf{Z}\}\). 
\citet{cheng2017variational} propose to generalize this so that each of the 
linear operators is optimized using different bases (sets of inducing points). Let \(\mathbf{Z}_{\alpha}\) and \(\mathbf{Z}_{\beta}\) be 
two sets of inducing points for the mean and the variance, respectively. 
The parameterization of \citet{cheng2017variational} is:
\begin{equation}
	\tilde{\mu} = \Phi_{\mathbf{Z}_\alpha}(\bm{a}) \quad \text{and} \quad \tilde{\Sigma} = (I + \Phi_{\mathbf{Z}_\beta}\bm{A}\Phi_{\mathbf{Z}_\beta}^T)^{-1}\,,
\end{equation}
where \(\Phi_{\mathbf{Z}_\alpha}:\mathbb{R}^{M_{\alpha}} \to \mathcal{H}\) and \(\Phi_{\mathbf{Z}_\beta}:\mathbb{R}^{M_{\beta}} \to \mathcal{H}\) 
are defined as \(\Phi_{\mathbf{Z}}\) using \(\mathbf{Z}_\alpha\) 
and \(\mathbf{Z}_\beta\), respectively. Now, there are two sets of inducing points, $M_\alpha$ for the mean and $M_\beta$ for the covariances, respectively. This parameterization is a generalization and cannot be 
obtained using the approach of \citet{titsias2009variational}. Here, $q$ must
be found by optimizing Gaussian measures \citep{cheng2016incremental}:
\begin{equation} \label{eq:opt_dual}
\begin{aligned}
    \max_{q(f)} \mathcal{L} (q(f)) &= \max_{q(f)} \int q(f) \log \frac{p(\mathbf{y}|f)p(f)}{q(f)} df  \\
    &= \max_{q(f)}\mathbb{E}_q \left[ \log p(\mathbf{y}|f)\right] - \text{KL}\left(q \mid p\right)\,,
\end{aligned}
\end{equation}
where the KL term is:
\begin{equation}\label{eq:opt_kl}
\begin{aligned}	
        \text{KL}\left(q \mid p\right) &= \frac{1}{2} \bm{a}^T \bm{K}_{\alpha} \bm{a} + \frac{1}{2} \log |\bm{I} + \bm{K}_{\beta} \bm{A}| \\
        &- \frac{1}{2} \text{tr}\left( \bm{K}_\beta(\bm{A}^{-1} + \bm{K}_\beta)^{-1} \right)\,,
\end{aligned}
\end{equation}
with \(\bm{K}_{\alpha}\) and \(\bm{K}_\beta\) matrices with the prior covariances
among $f(\mathbf{Z}_\alpha)$ and $f(\mathbf{Z}_\beta)$, respectively.

\subsection{Using Decoupled SGP and LLA}

We utilize the decoupled reparameterization of sparse GPs to establish a model 
where the mean of the approximated posterior distribution is anchored to a pre-trained MAP 
solution. We denote this method as \textit{variational LLA} (VaLLA).

\begin{proposition}\label{prop:decoupled}
	If \(g(\cdot, \hat{\bm{\theta}}) \in \mathcal{H}\), then \(\forall \epsilon > 0\) exists a set of \(M_\alpha\) inducing points 
	\(\mathbf{Z}_{\alpha}\) and a collection of scalar values \(\bm{a} \in \mathbb{R}^{M_\alpha}\) such that the dual representation 
	of the sparse Gaussian process defined by
    \begin{equation}
        \tilde{\mu} = \Phi_{\mathbf{Z}_{\alpha}}(\bm{a}) \quad \text{and} \quad \tilde{\Sigma} = (I + \Phi_{\mathbf{Z}_{\beta}}\bm{A}\Phi_{\mathbf{Z}_{\beta}}^T)^{-1}\,,
    \end{equation}
    corresponds to a GP posterior approximation with mean and covariance functions defined as
    \begin{align} \label{eq:pred_valla}
        m^{\star}(\mathbf{x}) &= h_\epsilon(\mathbf{x})\,,  \\
	    K^{\star}(\mathbf{x}, \mathbf{x}') &=K(\mathbf{x}, \mathbf{x}') -  K_{\mathbf{x}, \mathbf{Z}_\beta}(\bm{A}^{-1} + \bm{K}_\beta)^{-1} K_{\mathbf{Z}_\beta, \mathbf{x}'} \,, \nonumber
    \end{align}
	where \(\mathbf{Z}_{\beta}\) is a set of \(M_\beta\) inducing points, 
	\(\bm{A} \in \mathbb{R}^{M_\beta \times M_\beta}\), $K_{\mathbf{x}, \mathbf{Z}_\beta}$ is a vector with the covariances between $f(\mathbf{x})$ and 
	$f(\mathbf{Z}_\beta)$, and \(h_\epsilon\) verifies
	    $d_{\mathcal{H}}(g(\cdot, \hat{\bm{\theta}}), h_\epsilon) \leq \epsilon$,
	    with $d_{\mathcal{H}}(\cdot,\cdot)$ the distance in the RKHS \textup{(see proof in Appendix~\ref{app:proofs}).}
\end{proposition}
Proposition~\ref{prop:decoupled} implies that if \(g(\cdot, \hat{\bm{\theta}}) \in \mathcal{H}\) we can
find values for $\bm{a}$ and inducing points for the mean $\mathbf{Z}_\alpha$ \emph{s.t.}
$d_{\mathcal{H}}(g(\cdot, \hat{\bm{\theta}}), h_\epsilon)$ can be made as small as desired.
For sufficiently small \(\epsilon\), \(h_\epsilon(\cdot) \approx g(\cdot, \hat{\bm{\theta}})\), 
and \(g(\cdot, \hat{\bm{\theta}})\) can be used for prediction instead of $h_\epsilon(\mathbf{x})$. 
Thus, there is no need to optimize \(\bm{a}\) and \(\mathbf{Z}_{\alpha}\) in \eqref{eq:opt_dual}, 
and the posterior distribution of VaLLA uses \(g(\cdot, \hat{\bm{\theta}})\) as its mean function.
The optimal parameters $\mathbf{Z}_{\beta}$ and $\bm{A}$ can be found by optimizing 
\eqref{eq:opt_dual} with \(\bm{a}\) and \(\mathbf{Z}_{\alpha}\) held constant. From the following proposition, computing the optimal value of \(\mathbf{A}\) has cost $\mathcal{O}(NM_\beta^2 + M_\beta^3)$.



\begin{proposition}\label{prop:optimal}
The value of \(\bm{A}\) in Proposition~\ref{prop:decoupled} that minimizes~\eqref{eq:opt_dual} is 
\begin{align}
	\bm{A} &= \frac{1}{\sigma^2} \bm{K}_{\beta}^{-1} \bm{K}_{\bm{Z}_\beta, \bm{X}}\bm{K}_{\bm{X}, \bm{Z}_\beta} \bm{K}_{\beta}^{-1}\,,
\end{align}
where $\sigma^2$ is the noise variance and $\bm{K}_{\bm{X}, \bm{Z}_\beta}$ is a matrix with the prior covariances
between $f(\mathbf{X})$ and $f(\mathbf{Z}_\beta)$.  If $\bm{Z}_\beta=\mathbf{X}$, the covariance function of 
the predictive distribution in (\ref{eq:pred_valla}) is equal to that of the full GP \textup{(see proof in Appendix~\ref{app:proofs}).}
\end{proposition}


\paragraph{MAP solution and Hilbert space. } Proposition~\ref{prop:decoupled} presupposes that \(g(\cdot, \hat{\bm \theta}) \in \mathcal{H}\). In practice, this need not be the case.
Covariance functions such as squared exponential or Matérn are recognized for spanning the entire space of continuous functions. However, whether \(g(\cdot, \hat{\bm \theta}) \in \mathcal{H}\)
holds in general remains unknown. For further discussions on this matter, please refer to Appendix~\ref{app:map}. From here onwards, we assume that if $g(\cdot, \hat{\bm{\theta}}) \notin \mathcal{H}$, then $\mathcal{H}$ is sufficiently expressive to include a close approximation to $g(\cdot, \hat{\bm{\theta}})$. 
Consequently, $g(\cdot, \hat{\bm \theta})$ can be used as the sparse GP posterior mean.

\subsection{Hessian Approximation in VaLLA}

Despite the formulation using GPs function-space duality, VaLLA can be also understood as 
a Hessian approximation method. Note that the predictive covariances of VaLLA are given 
in \eqref{eq:pred_valla}. From this equation, we can make a connection with the exact 
posterior variances given by the full Hessian in \eqref{eq:posterior_gp}. From there, 
one can conclude that VaLLA's inverse negative Hessian approximation is given by: 
\begin{equation}
\sigma^2_0\bm{I}_P - \sigma^2_0\Phi_{\mathbf{Z}_\beta}(\bm{A}^{-1} + \sigma_0^2\Phi_{\mathbf{Z}_\beta}^T\Phi_{\mathbf{Z}_\beta})^{-1}\Phi_{\mathbf{Z}_\beta}^T\sigma_0^2\,,
\end{equation}
which is equal to \((\bm{I}_P/\sigma_0^2 + \Phi_{\mathbf{Z}_\beta}^T \bm{A} \Phi_{\mathbf{Z}_\beta})^{-1}\)
using the Woodbury formula, where \(\bm{A}\) is a free parameter adjusted by VaLLA by minimizing the KL divergence between stochastic processes, as described above. VaLLA's negative Hessian approximation is the inverse of the previous quantity. Namely, 
\begin{equation}
    \bm{I}_P/\sigma_0^2 + \Phi_{\mathbf{Z}_\beta}^T \bm{A} \Phi_{\mathbf{Z}_\beta}\,,
\end{equation}
which is similar in structure to the GGN approximation in~\eqref{eq:ggn} considered by LLA, where the data dependent term has been replaced by the inducing points and the matrix \(\bm{A}\).

\subsection{Hyper-parameter Tuning and \(\alpha\)-divergences}\label{sec:alpha}

VaLLA's predictive mean is anchored to the DNN output, making the maximization of the ELBO in (\ref{eq:opt_dual}) unsuitable for tuning hyper-parameters like the prior variance $\sigma_0^2$. Specifically, in a regression scenario with Gaussian noise with variance \(\sigma^2\) the first term in the r.h.s. of (\ref{eq:opt_dual}) becomes:
\begin{equation}
\sum_{i=1}^N -\frac{\log(2\pi\sigma^2)}{2} - \frac{(y_i - g(\mathbf{x}_i, \hat{\bm{\theta}}))^2}{2\sigma^2} 
	- \frac{K^\star(\mathbf{x}_i, \mathbf{x}_i)}{2\sigma^2}
	\label{eq:first_term}
\end{equation}
where \((y_i - g(\mathbf{x}_i, \hat{\bm{\theta}}))^2\) is constant. Maximizing (\ref{eq:first_term}) w.r.t. $\sigma_0^2$ results in the prior covariances, $\sigma_0^2 J_{\hat{\bm{\theta}}}(\mathbf{x})^\text{T} J_{\hat{\bm{\theta}}}(\mathbf{x}' )$, 
tending to $0$. This makes posterior covariances $K^\star(\mathbf{x}_i, \mathbf{x}_i)$ also tend to $0$, effectively cancelling the last term in (\ref{eq:first_term}). The KL term in (\ref{eq:opt_dual}) is also optimal and \(0\) if $\sigma_0^2 \rightarrow 0$. The reasoning is that, in sparse GPs, tuning hyper-parameters involves a trade-off between fitting the mean to the training data and reducing the predictive variance of the model.  Therefore, in VaLLA's setting, where the predictive mean is fixed, the optimal predictive variance tends to be zero. 

To address these issues, we propose an alternative objective to (\ref{eq:opt_dual}) that facilitates hyper-parameter optimization:
\begin{equation}\label{eq:alpha}
	\max_{q(f)} \quad  \sum_{i=1}^N\frac{1}{\alpha}\log \mathbb{E}_q \left[p(y_i|f)^\alpha\right] - \text{KL}\left(q| p\right)\,.
\end{equation}
Here $\alpha\in(0,1]$ is a parameter. Instead of minimizing $\text{KL}(q(f)|p(f|\mathbf{y}))$, this objective minimizes, in 
an approximate way, the $\alpha$-divergence between $p(f|\mathbf{y})$ and $q(f)$ \citep{liG17}. Remarkably, this can
be achieved by simply changing the data-dependent term in the objective of (\ref{eq:opt_dual}). 

The use of $\alpha$-divergences for approximate inference has been extensively studied \citep{bui2017,Villacampa2020,SantanaH22},
with observations suggesting that values of \(\alpha \approx 0\) result in better predictive mean estimation. Conversely, values of \(\alpha \approx 1\) provide superior predictive 
distributions in terms of the log-likelihood. Thus, in this work we opt for \(\alpha = 1\). In this case, (\ref{eq:alpha}) does not promote $\sigma_0^2 \rightarrow 0$, 
unlike (\ref{eq:opt_dual}), 
as the data-dependent term is the log-likelihood of the training data. An unexpected behavior, however, is that (\ref{eq:alpha}) may lead to overfitting. To alleviate this, we employ an early-stopping strategy using a validation set (see Appendix~\ref{app:early} for further details). Early stopping is also used in other LLA approximations \emph{s.a.} ELLA \citep{deng2022accelerated}.

\paragraph{Mini-batch Optimization.} The objective in \eqref{eq:alpha} supports mini-batch 
optimization with cost $\mathcal{O}(M_\beta^3)$. For $\alpha=1.0$, 
\begin{equation}
	N|\mathcal{B}|^{-1} \sum_{b \in \mathcal{B}} \log \mathbb{E}_q 
	\left[p(y_b|f(\mathbf{x}_b))\right] -\text{KL}\left(q| p\right)\,.
	\label{eq:valla_minibatch_objective}
\end{equation}
Here, $\mathcal{B}$ denotes a mini-batch, and the expectation can be computed in closed 
form in regression problems. In classification, an approximation is available via using the softmax approximation of \citet{daxberger2021}.  This sub-linear cost of VaLLA enables its use in very large datasets.

\paragraph{Prediction.}  Predictions for test points $(y^\star,\mathbf{x}^\star)$ 
are computed using (\ref{eq:pred_valla}) with the DNN output $g(\mathbf{x}^\star,\hat{\bm{\theta}})$ as the mean. $p(y^\star|\mathbf{x}^\star) \approx \mathds{E}_q[p(y^\star|f(\mathbf{x}^\star)]$ is 
evaluated as in training. 

\paragraph{Inducing Points.} The locations of the inducing points $\mathbf{Z}_\beta$
are found by optimizing (\ref{eq:valla_minibatch_objective}) with K-means initialization.

\subsection{Limitations of VaLLA}
\label{sec:limitations}

VaLLA is limited by three factors: (i) Computing the predictive distribution at each training iteration involves inverting \(\mathbf{A}^{-1} + K_{\mathbf{Z}_\beta}\) in \eqref{eq:pred_valla}, with cubic cost in the number of inducing points $M_{\beta}$. Therefore, VaLLA cannot accommodate a very large number of inducing points. (ii) The objective in \eqref{eq:alpha} requires a validation set and early-stopping for effective optimization of the prior variance $\sigma_0^2$, thus further increasing training time. (iii) VaLLA requires additional training compared to other LLA approximations. However, in this regard, early-stopping can also reduce the training time by cutting down the number of iterations. In the Taxi experiments performed in Section~\ref{sec:exp}, early-stopping is triggered when only \(16.6\%\) of the training data has been seen. (iii) Mini-batch optimization in \eqref{eq:valla_minibatch_objective} involves evaluating $K_{\mathbf{x},\mathbf{Z}_\beta}$  $\forall \mathbf{x} \in \mathcal{B}$  and $\bm{K}_{\beta}$. Hence, we require efficient evaluation of the (scaled) Neural Tangent Kernel, \(\kappa(\cdot, \cdot) = \sigma^2 J_{\hat{\bm{\theta}}}(\cdot)^\text{T} J_{\hat{\bm{\theta}}}(\cdot)\) and its gradients to find $\mathbf{Z}_\beta$. While there are libraries that use structure in the derivatives for the efficient computation of $\kappa(\cdot, \cdot)$, these are limited to a few DNN models \citep{novak2022fast}. A simple but inefficient approach to evaluate $\kappa(\cdot, \cdot)$ involves computing and storing all full Jacobians in memory, for each mini-batch instance and inducing point. This is tractable in our problems, but makes VaLLA infeasible for very large problems, \emph{e.g.}, ImageNet. Appendix~\ref{app:ntk} shows a very efficient \emph{layer-by-layer} method to obtain  $K_{\mathbf{x},\mathbf{Z}_\beta}$  $\forall \mathbf{x} \in \mathcal{B}$  and $\bm{K}_{\beta}$. However, this requires computing each layer's contribution to the Jacobian at hand, which is difficult for large and complex DNNs.


\section{Related Work}
\label{sec:related_work}

LA for DNNs was originally introduced by \citet{mackay1992bayesian}, applying it to small networks 
using the full Hessian. \citet{mackay1992evidence} also proposed an approximation similar to the generalized Gauss-Newton (GGN). The combination of scalable factorizations or diagonal Hessian approximations \citep{martens2015optimizing, botev2017practical} with the GGN approximation \citep{martens2020new} played a crucial role in the resurgence of LA for modern DNNs \citep{ritter2018scalable, khan2019approximate}. Recent works aim to relax the Gaussian assumption of LLA adopting a Riemannian-Laplace approximation, where samples naturally fall into weight regions with low negative log-posterior \citep{bergamin2023riemannian}. 

To address the underfitting issue associated with LA \citep{lawrence2001variational}, particularly when combined with the GGN approximation, \citet{ritter2018scalable} proposed a Kronecker 
factored (KFAC) LLA approximation. This approach outperforms LA with a diagonal Hessian matrix.


The GP interpretation of LLA \citep{khan2019approximate} allows using GP approximate methods to 
speed up the computations. \citet{immer2021improving} propose to use a subset of the training dataset as a scalable alternative to the true GP. \citet{lee2022trust} propose a Mixture of Experts approach where each expert is trained on a different soft-margin cluster. However, the proposed clustering algorithm, although more efficient than Kernel-K-means, has linear cost w.r.t. the training set size. VaLLA, on the other hand, has sub-linear training time w.r.t. training set size due to mini-batch training. Moreover, it is not clear how to consider neighboring clusters in high dimensional input spaces. The authors only provide code for a 1-dimensional problem. Third, fitting a local GP using the data of the corresponding cluster and its neighbors is expected to overestimate the predictive variance since the model has been trained with a smaller number of training instances (see Figure~\ref{fig:intro}). This is particularly the case in datasets with millions of training instances such as Taxi. This problem is also described by \citet{immer2021improving}. 

\citet{deng2022accelerated} proposed a Nystr\"om approximation of the true GP 
covariance matrix by using $M\ll N$ points chosen at random from the training set. 
The method, called ELLA, has cost $\mathcal{O}(NM^3)$. ELLA also requires computing the costly 
Jacobian vectors required in VaLLA, but does not need their 
gradients. Unlike VaLLA, the Nystr\"om approximation needs to visit each instance in the 
training set. However, as stated by \citet{deng2022accelerated}, ELLA suffers from over-fitting.  
An early-stopping strategy, using a validation set, is proposed to alleviate it. 
In this case, ELLA only considers a subset of the training data. ELLA does not 
allow for hyper-parameter optimization, unlike VaLLA. The prior variance $\sigma_0^2$ must be 
tuned using grid search and a validation set, which increases training time significantly. 

The recent work of \citet{scannell2024function} proposes a similar approach to VaLLA, where an inducing point 
sparse approach is used to construct a GP from a pre-trained DNN. However, two main points differentiate this 
work from our approach: (i) the pre-trained DNN is not kept as the posterior mean of the model, potentially losing 
prediction performance and also departing from LLA's post-hoc nature and goal; (ii) instead of using mini-batches to optimize 
variational parameters, they perform a full iteration over the training data to find optimal variational 
parameters. Thus, this results in a potentially slower method than VaLLA, which due to 
early-stopping and stochastic optimization, can avoid iterating over the full dataset.

\begin{figure*}
    \centering
    \begin{minipage}[t][][b]{0.65\textwidth}
    \scalebox{0.75}{
    \begin{tabular}{lccccccccc}
        \toprule
            & \multicolumn{3}{c}{Airline} & \multicolumn{3}{c}{Year} & \multicolumn{3}{c}{Taxi} \\[0.5mm]
            \cmidrule(lr){2-4} \cmidrule(lr){5-7} \cmidrule(lr){8-10}
        Model & NLL & CRPS & CQM & NLL & CRPS & CQM & NLL & CRPS & CQM\\
        \midrule
        MAP       & $5.087$ & $18.436$ & $0.158$ & $3.674$ & $5.056$ & $0.164$ & $3.763$ & $\color{purple}\mathbf{3.753}$ & $\color{teal}\mathbf{0.227}$ \\
        LLA Diag  & $5.096$ & $\color{purple}\mathbf{18.317}$ & $0.144$ & $3.650$ & $4.957$ & $0.122$ & $3.714$ & $3.979$ & $0.270$\\
        LLA KFAC  & $5.097$ & $\color{purple}\mathbf{18.317}$ & $0.144$ & $3.650$ & $\color{teal}\mathbf{4.955}$ & $0.121$ & $3.705$ & $3.977$ & $0.270$\\
        LLA$^*$   & $5.097$ & $\color{teal}\mathbf{18.319}$ & $0.144$ & $3.650$ & $\color{purple}\mathbf{4.954}$ & $0.120$ & $3.718$ & $3.975$ & $0.270$\\
        LLA$^*$ KFAC & $5.097$ & $\color{purple}\mathbf{18.317}$ & $0.144$ & $3.650$ & $\color{purple}\mathbf{4.954}$ & $0.120$ &  $3.718$ & $3.976$ & $0.270$\\
        ELLA      & $5.086$ & $18.437$ & $0.158$ & $3.674$ & $5.056$ & $0.164$ & $3.753$ & $\color{teal}\mathbf{3.754}$ & $\color{teal}\mathbf{0.227}$\\
        VaLLA 100 & $\color{purple}\mathbf{4.923}$ & $18.610$ & $\color{teal}\mathbf{0.109}$ & $\color{teal}\mathbf{3.527}$ & $5.071$ & $\color{teal}\mathbf{0.084}$ & $\color{teal}\mathbf{3.287}$ & $3.968$ & $\color{purple}\mathbf{0.188}$\\
        VaLLA 200 & $\color{purple}\mathbf{4.918}$ & $18.615$ & $\color{purple}\mathbf{0.107}$ & $\color{purple}\mathbf{3.493}$ & $5.026$ & $\color{purple}\mathbf{0.076}$ & $\color{purple}\mathbf{3.280}$ & $3.993$ & $\color{purple}\mathbf{0.188}$\\
        \bottomrule
    \end{tabular}
    }
    \end{minipage}
    \begin{minipage}[t][][b]{0.3\textwidth}
        \vspace{0.2cm}
        \includegraphics[width = 0.9\textwidth]{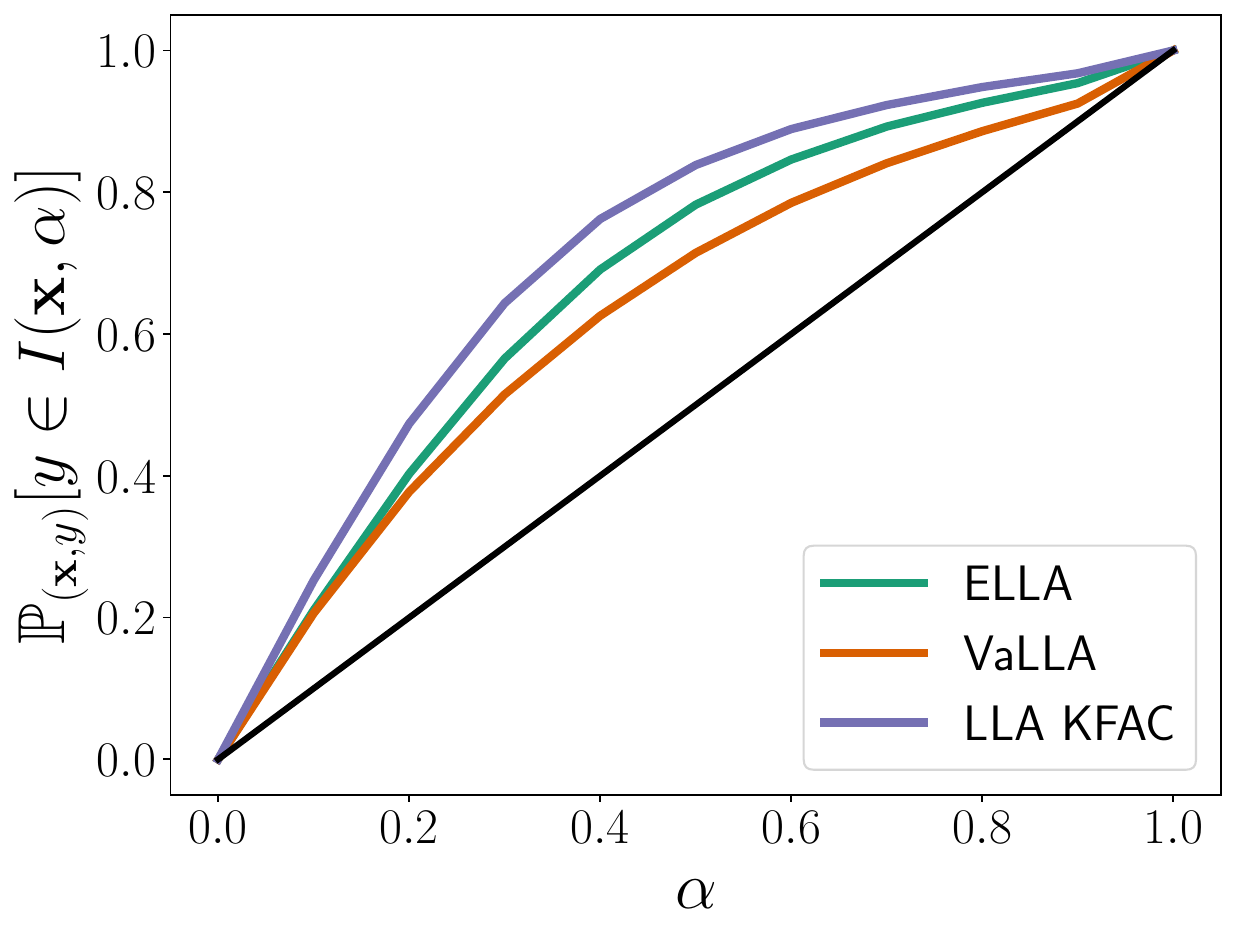}
    \end{minipage}
	\caption{(left) Results on regression datasets. (right) Illustration of CQM on Taxi.  
	Average results across 5 different random seeds (standard deviations always \(<10^{-4}\) and omitted). 
	Best value highlighted in {\color{purple}\textbf{purple}} and second to best in {\color{teal}\textbf{teal}}. $^*$ for Last Layer LLA.}
    \label{fig:regression}
\end{figure*}

Samples from a GP posterior can be efficiently computed using stochastic optimization, eluding
the explicit inversion of the kernel matrix \citep{lin2023sampling}. This approach can be
extended to LLA to generate samples from the GP posterior, avoiding the \(\mathcal{O}(N^3)\) cost \citep{antoran2023sampling}. However, this method cannot provide an estimate of the log-marginal likelihood for hyper-parameter optimization. To address this limitation, \citet{antoran2023sampling} propose
using the \emph{EM-algorithm}, where samples are generated (E-step) and hyper-parameters are
optimized afterwards (M-step) iteratively. The EM algorithm significantly
increases computational cost, as generating a single sample is as expensive as
training the original DNN on the full data.
Finally, the method of \citet{antoran2023sampling} only considers classification problems.

Another GP-based approach for obtaining prediction uncertainty in the context of DNNs is the Spectral-normalized Neural Gaussian Process (SNGP) \citep{liu2023simple}, 
where the last layer of a DNN is replaced by a GP. This approach allows to either (i) fine-tune a pre-trained DNN model, or (ii) train a full DNN model from scratch. 
We compare results with the former in our experiments. However, we have observed that replacing the last layer with a GP often reduces the prediction performance of 
the initial DNN. This is also observed in the results of \citet{liu2023simple}. As a result, this method also lies outside LLA-based methods' main objective, which 
is to preserve the initial DNN predictions.

\section{Experiments}\label{sec:exp}

We compare VaLLA with other methods using the LLA implementation by \citet{daxberger2021laplace}. VaLLA utilizes a batch size of \(100\). In regression, MNIST and FMNIST problems, we train our own DNN (standard multi-layer perceptron), which is stored for reproducibility. In the CIFAR10 experiments with ResNet, we report the results for the other methods given by \citet{deng2022accelerated} and use the same DNN for VaLLA. Hyper-parameters in all LLA variants (diagonal, KFAC, last-layer LLA) are optimized using the marginal log-likelihood estimate. Additional experimental details are given in Appendix~\ref{app:details}. VaLLA's code is available at \url{https://github.com/Ludvins/Variational-LLA}.

\begin{figure*}[t!]
    \centering
    \begin{minipage}[t][][b]{0.45\textwidth}
    \scalebox{0.75}{
        \begin{tabular}{lccccc}
        \toprule
        Model & ACC & NLL & ECE & BRIER & OOD-AUC\\
        \midrule
        MAP  & $\color{teal}\mathbf{97.6}$ & $\color{teal}\mathbf{0.076}$ & $\color{purple}\mathbf{0.008}$ & $\color{teal}\mathbf{0.036}$ & $0.905$\\
        LLA Diag  & $97.4$ & $0.143$ & $0.072$ & $0.053$ & $0.922$\\
        LLA KFAC  & $97.5$ & $0.094$ & $0.029$ & $0.041$ & $\color{teal}\mathbf{0.949}$\\
        LLA$^*$   & $\color{teal}\mathbf{97.6}$ & $0.081$ & $0.015$ & $0.037$ & $0.909$\\
        LLA$^*$ KFAC  & $\color{teal}\mathbf{97.6}$ & $0.081$ & $0.015$ & $0.037$ & $0.909$\\
        ELLA      & $\color{teal}\mathbf{97.6}$ & $\color{teal}\mathbf{0.076}$ & $\color{purple}\mathbf{0.008}$ & $\color{teal}\mathbf{0.036}$ & $0.905$\\
        Sampled LLA & $\color{teal}\mathbf{97.6}$ & $0.087$ & $0.026$ & $0.040$ & $\color{purple} \mathbf{0.954}$\\
        VaLLA 100 & $\color{purple}\mathbf{97.7}$ & $\color{teal}\mathbf{0.076}$ & $\color{teal}\mathbf{0.010}$ & $\color{teal}\mathbf{0.036}$ & $0.916$\\
        VaLLA 200 & $\color{purple}\mathbf{97.7}$ & $\color{purple}\mathbf{0.075}$ & $\color{teal}\mathbf{0.010}$ & $\color{purple}\mathbf{0.035}$ & $0.921$\\
        \bottomrule
    \end{tabular}
    }
    \end{minipage}
    \begin{minipage}[t][][b]{0.49\textwidth}
        \centering
        \includegraphics[width = 0.85\textwidth]{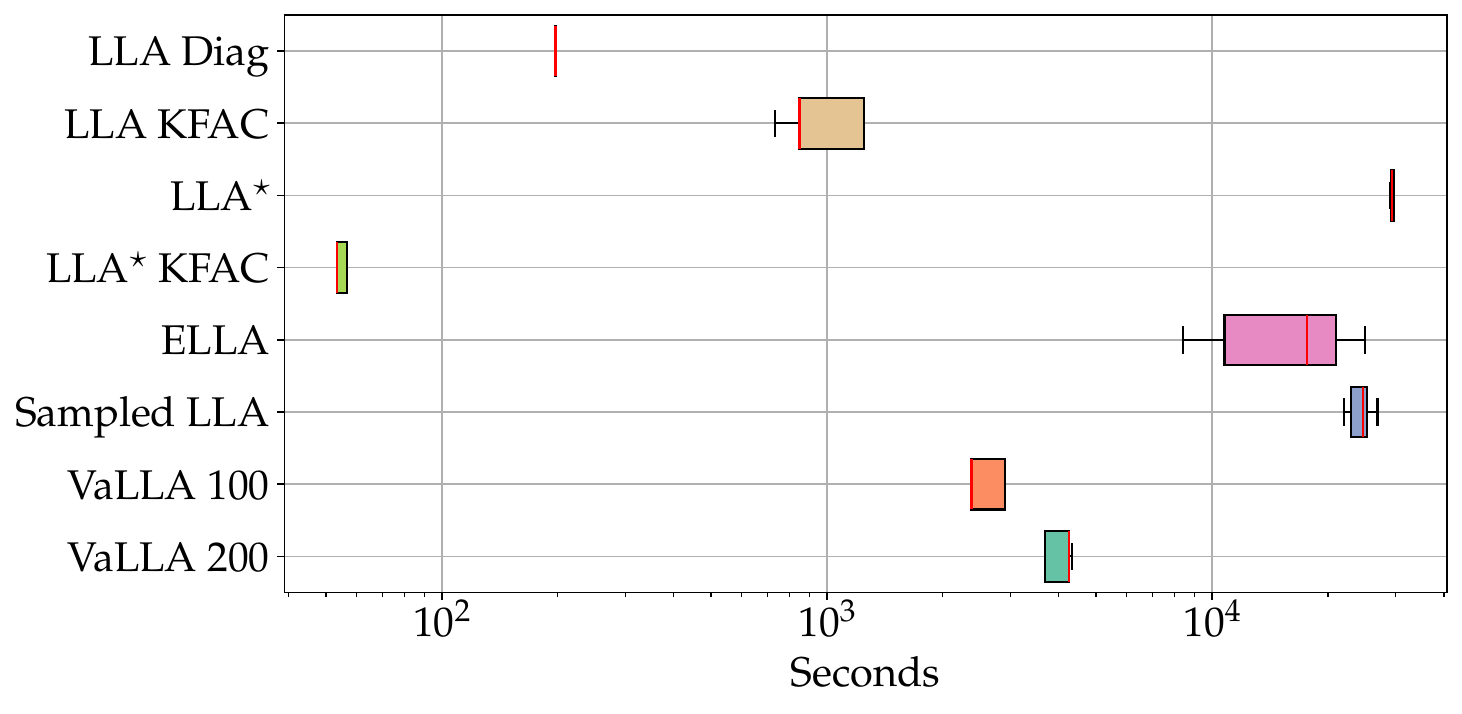}
    \end{minipage}
	\caption{(left) MNIST experiments.  
	Results averaged over 5 different random seeds (standard deviations \(<10^{-4}\) in all cases and omitted). 
	(right) Box-plots of training times in seconds. ELLA considers 10 prior values chosen using a validation set.
	Sampled-LLA uses 8 EM steps and 32 samples. Best value is highlighted in {\color{purple}\textbf{purple}} and second to best in {\color{teal}\textbf{teal}}. $^*$ for Last Layer LLA.}
    \label{fig:mnist}
\end{figure*}
\begin{figure*}[t!]
    \centering
    \begin{minipage}[t][][b]{0.45\textwidth}
    \hspace*{0.5cm}\scalebox{0.7}{
        \centering
        \begin{tabular}{lccccc}
        \toprule
        Model & ACC & NLL & ECE & BRIER & OOD-AUC\\
        \midrule
        MAP       & $86.6$ & $0.373$ & $\color{teal}\mathbf{0.009}$ & $0.193$ & $0.874$ \\
        LLA Diag  & $86.2$ & $0.397$ & $0.043$ & $0.201$ & $0.914$ \\
        LLA KFAC  & $86.5$ & $0.377$ & $0.014$ & $0.194$ & $\color{teal}\mathbf{0.932}$ \\
        LLA$^*$   & $86.6$ & $0.373$ & $\color{purple}\mathbf{0.008}$ & $0.193$ & $0.882$\\
        LLA$^*$ KFAC  & $86.6$ & $0.373$ & $\color{purple}\mathbf{0.008}$ & $0.193$ & $0.880$ \\
        ELLA      & $86.6$ & $0.373$ & $\color{purple}\mathbf{0.008}$ & $0.193$ & $0.874$\\
        VaLLA 100 & $\color{teal}\mathbf{87.4}$ & $\color{teal}\mathbf{0.335}$ & $0.011$ & $\color{teal}\mathbf{0.182}$ & $0.923$ \\
        VaLLA 200 & $\color{purple}\mathbf{87.6}$ & $\color{purple}\mathbf{0.332}$ & $0.013$ & $\color{purple}\mathbf{0.181}$ & $\color{purple}\mathbf{0.933}$ \\
        \bottomrule
    \end{tabular}
    }
    \end{minipage}
    \begin{minipage}[t][][b]{0.54\textwidth}
        \includegraphics[width = 1\textwidth]{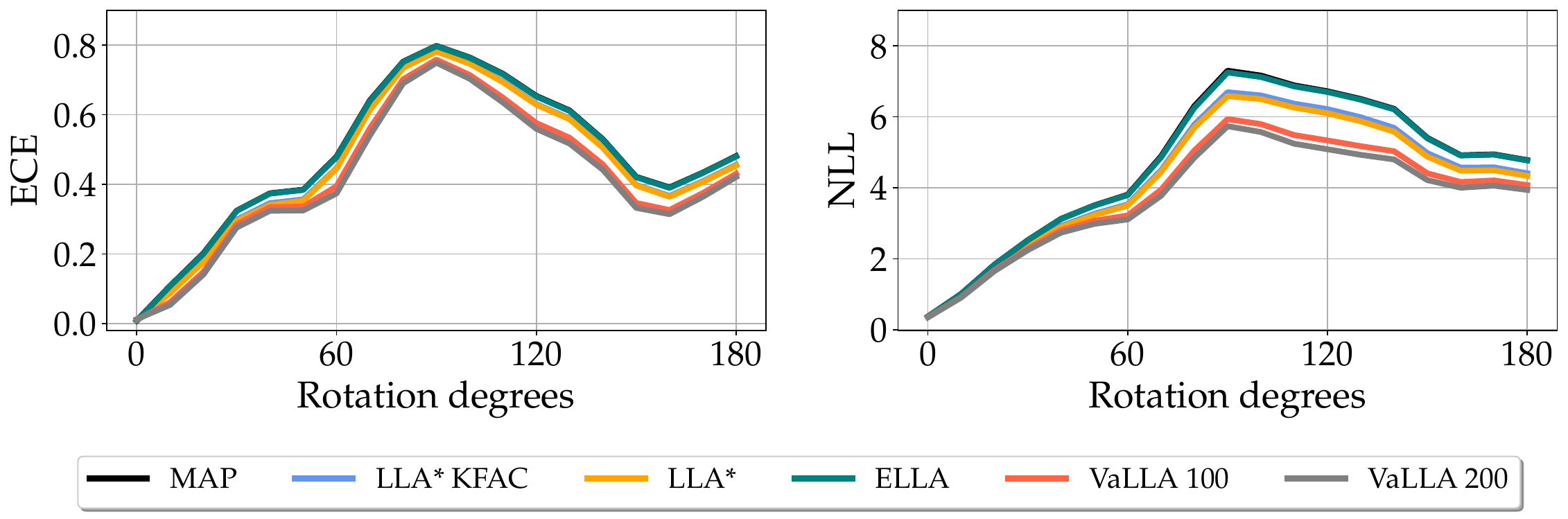}
    \end{minipage}
	\caption{(left) Results on FMNIST. Results are averaged over 5 different random seeds (standard deviations are lower than \(10^{-4}\) and omitted). 
	Best value is highlighted in {\color{purple}\textbf{purple}} and second to best in {\color{teal}\textbf{teal}}. $^*$ for Last Layer LLA.  (right) ECE and NLL for rotated FMNIST.}
    \label{fig:fmnist}
\end{figure*}

\subsection{Synthetic Regression}

We compare the predictive distribution of VaLLA with that of LLA (which is considered the optimal method),
other LLA variants and ELLA, on the 1-D regression problem of \citet{izmailov2020subspace}. 
In ELLA and VaLLA, we use the optimal hyper-parameters from LLA. 
The results in Figure~\ref{fig:intro} illustrate that VaLLA's predictive distribution closely aligns with that of LLA.
Figure~\ref{fig:inducing} (see Appendix~\ref{app:inducing}) depicts the predictive distributions of VaLLA and ELLA for varying numbers of inducing points and points in the Nystr\"om approximation, respectively. 
It shows that VaLLA converges to the true posterior faster than ELLA, with VaLLA tending to overestimate the predictive variance while ELLA underestimates it. In Figure~\ref{fig:val} (see Appendix~\ref{app:inducing}) we observe the effect of tuning the prior variance in VaLLA in another toy 1-D problem, with and without early-stopping. Notably, early stopping, using a validation set, prevents overly small predictive variances in VaLLA. Finally, we observe that when VaLLA estimates the prior variance by maximizing (\ref{eq:alpha}), it tends to underestimate LLA's predictive variance.

\subsection{Airline, Year and Taxi Regression Problems}

We carry out experiments on large regression datasets. (i) The \textit{Year} dataset (UCI) with $515,345$ instances and $90$ features. We use the original train/test splits. 
(ii) The \textit{US flight delay (Airline)} dataset \citep{dutordoir2020sparse}. Following 
\citet{ortega2023deep} we use the first $700,000$ instances for training and the next $100,000$ for testing. 
$8$ features are considered: month, day of the month, day of the week, plane age, air time, distance, arrival 
time and departure time. (iii) The \textit{Taxi dataset}, with data recorded on January, 2023 \citep{salimbeni2017doubly}. $9$ attributes are considered: time of day, day of week, day of month, month, PULocationID, DOLocationID, distance and duration. We filter trips shorter than 10 seconds and larger than 5 hours, resulting in $3$ million instances. The first \(80\%\) is used as train data, the next \(10\%\) as validation data,  and the last (10\%) as test data. In all experiments, a 3-layer DNN with 200 units, \emph{tanh} activations and L2 regularization is employed. VaLLA and ELLA use \(100\) inducing points and 100 random points, respectively. We carry out \(40,000\) iterations of mini-batch size 100 in VaLLA. However, in the Taxi dataset, with nearly \(3\) million data instances, due to early-stopping, training finishes at \(5,000\) iterations for one of the random seed initializations (this value differs for each random). This means that 500,000 points are visited during training for that seed, which is only $16.6\%$ of the complete dataset.

The table on the l.h.s. of Figure \ref{fig:regression} presents the averaged results over 5 random seeds. LLA is not considered here due to intractability. Negative log likelihood (NLL), continuous ranked probability score (CRPS) \citep{gneiting2007strictly} and a centered quantile metric (CQM), described below, are reported. We observe that VaLLA performs best according to NLL and CQM, while it gives worse results in terms of CRPS compared to the other methods. 

\paragraph{Centered Quantile Metric (CQM).} In regression, CQM assesses the calibration of the predictions, extending the Expected Calibration Error (ECE) to regression problems with Gaussian predictions \emph{with the same mean} but different variance. CQM calculates the \emph{centered interval around the mean} with probability mass $\alpha \in (0, 1) $. For Gaussian predictions \(\mathcal{ N}(\mu(\mathbf{x}), \sigma^2(\mathbf{x}))\), the open interval is defined as \(I(\mathbf{x}, \alpha)=(\lambda(-\alpha), \lambda(\alpha))\), where $\lambda(\alpha) = \Phi_{\mu(\mathbf{x}),\sigma^2(\mathbf{x})}^{-1}(\tfrac{1 + \alpha}{2})$ with \(\Phi_{\mu(\mathbf{x}),\sigma^2(\mathbf{x})}\) the CDF of a Gaussian with mean $\mu(\mathbf{x})$ and variance $\sigma^2(\mathbf{x})$. The fraction $\gamma$ of test points falling inside the interval is then computed. If the predictive distribution is well calibrated, \(\gamma \approx \alpha\). Formally, 
\begin{align}
	\text{CQM} & = \int_0^1 \ \Big|\mathbb{P}_{(\mathbf{x}^\star, y^\star)}\left[ y^\star \in I(\mathbf{x}^\star, \alpha) \right] - \alpha\Big| \ d\alpha\,.\label{eq:CQM}
\end{align}
All methods utilize the pre-trained DNN solution as predictive mean. Thus, the differences in \(I(\mathbf{x}, \alpha)\) stem from the predictive variance. Evaluating the integrand in (\ref{eq:CQM}) on a grid of \(\alpha\) values allows us to visually interpret the uncertainty estimation of each method. The r.h.s. of Figure~\ref{fig:regression} shows \(\mathbb{P}_{(\mathbf{x}, y)}\left[ y \in I(\mathbf{x}, \alpha) \right]\) for several models on the Taxi dataset. 
In general, all methods tend to overestimate the actual predictive variance, as evidenced by the values above the diagonal. The l.h.s. of Figure \ref{fig:regression} shows CQM estimated using trapezoid integration with $11$ points. We refer to Appendix~\ref{app:Q} for more details on the CQM metric.

\begin{table*}[t!]
    \centering
    \scalebox{0.78}{
    \begin{tabular}{lcccccccccccc|c}
        \toprule
            & \multicolumn{3}{c}{ResNet-20} & \multicolumn{3}{c}{ResNet-32} & \multicolumn{3}{c}{ResNet-44} & \multicolumn{3}{c|}{ResNet-56} & \multirow{2}{*}{Mean Rank}\\[0.5mm]
            \cmidrule(lr){2-4} \cmidrule(lr){5-7} \cmidrule(lr){8-10} \cmidrule(lr){11-13}
        Method & ACC & NLL & ECE & ACC & NLL & ECE & ACC & NLL & ECE & ACC & NLL & ECE \\
        \midrule
        MAP       & $\color{teal} \mathbf{92.6}$ & $0.282$ & $0.039$ & $\color{purple} \mathbf{93.5}$ & $0.292$ & $0.041$ & $\color{purple} \mathbf{94.0}$ & $0.275$ & $0.039$ & $\color{purple} \mathbf{94.4}$ & $0.252$ & $0.037$ & $-$\\
        MF-VI       & $\color{purple} \mathbf{92.7}$ & $\color{teal}\mathbf{0.231}$ & $0.016$ &$\color{purple} \mathbf{93.5}$ & $0.222$ & $0.020$ & $\color{teal} \mathbf{93.9}$  & $0.206$ & $0.018$ & $\color{purple} \mathbf{94.4}$ & $0.188$ & $0.016$ & $-$\\
        SNGP       & $92.4$ & $0.266$ & $0.024$ &$93.2$ & $0.256$ & $0.025$ & $93.8$  & $0.242$ & $0.028$ & $93.8$ & $0.229$ & $0.022$ & $-$\\
        \midrule
        GP - Subset   & $\color{teal} \mathbf{92.6}$ & $0.555$ & $0.299$ &$\color{teal} \mathbf{93.4}$ & $0.462$ & $0.247$ & $93.6$  & $0.424$ & $0.225$ & $\color{purple} \mathbf{94.4}$ & $0.403$ & $0.221$ & $-$\\
        LLA Diag  & $92.2$ & $0.728$ & $0.404$ & $92.7$ & $0.755$ & $0.430$ & $92.8$ & $0.778$ & $0.445$ & $\color{teal}\mathbf{92.9}$ & $0.843$ & $0.480$ & $-$\\
        LLA KFAC  & $92.0$ & $0.852$ & $0.467$ & $91.8$ & $1.027$ & $0.547$ & $91.4$ & $1.091$ & $0.566$ & $89.8$ & $1.174$ & $0.579$ & $-$\\
        LLA$^*$   & $\color{teal} \mathbf{92.6}$ & $0.269$ & $0.034$ & $\color{purple} \mathbf{93.5}$ & $0.259$ & $0.033$ & $\color{purple} \mathbf{94.0}$ & $0.237$ & $0.028$ & $\color{purple} \mathbf{94.4}$ & $0.213$ & $0.022$ & $-$\\
        LLA$^*$ KFAC &$\color{teal} \mathbf{92.6}$ & $0.271$ & $0.035$ & $\color{purple} \mathbf{93.5}$ & $0.260$ & $0.033$ & $\color{purple} \mathbf{94.0}$ & $0.232$ & $0.028$ & $\color{purple} \mathbf{94.4}$ & $0.202$ & $0.024$ & $-$\\
        \midrule
        ELLA      & $92.5$ & $0.233$ & $0.009$ & $\color{purple} \mathbf{93.5}$ & $\color{teal}\mathbf{0.215}$ & $\color{teal}\mathbf{0.008}$ & $\color{teal} \mathbf{93.9}$ & $0.204$ & $\color{purple} \mathbf{0.007}$ & $\color{purple} \mathbf{94.4}$ & $0.187$ & $\color{purple} \mathbf{0.007}$ & $2.375$\\
        Sampled LLA & $92.5$ & $\color{teal} \mathbf{0.231}$ & $\color{purple}\color{purple} \mathbf{0.006}$ & $\color{purple}\mathbf{93.5}$ & $0.217$ & $\color{teal}\mathbf{0.008}$ & $\color{purple}\mathbf{94.0}$ & $\color{teal} \mathbf{0.200}$ & $\color{purple}\mathbf{0.007}$ & $\color{purple}{\mathbf{94.4}}$ & $\color{teal} \mathbf{0.185}$ & $0.015$ & $\color{teal}\mathbf{2.000}$\\
        VaLLA     & $\color{teal} \mathbf{92.6}$ & $\color{purple} \mathbf{0.228}$ & $\color{teal}\mathbf{0.007}$ & $\color{purple} \mathbf{93.5}$ & $\color{purple} \mathbf{0.211}$ & $\color{purple} \mathbf{0.007}$ & $\color{purple} \mathbf{94.0}$ & $\color{purple} \mathbf{0.198}$ & $\color{teal}\mathbf{0.008}$ & $\color{purple} \mathbf{94.4}$ & $\color{purple} \mathbf{0.183}$ & $\color{teal} \mathbf{0.009}$ & $\color{purple}\mathbf{1.375}$\\
        \bottomrule
    \end{tabular}
    }
    \caption{Results on CIFAR10. ACC, NLL and ECE are computed using Monte-Carlo estimation. 
	Best value highlighted in {\color{purple}\textbf{purple}} and second to best in {\color{teal}\textbf{teal}}. 
	Sampled LLA uses \(64\) samples. ELLA uses \(M=2000\) points and \(K=20\). 
	Average results over \(5\) different random seeds (standard deviations \(<10^{-3}\) in all cases and omitted). $^*$ for Last Layer LLA.
	Mean rank only considers both NLL and ECE.}
    \label{fig:resnet}
	\vspace{-.1cm}
\end{table*}

\subsection{Image Classification Problems}


\paragraph{MNIST and FMNIST.}
We employ a 2-layer fully connected DNN with $200$ units in each layer and \emph{tanh} 
activations. In VaLLA we considered 100 and 200 inducing points, while in ELLA, 2000 random points are used.
The Out-of-distribution (OOD) detection ability of each method is evaluated using the entropy of the predictive distribution as a score. We compute the area under the ROC curve (AUC) of the binary 
problem that distinguishes between instances from pairs of datasets MNIST/FMNIST and FMNIST/MNIST \citep{immer2021improving}. Moreover, in FMNIST we also assess the robustness of the predictive distribution by rotating the test images up to \(180\) degrees and computing the ECE and NLL on rotated images \citep{ovadia2019can}. 

The left table in Figure~\ref{fig:mnist} shows the results on MNIST. VaLLA gives better uncertainty estimates in terms of NLL and the Brier score but performs less effectively in terms of ECE. Remarkably, VaLLA improves prediction accuracy (ACC) due to the approximation of \citet{daxberger2021} to compute 
class probabilities in multi-class problems. In terms of OOD-AUC VaLLA outperforms the MAP solution but lags behind other methods \emph{s.a.} Sampled-LLA or LLA with Kronecker approximations. 
Figure~\ref{fig:mnist} (right) illustrates the training times for each method, with VaLLA being faster than ELLA, Sampled-LLA or Last-Layer LLA.

Finally, the left table in Figure~\ref{fig:fmnist} displays the results on FMNIST. Here, VaLLA excels in prediction accuracy and provides the best uncertainty estimates in terms of NLL and the Brier score. Although 
it does not perform as well in terms of ECE, the differences are small. VaLLA also achieves the best 
results in OOD-AUC. Figure~\ref{fig:fmnist} (right) shows VaLLA holds better performance in terms of ECE and NLL as the test images' corruption increases (rotation level), indicating the greater robustness of VaLLA's predictive distribution.

\begin{figure}
    \centering
    \includegraphics[width = 0.45\textwidth]{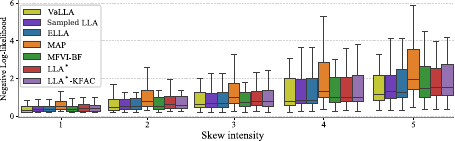}
    \caption{Results on corrupted CIFAR10 with ResNet56. Sampled LLA uses \(64\) samples 
	and ELLA uses \(M=2000\) and \(K=20\).}
    \label{fig:resnet_corruptions}
\end{figure}

\paragraph{CIFAR10 and ResNet.} Various ResNets architectures are used, and the corresponding pre-trained models are those 
of \citet{deng2022accelerated} (accessible \href{https://github.com/chenyaofo/pytorch-cifar-models}{here}). 
Table~\ref{fig:resnet} shows ACC, NLL and ECE for each method, including LLA variants, a mean-field VI approach \citep{deng2023bayesadapter}, fine-tuned SNGP \citep{liu2023simple} and \citet{immer2021improving}'s approach of using the GP interpretation of LLA and a random subset of \(500\) instances from the training set (GP-Subset). For the latter, we used as the prior parameter the weight decay value used when training the MAP solution,
\(\sigma_0^2 = 0.04\), which is the one suggested by \citet{immer2021improving}. Finally, we did not scale the prior by the subset size, as it resulted in worse results. For this experiment, VaLLA is trained for \(40.000\) iterations or until Early-Stopping raises. Since we use the same pre-trained models, the results of all other methods are consistent with those reported by \citet{deng2022accelerated}. VaLLA with \(M_\beta=100\) outperforms other methods in most cases, always being either the best or second-best method.
Figure~\ref{fig:resnet_corruptions} shows the NLL on the perturbed test set with 5 increasing levels of \(19\) image corruptions \citep{deng2022accelerated}. Each box-plot summarizes the test NLL for each intensity level across all $19$ corruptions. The results again highlight VaLLA's robust predictive distribution, achieving also lower NLL compared to the other methods.

\section{Conclusions}

We introduced VaLLA, a method derived from the formulation of a generalized sparse GP that offers the flexibility to fix the predictive mean to any desired function in the RKHS. 
VaLLA excels in computing error bars for pre-trained DNNs with a vast number of parameters on extensive datasets, handling even millions of training instances. 
VaLLA's applicability spans both regression and classification problems, showcasing costs independent of the number of training points $N$. 
In comparison, the Nystr\"om approximation by \citet{deng2022accelerated} incurs a linear cost in $N$, unless early-stopping is employed. 
Furthermore, VaLLA surpasses the sample-then-optimize method of \citet{antoran2023sampling} in terms of speed, while also providing predictive distributions robust to input corruptions. 
In essence, VaLLA stands out by delivering robust predictive distributions akin to LLA, all while maintaining noteworthy computational efficiency.

VALLA may also offer valuable advantages for Bayesian optimization (BO). 
With more accurate uncertainty assessments, VaLLA is expected to guide the optimization process more efficiently, 
potentially reducing the number of evaluations needed to find optimal solutions. 
VaLLA may also be useful for scaling BO to problems where cheap evaluations are available,
since in that setting it is required the fast fitting of probabilistic models to thousands of training instances.

\section*{Acknowledgements}
Authors gratefully acknowledge the use of the facilities of Centro de Computacion Cientifica (CCC) at Universidad Autónoma de Madrid. The authors acknowledge financial support from project PID2022-139856NB-I00 funded by MCIN/ AEI / 10.13039/501100011033 / FEDER, UE and project PID2019-106827GB-I00 / AEI / 10.13039/501100011033 and from the Autonomous Community of Madrid (ELLIS Unit Madrid). The authors also acknowledge financial support from project TED2021-131530B-I00, funded by MCIN/AEI /10.13039/501100011033 and by the European Union NextGenerationEU PRTR.

\section*{Impact Statement}
As machine learning models play an ever-growing role in influencing decisions with substantial consequences for society, industry, and individuals ---such as ensuring the safety of autonomous vehicles \citep{mcallister2017concrete} and improving disease detection \citep{sajda2006machine, singh2021better}--- it becomes imperative to possess a comprehensive understanding of the employed methodologies and be capable of offering robust performance assurances. Our diligent examination of VaLLA's performance across diverse datasets and tasks as part of our empirical assessment showcases its adaptability to various domain-specific datasets.

\bibliographystyle{icml2024}
\bibliography{refs}

\begin{thebibliography}{50}
\providecommand{\natexlab}[1]{#1}
\providecommand{\url}[1]{\texttt{#1}}
\expandafter\ifx\csname urlstyle\endcsname\relax
  \providecommand{\doi}[1]{doi: #1}\else
  \providecommand{\doi}{doi: \begingroup \urlstyle{rm}\Url}\fi

\bibitem[Antor{\'{a}}n et~al.(2023)Antor{\'{a}}n, Padhy, Barbano, Nalisnick, Janz, and Hern{\'{a}}ndez{-}Lobato]{antoran2023sampling}
Antor{\'{a}}n, J., Padhy, S., Barbano, R., Nalisnick, E.~T., Janz, D., and Hern{\'{a}}ndez{-}Lobato, J.~M.
\newblock Sampling-based inference for large linear models, with application to linearised {L}aplace.
\newblock In \emph{International Conference on Learning Representations}, 2023.

\bibitem[Bergamin et~al.(2023)Bergamin, Moreno-Mu{\~n}oz, Hauberg, and Arvanitidis]{bergamin2023riemannian}
Bergamin, F., Moreno-Mu{\~n}oz, P., Hauberg, S., and Arvanitidis, G.
\newblock Riemannian {L}aplace approximations for {B}ayesian neural networks.
\newblock \emph{Advances in Neural Information Processing Systems}, 2023.

\bibitem[Bishop(2006)]{bishop2006}
Bishop, C.~M.
\newblock \emph{Pattern Recognition and Machine Learning (Information Science and Statistics)}.
\newblock Springer, 2006.

\bibitem[Blundell et~al.(2015)Blundell, Cornebise, Kavukcuoglu, and Wierstra]{blundell2015weight}
Blundell, C., Cornebise, J., Kavukcuoglu, K., and Wierstra, D.
\newblock Weight uncertainty in neural network.
\newblock In \emph{International Conference on Machine Learning}, pp.\  1613--1622, 2015.

\bibitem[Botev et~al.(2017)Botev, Ritter, and Barber]{botev2017practical}
Botev, A., Ritter, H., and Barber, D.
\newblock Practical {G}auss-{N}ewton optimisation for deep learning.
\newblock In \emph{International Conference on Machine Learning}, pp.\  557--565, 2017.

\bibitem[Bui et~al.(2017)Bui, Yan, and Turner]{bui2017}
Bui, T.~D., Yan, J., and Turner, R.~E.
\newblock A unifying framework for {G}aussian process pseudo-point approximations using power expectation propagation.
\newblock \emph{Journal of Machine Learning Research}, 18:\penalty0 1--72, 2017.

\bibitem[Chen et~al.(2014)Chen, Fox, and Guestrin]{chen2014stochastic}
Chen, T., Fox, E., and Guestrin, C.
\newblock Stochastic gradient hamiltonian monte carlo.
\newblock In \emph{International Conference on Machine Learning}, pp.\  1683--1691, 2014.

\bibitem[Cheng \& Boots(2016)Cheng and Boots]{cheng2016incremental}
Cheng, C.-A. and Boots, B.
\newblock Incremental variational sparse {G}aussian process regression.
\newblock \emph{Advances in Neural Information Processing Systems}, 29:\penalty0 4403--4411, 2016.

\bibitem[Cheng \& Boots(2017)Cheng and Boots]{cheng2017variational}
Cheng, C.-A. and Boots, B.
\newblock Variational inference for {G}aussian process models with linear complexity.
\newblock \emph{Advances in Neural Information Processing Systems}, 30:\penalty0 5184--5194, 2017.

\bibitem[Daxberger et~al.(2021{\natexlab{a}})Daxberger, Kristiadi, Immer, Eschenhagen, Bauer, and Hennig]{daxberger2021laplace}
Daxberger, E., Kristiadi, A., Immer, A., Eschenhagen, R., Bauer, M., and Hennig, P.
\newblock Laplace {R}edux - {E}ffortless {B}ayesian deep learning.
\newblock \emph{Advances in Neural Information Processing Systems}, 34:\penalty0 20089--20103, 2021{\natexlab{a}}.

\bibitem[Daxberger et~al.(2021{\natexlab{b}})Daxberger, Nalisnick, Allingham, Antoran, and Hernandez-Lobato]{daxberger2021}
Daxberger, E., Nalisnick, E., Allingham, J.~U., Antoran, J., and Hernandez-Lobato, J.~M.
\newblock {B}ayesian deep learning via subnetwork inference.
\newblock In \emph{International Conference on Machine Learning}, pp.\  2510--2521, 2021{\natexlab{b}}.

\bibitem[Deng \& Zhu(2023)Deng and Zhu]{deng2023bayesadapter}
Deng, Z. and Zhu, J.
\newblock Bayesadapter: Being {B}ayesian, inexpensively and reliably, via {B}ayesian fine-tuning.
\newblock In \emph{Asian Conference on Machine Learning}, pp.\  280--295, 2023.

\bibitem[Deng et~al.(2022)Deng, Zhou, and Zhu]{deng2022accelerated}
Deng, Z., Zhou, F., and Zhu, J.
\newblock Accelerated linearized {L}aplace approximation for {B}ayesian deep learning.
\newblock \emph{Advances in Neural Information Processing Systems}, 35:\penalty0 2695--2708, 2022.

\bibitem[Dutordoir et~al.(2020)Dutordoir, Durrande, and Hensman]{dutordoir2020sparse}
Dutordoir, V., Durrande, N., and Hensman, J.
\newblock Sparse {G}aussian processes with spherical harmonic features.
\newblock In \emph{International Conference on Machine Learning}, pp.\  2793--2802, 2020.

\bibitem[Foong et~al.(2019)Foong, Li, Hern{\'a}ndez-Lobato, and Turner]{foong2019between}
Foong, A.~Y., Li, Y., Hern{\'a}ndez-Lobato, J.~M., and Turner, R.~E.
\newblock `{I}n-{B}etween' {U}ncertainty in {B}ayesian neural networks.
\newblock \emph{ICML Workshop on Uncertainty and Robustness in Deep Learning}, 2019.

\bibitem[Gneiting \& Raftery(2007)Gneiting and Raftery]{gneiting2007strictly}
Gneiting, T. and Raftery, A.~E.
\newblock Strictly proper scoring rules, prediction, and estimation.
\newblock \emph{Journal of the American statistical Association}, 102:\penalty0 359--378, 2007.

\bibitem[Graves(2011)]{graves2011practical}
Graves, A.
\newblock Practical variational inference for neural networks.
\newblock \emph{Advances in Neural Information Processing Systems}, 24:\penalty0 2348--2356, 2011.

\bibitem[Guo et~al.(2017)Guo, Pleiss, Sun, and Weinberger]{guo2017calibration}
Guo, C., Pleiss, G., Sun, Y., and Weinberger, K.~Q.
\newblock On calibration of modern neural networks.
\newblock In \emph{International conference on machine learning}, pp.\  1321--1330, 2017.

\bibitem[He et~al.(2016)He, Zhang, Ren, and Sun]{he2016deep}
He, K., Zhang, X., Ren, S., and Sun, J.
\newblock Deep residual learning for image recognition.
\newblock In \emph{Proceedings of the IEEE Conference on Computer Vision and Pattern Recognition}, pp.\  770--778, 2016.

\bibitem[Immer et~al.(2021)Immer, Korzepa, and Bauer]{immer2021improving}
Immer, A., Korzepa, M., and Bauer, M.
\newblock Improving predictions of {B}ayesian neural nets via local linearization.
\newblock In \emph{International Conference on Artificial Intelligence and Statistics}, pp.\  703--711, 2021.

\bibitem[Izmailov et~al.(2020)Izmailov, Maddox, Kirichenko, Garipov, Vetrov, and Wilson]{izmailov2020subspace}
Izmailov, P., Maddox, W.~J., Kirichenko, P., Garipov, T., Vetrov, D., and Wilson, A.~G.
\newblock Subspace inference for {B}ayesian deep learning.
\newblock In \emph{Uncertainty in Artificial Intelligence}, pp.\  1169--1179, 2020.

\bibitem[Kendall \& Gal(2017)Kendall and Gal]{kendall2017uncertainties}
Kendall, A. and Gal, Y.
\newblock What uncertainties do we need in {B}ayesian deep learning for computer vision?
\newblock \emph{Advances in Neural Information Processing Systems}, 30:\penalty0 5574--5584, 2017.

\bibitem[Khan et~al.(2019)Khan, Immer, Abedi, and Korzepa]{khan2019approximate}
Khan, M. E.~E., Immer, A., Abedi, E., and Korzepa, M.
\newblock Approximate inference turns deep networks into {G}aussian processes.
\newblock \emph{Advances in Neural Information Processing Systems}, 32, 2019.

\bibitem[Kingma \& Ba(2015)Kingma and Ba]{kingma2014adam}
Kingma, D.~P. and Ba, J.
\newblock Adam: {A} method for stochastic optimization.
\newblock \emph{International Conference for Learning Representations}, 2015.

\bibitem[Lawrence(2001)]{lawrence2001variational}
Lawrence, N.~D.
\newblock \emph{Variational inference in probabilistic models}.
\newblock PhD thesis, Citeseer, 2001.

\bibitem[Lee et~al.(2022)Lee, Feng, Humt, M{\"u}ller, and Triebel]{lee2022trust}
Lee, J., Feng, J., Humt, M., M{\"u}ller, M.~G., and Triebel, R.
\newblock Trust your robots! predictive uncertainty estimation of neural networks with sparse {G}aussian processes.
\newblock In \emph{Conference on Robot Learning}, pp.\  1168--1179, 2022.

\bibitem[Leibig et~al.(2017)Leibig, Allken, Ayhan, Berens, and Wahl]{leibig2017leveraging}
Leibig, C., Allken, V., Ayhan, M.~S., Berens, P., and Wahl, S.
\newblock Leveraging uncertainty information from deep neural networks for disease detection.
\newblock \emph{Scientific reports}, 7:\penalty0 1--14, 2017.

\bibitem[Li \& Gal(2017)Li and Gal]{liG17}
Li, Y. and Gal, Y.
\newblock Dropout inference in {B}ayesian neural networks with alpha-divergences.
\newblock In \emph{International Conference on Machine Learning}, pp.\  2052--2061, 2017.

\bibitem[Lin et~al.(2024)Lin, Antor{\'a}n, Padhy, Janz, Hern{\'a}ndez-Lobato, and Terenin]{lin2023sampling}
Lin, J.~A., Antor{\'a}n, J., Padhy, S., Janz, D., Hern{\'a}ndez-Lobato, J.~M., and Terenin, A.
\newblock Sampling from {G}aussian process posteriors using stochastic gradient descent.
\newblock \emph{Advances in Neural Information Processing Systems}, 36, 2024.

\bibitem[Liu et~al.(2023)Liu, Padhy, Ren, Lin, Wen, Jerfel, Nado, Snoek, Tran, and Lakshminarayanan]{liu2023simple}
Liu, J.~Z., Padhy, S., Ren, J., Lin, Z., Wen, Y., Jerfel, G., Nado, Z., Snoek, J., Tran, D., and Lakshminarayanan, B.
\newblock A simple approach to improve single-model deep uncertainty via distance-awareness.
\newblock \emph{Journal of Machine Learning Research}, 24:\penalty0 1--63, 2023.

\bibitem[MacKay(1992{\natexlab{a}})]{mackay1992evidence}
MacKay, D.~J.
\newblock The evidence framework applied to classification networks.
\newblock \emph{Neural computation}, 4:\penalty0 720--736, 1992{\natexlab{a}}.

\bibitem[MacKay(1992{\natexlab{b}})]{mackay1992practical}
MacKay, D.~J.
\newblock A practical {B}ayesian framework for backpropagation networks.
\newblock \emph{Neural computation}, 4:\penalty0 448--472, 1992{\natexlab{b}}.

\bibitem[Mackay(1992)]{mackay1992bayesian}
Mackay, D. J.~C.
\newblock \emph{{B}ayesian methods for adaptive models}.
\newblock California Institute of Technology, 1992.

\bibitem[Martens(2020)]{martens2020new}
Martens, J.
\newblock New insights and perspectives on the natural gradient method.
\newblock \emph{The Journal of Machine Learning Research}, 21:\penalty0 5776--5851, 2020.

\bibitem[Martens \& Grosse(2015)Martens and Grosse]{martens2015optimizing}
Martens, J. and Grosse, R.
\newblock Optimizing neural networks with kronecker-factored approximate curvature.
\newblock In \emph{International Conference on Machine Learning}, pp.\  2408--2417, 2015.

\bibitem[McAllister et~al.(2017)McAllister, Gal, Kendall, Van Der~Wilk, Shah, Cipolla, and Weller]{mcallister2017concrete}
McAllister, R., Gal, Y., Kendall, A., Van Der~Wilk, M., Shah, A., Cipolla, R., and Weller, A.
\newblock Concrete problems for autonomous vehicle safety: advantages of bayesian deep learning.
\newblock In \emph{Proceedings of the 26th International Joint Conference on Artificial Intelligence}, pp.\  4745--4753, 2017.

\bibitem[Neal(2012)]{neal2012bayesian}
Neal, R.~M.
\newblock \emph{{B}ayesian learning for neural networks}, volume 118.
\newblock Springer Science \& Business Media, 2012.

\bibitem[Novak et~al.(2022)Novak, Sohl-Dickstein, and Schoenholz]{novak2022fast}
Novak, R., Sohl-Dickstein, J., and Schoenholz, S.~S.
\newblock Fast finite width neural tangent kernel.
\newblock In \emph{International Conference on Machine Learning}, pp.\  17018--17044, 2022.

\bibitem[Ortega et~al.(2023)Ortega, Santana, and Hern{\'a}ndez-Lobato]{ortega2023deep}
Ortega, L.~A., Santana, S.~R., and Hern{\'a}ndez-Lobato, D.
\newblock Deep variational implicit processes.
\newblock In \emph{International Conference of Learning Representations}, 2023.

\bibitem[Ovadia et~al.(2019)Ovadia, Fertig, Ren, Nado, Sculley, Nowozin, Dillon, Lakshminarayanan, and Snoek]{ovadia2019can}
Ovadia, Y., Fertig, E., Ren, J., Nado, Z., Sculley, D., Nowozin, S., Dillon, J., Lakshminarayanan, B., and Snoek, J.
\newblock Can you trust your model's uncertainty? evaluating predictive uncertainty under dataset shift?
\newblock \emph{Advances in Neural Information Processing Systems}, pp.\  13969--13980, 2019.

\bibitem[Ritter et~al.(2018)Ritter, Botev, and Barber]{ritter2018scalable}
Ritter, H., Botev, A., and Barber, D.
\newblock A scalable {L}aplace approximation for neural networks.
\newblock In \emph{International Conference on Learning Representations}, volume~6, 2018.

\bibitem[Sajda(2006)]{sajda2006machine}
Sajda, P.
\newblock Machine learning for detection and diagnosis of disease.
\newblock \emph{Annu. Rev. Biomed. Eng.}, 8:\penalty0 537--565, 2006.

\bibitem[Salimbeni \& Deisenroth(2017)Salimbeni and Deisenroth]{salimbeni2017doubly}
Salimbeni, H. and Deisenroth, M.
\newblock Doubly stochastic variational inference for deep {G}aussian processes.
\newblock \emph{Advances in Neural Information Processing Systems}, 30:\penalty0 4588--4599, 2017.

\bibitem[Santana \& Hern{\'{a}}ndez{-}Lobato(2022)Santana and Hern{\'{a}}ndez{-}Lobato]{SantanaH22}
Santana, S.~R. and Hern{\'{a}}ndez{-}Lobato, D.
\newblock Adversarial {\(\alpha\)}-divergence minimization for {B}ayesian approximate inference.
\newblock \emph{Neurocomputing}, 471:\penalty0 260--274, 2022.

\bibitem[Scannell et~al.(2024)Scannell, Mereu, Chang, Tamir, Pajarinen, and Solin]{scannell2024function}
Scannell, A., Mereu, R., Chang, P., Tamir, E., Pajarinen, J., and Solin, A.
\newblock Function-space parameterization of neural networks for sequential learning.
\newblock \emph{International Conference on Learning Representations}, 2024.

\bibitem[Singh(2021)]{singh2021better}
Singh, P.~N.
\newblock Better application of {B}ayesian deep learning to diagnose disease.
\newblock In \emph{2021 5th International Conference on Computing Methodologies and Communication (ICCMC)}, pp.\  928--934. IEEE, 2021.

\bibitem[Titsias(2009)]{titsias2009variational}
Titsias, M.
\newblock Variational learning of inducing variables in sparse {G}aussian processes.
\newblock In \emph{Artificial Intelligence and Statistics}, pp.\  567--574, 2009.

\bibitem[Vaswani et~al.(2017)Vaswani, Shazeer, Parmar, Uszkoreit, Jones, Gomez, Kaiser, and Polosukhin]{Vaswani2017}
Vaswani, A., Shazeer, N., Parmar, N., Uszkoreit, J., Jones, L., Gomez, A.~N., Kaiser, L.~u., and Polosukhin, I.
\newblock Attention is {A}ll you need.
\newblock In \emph{Advances in Neural Information Processing Systems}, pp.\  5998--6008, 2017.

\bibitem[Villacampa{-}Calvo \& Hern{\'{a}}ndez{-}Lobato(2020)Villacampa{-}Calvo and Hern{\'{a}}ndez{-}Lobato]{Villacampa2020}
Villacampa{-}Calvo, C. and Hern{\'{a}}ndez{-}Lobato, D.
\newblock Alpha divergence minimization in multi-class {G}aussian process classification.
\newblock \emph{Neurocomputing}, 378:\penalty0 210--227, 2020.

\bibitem[Williams \& Rasmussen(2006)Williams and Rasmussen]{williams2006gaussian}
Williams, C.~K. and Rasmussen, C.~E.
\newblock \emph{Gaussian processes for machine learning}, volume~2.
\newblock MIT press Cambridge, MA, 2006.

\end{thebibliography}

\newpage

\onecolumn

\appendix

\section{Proofs}\label{app:proofs}

\textbf{Theorem 1 {\citet{cheng2016incremental}}.}
    Using a sparse GP approximation with \(q(\mathbf{f}, \mathbf{u}) = p(\mathbf{f}|\mathbf{u})q(\mathbf{u})\) is 
	equivalent to restricting the mean and covariance functions of the dual representation in the RKHS to
    \begin{equation}
        \tilde{\mu} = \Phi_{\mathbf{Z}}(\bm{a}) \quad \text{and} \quad \tilde{\Sigma} = I + \Phi_{\mathbf{Z}}\bm{A}\Phi_{\mathbf{Z}}^T\,,
    \end{equation}
    where the functional \(\Phi_{\mathbf{Z}}: \mathbb{R}^M \to \mathcal{H} \) defines a linear combination of basis functions as \(\Phi_{\mathbf{Z}}(\bm{a}) = \sum_{m=1}^{M} a_m \phi_{\mathbf{z}_m}\), with \(\bm{a} = (a_1,\dots,a_M) \in \mathbb{R}^M\) and the functional \(\Phi_{\mathbf{Z}}\bm{A}\Phi_{\mathbf{Z}}^T= \sum_{i=1}^{M}\sum_{j=1}^M\phi_{\mathbf{z}_i} A_{i,j} \phi_{\mathbf{z}_j}^T\), defines a quadratic expression where \(\bm{A} \in \mathbb{R}^{M\times M}\) such that \(\tilde{\Sigma} \geq 0\).

\begin{proof}
    First of all, notice that \(\Phi_{\mathbf{Z}} = \begin{pmatrix} \phi_{\mathbf{z}_1}, \cdots, \phi_{\mathbf{z}_M}\end{pmatrix} \in \mathcal{H}^M\), leading to 
    \begin{equation}\label{eq:K_matrix}
    K_{\mathbf{Z}} := \Phi_{\mathbf{Z}} \Phi_{\mathbf{Z}}^T = \begin{pmatrix}
        \braket{\phi_{\mathbf{z}_1}, \phi_{\mathbf{z}_1}} & \braket{\phi_{\mathbf{z}_1}, \phi_{\mathbf{z}_2}} & \cdots & \braket{\phi_{\mathbf{z}_1}, \phi_{\mathbf{z}_M}}\\
        \vdots & \vdots &  & \vdots\\
        \braket{\phi_{\mathbf{z}_M}, \phi_{\mathbf{z}_1}} & \braket{\phi_{\mathbf{z}_M}, \phi_{\mathbf{z}_2}} & \cdots & \braket{\phi_{\mathbf{z}_M}, \phi_{\mathbf{z}_M}}\\
    \end{pmatrix} \in \mathbb{R}^{M \times M}\,.
    \end{equation}  
    Furthermore, \(K_{\mathbf{x}, \mathbf{Z}}\) defined as \(\mathbf{v} \to \braket{\phi_{\mathbf{x}},\Phi_{\mathbf{Z}}(\mathbf{v})}\) can be seen as a vector of \(\mathbb{R}^{M}\), considering the image of any orthonormal basis of \(\mathbb{R}^M\). In fact, let \(\mathbf{e}_1, \dots, \mathbf{e}_M\) be the usual basis of \(\mathbb{R}^M\):
    \begin{equation}\label{eq:K_vector}
    K_{\mathbf{x}, \mathbf{Z}} \cong \begin{pmatrix}
        \braket{\phi_{\mathbf{x}},\Phi_{\mathbf{Z}}(\mathbf{e}_1)}\\
        \braket{\phi_{\mathbf{x}},\Phi_{\mathbf{Z}}(\mathbf{e}_2)}\\
        \vdots\\
        \braket{\phi_{\mathbf{x}},\Phi_{\mathbf{Z}}(\mathbf{e}_M)}
    \end{pmatrix} = \begin{pmatrix}
        \braket{\phi_{\mathbf{x}}, \sum_{i=1}^M \mathbf{e}_{1,i} \phi_{\mathbf{z}_1}}\\
        \braket{\phi_{\mathbf{x}},\sum_{i=1}^M \mathbf{e}_{2,i} \phi_{\mathbf{z}_2}}\\
        \vdots\\
        \braket{\phi_{\mathbf{x}},\sum_{i=1}^M \mathbf{e}_{M,i} \phi_{\mathbf{z}_M}}\\
    \end{pmatrix}= \begin{pmatrix}
        \braket{\phi_{\mathbf{x}},\phi_{\mathbf{z}_1}}\\
        \braket{\phi_{\mathbf{x}},\phi_{\mathbf{z}_2}}\\
        \vdots\\
        \braket{\phi_{\mathbf{x}},\phi_{\mathbf{z}_M}}\\
    \end{pmatrix} \in \mathbb{R}^M\,.
    \end{equation}
    Assume a variational distribution \(q(\mathbf{u}) = \mathcal{N}(\mathbf{u}|\tilde{\bm{m}}, \tilde{\bm{S}})\) with \(\tilde{\bm{m}} \in \mathbb{R}^M\) and \(\tilde{\bm{S}} \in \mathbb{R}^{M \times M}\). Then, defining the correspondent dual vectors as
    \begin{equation}
        \bm{a} = K_{\mathbf{Z}}^{-1}\tilde{\bm{m}} \in \mathbb{R}^M, \quad \bm{A} = K_{\mathbf{Z}}^{-1}\tilde{\bm{S}}K_{\mathbf{Z}}^{-1} - K_{\mathbf{Z}}^{-1} \in \mathbb{R}^{M \times M}\,,
    \end{equation}
    the mean and covariance functions in the dual formulation of the sparse GP \(q(f)\) are
    \begin{equation}
        m^{\star}(\mathbf{x}) = \braket{\phi_{\mathbf{x}}, \tilde{\mu}} = \braket{\phi_{\mathbf{x}}, \Phi_{\mathbf{Z}}(K_{\mathbf{Z}}^{-1}\tilde{\bm{m}}) } = K_{\mathbf{x}, \mathbf{Z}} K_{\mathbf{Z}}^{-1}\tilde{\bm{m}}\,,
    \end{equation}
    and
        \begin{align}
            K^{\star}(\mathbf{x}, \mathbf{x}') &= \braket{\phi_{\mathbf{x}},\Sigma (\phi_{\mathbf{x}'})} = \braket{\phi_{\mathbf{x}},\phi_{\mathbf{x}'} + \Phi_{\mathbf{Z}}\bm{A}\Phi_{\mathbf{Z}}^T\phi_{\mathbf{x}'} } = \braket{\phi_{\mathbf{x}},\phi_{\mathbf{x}'}} + \braket{\phi_{\mathbf{x}},\Phi_{\mathbf{Z}}\bm{A}\Phi_{\mathbf{Z}}^T\phi_{\mathbf{x}'} } \\
            &= K(\mathbf{x}, \mathbf{x}') + K_{\mathbf{x}, \mathbf{Z}} \bm{A} K_{\mathbf{x}', \mathbf{Z}}^T =  K(\mathbf{x}, \mathbf{x}') +  K_{\mathbf{x}, \mathbf{Z}}( K_{\mathbf{Z}}^{-1}\tilde{\bm{S}}K_{\mathbf{Z}}^{-1} - K_{\mathbf{Z}}^{-1})  K_{\mathbf{Z}, \mathbf{x}'}\,,
        \end{align}
	which is the same approximate GP posterior \( p(\mathbf{f}) = \int p(\mathbf{f}|\mathbf{u})q(\mathbf{u})\ d\mathbf{u}\) found in Equation ($6$) 
    of \citet{titsias2009variational}.
\end{proof}

\setcounter{proposition}{0}

\begin{proposition}
	If \(g(\cdot, \hat{\bm{\theta}}) \in \mathcal{H}\), then \(\forall \epsilon > 0\) exists a set of \(M_\alpha\) inducing points 
	\(\mathbf{Z}_{\alpha}\) and a collection of scalar values \(\bm{a} \in \mathbb{R}^{M_\alpha}\) such that the dual representation 
	of the sparse Gaussian process defined by
    \begin{equation}
        \tilde{\mu} = \Phi_{\mathbf{Z}_{\alpha}}(\bm{a}) \quad \text{and} \quad \tilde{\Sigma} = (I + \Phi_{\mathbf{Z}_{\beta}}\bm{A}\Phi_{\mathbf{Z}_{\beta}}^T)^{-1}\,,
    \end{equation}
    corresponds to a GP posterior approximation with mean and covariance functions defined as
    \begin{align}
        m^{\star}(\mathbf{x}) &= h_\epsilon(\mathbf{x})\,,\\
	    K^{\star}(\mathbf{x}, \mathbf{x}') &=K(\mathbf{x}, \mathbf{x}') -  K_{\mathbf{x}, \mathbf{Z}_\beta}(\bm{A}^{-1} + \bm{K}_\beta)^{-1} K_{\mathbf{Z}_\beta, \mathbf{x}'} \,, \nonumber
    \end{align}
	where \(\mathbf{Z}_{\beta}\) is a set of \(M_\beta\) inducing points, 
	\(\bm{A} \in \mathbb{R}^{M_\beta \times M_\beta}\), $K_{\mathbf{x}, \mathbf{Z}_\beta}$ is a vector with the covariances between $f(\mathbf{x})$ and 
	$f(\mathbf{Z}_\beta)$, and \(h_\epsilon\) verifies
	    $d_{\mathcal{H}}(g(\cdot, \hat{\bm{\theta}}), h_\epsilon) \leq \epsilon$,
	    with $d_{\mathcal{H}}(\cdot,\cdot)$ the distance in the RKHS.
\end{proposition}


\begin{proof}
	First of all, if \(g(\cdot, \hat{\bm{\theta}}) \in \mathcal{H}\), the reproducing property of the RKHS verifies that \(\forall \epsilon > 0 \) there exists \(\mathbf{Z}_{\alpha} \subset \mathcal{X}\), with $\mathcal{X}$ the input space, and \(\{\bm{a}_i\}_{i \in \mathbb{N}}\) such that \(h_\epsilon := \sum_{i=1}^{M_\alpha} a_i \phi_{\mathbf{z}_i} = \Phi_{\mathbf{Z}_{\alpha}}(\bm{a})\) verifies
\begin{equation}
	d_{\mathcal{H}}(g(\cdot, \hat{\bm{\theta}}), h_\epsilon) \leq \epsilon\,.
\end{equation}
As a result, the mean function of the approximate posterior is
\begin{equation}
	m^{\star}(\mathbf{x}) = \braket{\phi_{\mathbf{x}}, \tilde{\mu}} = \tilde{\mu}(\mathbf{x}) = h_\epsilon(\mathbf{x}) \approx g(\mathbf{x}, \hat{\bm{\theta}})\,.
 \end{equation}
 On the other hand, using that
 \begin{equation}
     (I + \Phi_{\mathbf{Z}_{\beta}}\bm{A}\Phi_{\mathbf{Z}_{\beta}}^T)^{-1} = I - \Phi_{\mathbf{Z}_{\beta}}(\bm{A}^{-1} + \Phi_{\mathbf{Z}_{\beta}}^T\Phi_{\mathbf{Z}_{\beta}})^{-1}\Phi_{\mathbf{Z}_{\beta}}^T\,,
 \end{equation}
 the covariance function is
    \begin{align}
     K^{\star}(\mathbf{x}, \mathbf{x}' ) &= \braket{\phi_{\mathbf{x}},\tilde{\Sigma} (\phi_{\mathbf{x}'})} = \braket{\phi_{\mathbf{x}},\phi_{\mathbf{x}'}} - \braket{\phi_{\mathbf{x}},\Phi_{\mathbf{Z}_{\beta}}(\bm{A}^{-1} + \Phi_{\mathbf{Z}_{\beta}}^T\Phi_{\mathbf{Z}_{\beta}})^{-1}\Phi_{\mathbf{Z}_{\beta}}^T \phi_{\mathbf{x}'}}\\
     &=  K(\mathbf{x}, \mathbf{x}') -  K_{\mathbf{x}, \mathbf{Z}_\beta}(\bm{A}^{-1} - K_{\mathbf{Z}_\beta})^{-1} K_{\mathbf{Z}_\beta, \mathbf{x}'}\,.
    \end{align}
 Where the characterization of \(K_{\mathbf{Z}_\beta}\) and \(K_{\mathbf{x},\mathbf{Z}_\beta}\) as elements of \(\mathbb{R}^{M_\beta \times M_\beta}\) and \(\mathbb{R}^{M_\beta}\) of \eqref{eq:K_matrix} and \eqref{eq:K_vector} are used.
\end{proof}

\begin{proposition}
The value of \(\bm{A}\) in Proposition~\ref{prop:decoupled} that minimizes~\eqref{eq:opt_dual} is 
\begin{align}
	\bm{A} &= \frac{1}{\sigma^2} \bm{K}_{\beta}^{-1} \bm{K}_{\bm{Z}_\beta, \bm{X}}\bm{K}_{\bm{X}, \bm{Z}_\beta} \bm{K}_{\beta}^{-1}\,,
\end{align}
where $\sigma^2$ is the noise variance and $\bm{K}_{\bm{X}, \bm{Z}_\beta}$ is a matrix with the prior covariances
between $f(\mathbf{X})$ and $f(\mathbf{Z}_\beta)$.  If $\bm{Z}_\beta=\mathbf{X}$, the covariance function of 
the predictive distribution in (\ref{eq:pred_valla}) is equal to that of the full GP.
\end{proposition} 

\begin{proof}
    For simplicity, assume a non-inverse reparameterization where
    \begin{equation}
        q(f) = \mathcal{N}\left(\, f \, \big|\, \Phi_{\bm{Z}_\alpha}(\bm{a}) ,\, I + \Phi_{\bm{Z}_\beta}\hat{\bm{A}}\Phi_{\bm{Z}_\beta}\right)\,.
    \end{equation}
    We will find the optimal value for \(\hat{\bm{A}}\) and compute the corresponding value for \(\bm{A}\) (using the inverse reparameterization). First, we will show that the true GP posterior can be written in the dual formulation as the following
    \begin{equation}
        p(f|y) = \mathcal{N}\left(f| (\Phi_{\bm{X}}\Phi_{\bm{X}}^T + \sigma^2\bm{I})^{-1}\Phi_{\bm{X}}y ),\, \sigma^2(\Phi_{\bm{X}}\Phi_{\bm{X}}^T + \sigma^2\bm{I})^{-1}\right)\,,
    \end{equation}
    where it verifies that the GP posterior mean function is
        \begin{align}
            m(\mathbf{x}) &= \braket{\phi_{\mathbf{x}}, (\Phi_{\bm{X}}\Phi_{\bm{X}}^T + \sigma^2\bm{I})^{-1}\Phi_{\bm{X}}y} = (\Phi_{\bm{X}}\Phi_{\bm{X}}^T + \sigma^2\bm{I})^{-1}\Phi_{\bm{X}}\phi_{\mathbf{x}}y \\
            &= (K_{\mathbf{X}, \mathbf{X}} + \sigma^2 \bm{I})^{-1} K_{\mathbf{X}, \mathbf{x}} y = K_{\mathbf{x}, \mathbf{X}}(K_{\mathbf{X}, \mathbf{X}} + \sigma^2 \bm{I})^{-1}  y\,.
        \end{align}
     Where the characterization of \(K_{\mathbf{X}, \mathbf{X}}\) and \(K_{\mathbf{X}, \mathbf{x}}\) as elements of \(\mathbb{R}^{N \times N}\) and \(\mathbb{R}^{N}\) is used, similarly to Equations \eqref{eq:K_matrix} and \eqref{eq:K_vector}. Furthermore, using Woodbury matrix identity,
     \begin{equation}
         \sigma^2(\Phi_{\bm{X}}\Phi_{\bm{X}}^T + \sigma^2\bm{I})^{-1} = \bm{I} -\Phi_{\bm{X}}(K_{\mathbf{X}, \mathbf{X}} + \sigma^{2}\bm{I})^{-1}\Phi_{\bm{X}}^T\,,
     \end{equation}
     where again, correspondence between operators and matrices is used. This leads to the covariance function:
     \begin{align}
      K(\mathbf{x}, \mathbf{x}' ) &= \braket{\phi_{\mathbf{x}},\sigma^2(\Phi_{\bm{X}}\Phi_{\bm{X}}^T + \sigma^2\bm{I})^{-1} (\phi_{\mathbf{x}'})} = \braket{\phi_{\mathbf{x}},\bm{I} -\Phi_{\bm{X}}(K_{\mathbf{X}, \mathbf{X}} + \sigma^{2}\bm{I})^{-1}\Phi_{\bm{X}}^T (\phi_{\mathbf{x}'})} \\
      &= K_{\mathbf{x}, \mathbf{x}'} - K_{\mathbf{x}, \mathbf{X}} (K_{\mathbf{X}, \mathbf{X}} + \sigma^{2}\bm{I})^{-1} K_{\mathbf{X}, \mathbf{x}'}\,.
     \end{align}
     This mean and covariance function are exactly the ones obtained from the original GP formulation \citep{titsias2009variational}. Using the ELBO is equivalent to the KL divergence between the true posterior and the variational approximation, we got
    \begin{equation}
        \text{KL}(q(f)\mid p(f|y)) \propto  \frac{1}{2} \text{tr}\Big(\bm{B}^{-1} (\bm{I} + \Phi_{\bm{Z}_\beta} \hat{\bm{A}} \Phi_{\bm{Z}_\beta}^T)\Big) - \frac{1}{2}\ln\Big(\Big|\bm{I} + \Phi_{\bm{Z}_\beta} \hat{\bm{A}} \Phi_{\bm{Z}_\beta}^T\Big|\Big)\,,
    \end{equation}
    where
    \begin{equation}
        \bm{B} = \sigma^2(\Phi_{\bm{X}}\Phi_{\bm{X}}^T + \sigma^2\bm{I})^{-1}\,.
    \end{equation}
    Naming \(\bm{M} = \mathbf{I} + \Phi_{\bm{Z}_\beta} \hat{\bm{A}} \Phi_{\bm{Z}_\beta}^T \), it is important to notice that given \(\Phi_{\bm{Z}_\beta} \hat{\bm{A}} \Phi_{\bm{Z}_\beta}^T= \sum_{i=1}^{M}\sum_{j=1}^M\phi_{\mathbf{z}_i} \hat{a}_{i,j} \phi_{\mathbf{z}_j}^T\), despite being an operator, \(\bm{M}\) can be seen as a matrix whose entries are the application of \(\mathbf{I} +  \Phi_{\bm{Z}_\beta} (\cdot) \Phi_{\bm{Z}_\beta}^T\) to the usual basis of matrices. In short
    \begin{equation}
	    \bm{M} = \mathbf{I} +
        \begin{pmatrix}
            \phi_{\bm{z}_1}  \hat{a}_{1,1}\phi_{\bm{z}_1}^T & \cdots & \phi_{\bm{z}_1}  \hat{a}_{1,M}\phi_{\bm{z}_M}^T \\
            \vdots & & \vdots\\
            \phi_{\bm{z}_M}  \hat{a}_{M,1}\phi_{\bm{z}_1}^T & \cdots & \phi_{\bm{z}_M}  \hat{a}_{M,M}\phi_{\bm{z}_M}^T \\
        \end{pmatrix} \in \mathbb{R}^{M_{\beta} \times M_{\beta}}\,.
    \end{equation}
    
    The partial derivative of \(\bm{M}\) w.r.t. a single position in the matrix \(\hat{\bm{A}}\) is \(\frac{ \partial \bm{M}}{\partial \hat{a}_{ij}} = (\Phi_{\bm{Z}_\beta} \delta_i) ( \Phi_{\bm{Z}_\beta}\delta_j)^T\), where \(\delta_j\) denotes a zero-vector with a single \(1\) at position \(j\). Then, using the chain rule for matrices,
    \begin{equation}
        \frac{\partial g(U)}{\partial X_{ij}} = \text{tr}\left( \frac{\partial g(U)}{\partial U}^T \frac{\partial U}{\partial X_{ij}} \right),
    \end{equation}
    we can compute the optimum in the two terms in the KL. The optimum for the logarithm term can be computed as
    \begin{align}
        \frac{ \partial \ln(|\bm{M}|)}{\partial \hat{a}_{ij}} &= \text{tr}\left(\frac{ \partial \ln(|\bm{M}|)}{\partial \bm{M}}^T\frac{ \partial \bm{M}}{\partial \hat{a}_{ij}}\right)\\
        &= \text{tr}(\bm{M}^{-1^T} (\Phi_{\bm{Z}_\beta} \delta_i) ( \Phi_{\bm{Z}_\beta}\delta_j)^T) \\
        &=  \text{tr}(( \Phi_{\bm{Z}_\beta}\delta_j)^T \bm{M}^{-1} (\Phi_{\bm{Z}_\beta} \delta_i) ) \\
        &=  (\Phi_{\bm{Z}_\beta}^T \bm{M}^{-1}\Phi_{\bm{Z}_\beta})\delta_{ji}\,,
    \end{align}
    leading to 
    \begin{equation}
         \frac{ \partial \ln(|\bm{M}|)}{\partial \hat{\bm{A}}} =  \Phi_{\bm{Z}_\beta}^T \bm{M}^{-1}\Phi_{\bm{Z}_\beta} = \Phi_{\bm{Z}_\beta}^T ( \bm{I} + \Phi_{\bm{Z}_\beta} \hat{\bm{A}} \Phi_{\bm{Z}_\beta}^T)^{-1}\Phi_{\bm{Z}_\beta}\,.
    \end{equation}
    On the other hand, the optimum for the trace term is    
        \begin{align}
        \frac{\partial \text{tr}(\bm{B}^{-1}\bm{M})}{\partial \hat{a}_{ij}} &= \text{tr}\left(\frac{ \partial \text{tr}(\bm{B}^{-1}\bm{M})}{\partial \bm{M}}^T\frac{ \partial \bm{M}}{\partial \hat{a}_{ij}}\right)\\
        &= \text{tr}\left(\bm{B}^{-1} (\Phi_{\bm{Z}_\beta} \delta_i) ( \Phi_{\bm{Z}_\beta}\delta_j)^T\right)\\
        &= \text{tr}(( \Phi_{\bm{Z}_\beta}\delta_j)^T \bm{B}^{-1^T} (\Phi_{\bm{Z}_\beta} \delta_i) ) \\
        &= (\Phi_{\bm{Z}_\beta}^T \bm{B}^{-1}\Phi_{\bm{Z}_\beta})\delta_{ji}\,,
        \end{align}
    where we used that \(\bm{B} = \bm{B}^T\). As a result,
    \begin{align}
         \frac{\partial \text{tr}(\bm{B}^{-1}\bm{M})}{\partial \hat{\bm{A}}} &=  \Phi_{\bm{Z}_\beta}^T \bm{B}^{-1}\Phi_{\bm{Z}_\beta}\, \\ 
         & =  \sigma^2 \Phi_{\bm{Z}_\beta}^T (\Phi_{\bm{X}}\Phi_{\bm{X}}^T + \sigma^2\bm{I})^{-1}\ \Phi_{\bm{Z}_\beta}\,.
    \end{align}
    Using all the derivations
    \begin{equation}
        \frac{\partial \text{KL}(q(f)\mid p(f|y)) }{\partial \hat{\bm{A}}} = 0 \iff \Phi_{\bm{Z}_\beta}^T (\bm{B}^{-1}- ( \bm{I} + \Phi_{\bm{Z}_\beta} \hat{\bm{A}} \Phi_{\bm{Z}_\beta}^T)^{-1})\Phi_{\bm{Z}_\beta} = 0\,.
    \end{equation}
    Using the Woodbury matrix identity 
        \begin{align}
            0 &= \Phi_{\bm{Z}_\beta}^T (\bm{B}^{-1}- ( \bm{I} + \Phi_{\bm{Z}_\beta} \hat{\bm{A}} \Phi_{\bm{Z}_\beta}^T)^{-1})\Phi_{\bm{Z}_\beta}\\
            &= \Phi_{\bm{Z}_\beta}^T (\bm{B}^{-1}- \bm{I} + \Phi_{\bm{Z}_\beta} (\hat{\bm{A}}^{-1} + \bm{K}_{\beta})^{-1} \Phi_{\bm{Z}_\beta}^T )\Phi_{\bm{Z}_\beta}\\
            &= \Phi_{\bm{Z}_\beta}^T \bm{B}^{-1} \Phi_{\bm{Z}_\beta} - \bm{K}_{\beta} + \bm{K}_{\beta} (\hat{\bm{A}}^{-1} + \bm{K}_{\beta})^{-1} \bm{K}_{\beta}\,.
        \end{align}
    
    Thus, the value of \(\hat{\bm{A}}\) where ${\partial \text{KL}} / {\partial \hat{\bm{A}}} = 0$ verifies
    \begin{equation}
        \hat{\bm{A}} = ((\bm{K}_{\beta}^{-1} - \bm{K}_{\beta}^{-1} \Phi_{\bm{Z}_\beta}^T \bm{B}^{-1} \Phi_{\bm{Z}_\beta} \bm{K}_{\beta}^{-1})^{-1} - \bm{K}_{\beta})^{-1}\,.
    \end{equation}
    Using that \(\bm{B}^{-1} = \sigma^{-2}(\Phi_{\bm{X}}\Phi_{\bm{X}}^T + \sigma^2\bm{I})\), we can take further derivations on the expression of \(\hat{\bm{A}}\) as
        \begin{align}
        \hat{\bm{A}} &= ((\bm{K}_{\beta}^{-1} - \bm{K}_{\beta}^{-1} \Phi_{\bm{Z}_\beta}^T \bm{B}^{-1} \Phi_{\bm{Z}_\beta} \bm{K}_{\beta}^{-1})^{-1} - \bm{K}_{\beta})^{-1} \\ 
        &= ((\bm{K}_{\beta}^{-1} - \bm{K}_{\beta}^{-1} \Phi_{\bm{Z}_\beta}^T  \sigma^{-2}(\Phi_{\bm{X}}\Phi_{\bm{X}}^T + \sigma^2\bm{I}) \Phi_{\bm{Z}_\beta} \bm{K}_{\beta}^{-1})^{-1} - \bm{K}_{\beta})^{-1}\\
        &= ((\bm{K}_{\beta}^{-1} - \sigma^{-2} \bm{K}_{\beta}^{-1}\bm{K}_{\bm{Z}_\beta, \bm{X}}\bm{K}_{\bm{X}, \bm{Z}_\beta}\bm{K}_{\beta}^{-1} - \bm{K}_{\beta}^{-1})^{-1} - \bm{K}_{\beta})^{-1}\\
        &= ((- \sigma^{-2} \bm{K}_{\beta}^{-1}\bm{K}_{\bm{Z}_\beta, \bm{X}}\bm{K}_{\bm{X}, \bm{Z}_\beta}\bm{K}_{\beta}^{-1})^{-1} - \bm{K}_{\beta})^{-1}\\
        &= -((\sigma^{-2} \bm{K}_{\beta}^{-1}\bm{K}_{\bm{Z}_\beta, \bm{X}}\bm{K}_{\bm{X}, \bm{Z}_\beta}\bm{K}_{\beta}^{-1})^{-1} + \bm{K}_{\beta})^{-1}\,.
        \end{align}
    Applying again Woodbury matrix identity:
        \begin{align}
        \hat{\bm{A}} &= -((\sigma^{-2} \bm{K}_{\beta}^{-1}\bm{K}_{\bm{Z}_\beta, \bm{X}}\bm{K}_{\bm{X}, \bm{Z}_\beta}\bm{K}_{\beta}^{-1})^{-1} + \bm{K}_{\beta})^{-1}\\
        &= -(\bm{K}_{\beta}^{-1} - \bm{K}_{\beta}^{-1}(\sigma^{-2} \bm{K}_{\beta}^{-1}\bm{K}_{\bm{Z}_\beta, \bm{X}}\bm{K}_{\bm{X}, \bm{Z}_\beta}  \bm{K}_{\beta}^{-1} + \bm{K}_{\beta}^{-1} )^{-1}  \bm{K}_{\beta}^{-1})\\
        &=  -(\bm{K}_{\beta}^{-1} - (\sigma^{-2}\bm{K}_{\bm{Z}_\beta, \bm{X}}\bm{K}_{\bm{X}, \bm{Z}_\beta}  + \bm{K}_{\beta} )^{-1})  \\
        &=  -\bm{K}_{\beta}^{-1} + (\sigma^{-2}\bm{K}_{\bm{Z}_\beta, \bm{X}}\bm{K}_{\bm{X}, \bm{Z}_\beta}  + \bm{K}_{\beta} )^{-1}\,.
        \end{align}
    If we substitute this value on the predictive distribution
    \begin{align}
        K^{\star}(\mathbf{x}, \mathbf{x}') &= \braket{\phi_{\mathbf{x}},\bm{M} (\phi_{\mathbf{x}'})} =  K(\mathbf{x}, \mathbf{x}') +  K_{\mathbf{x}, \bm{Z}_\beta} \hat{\bm{A}}  K_{\bm{Z}_\beta, \mathbf{x}'}\\
        &=  K(\mathbf{x}, \mathbf{x}') +  K_{\mathbf{x}, \mathbf{Z}} ( -\bm{K}_{\beta}^{-1} + (\sigma^{-2}\bm{K}_{\bm{Z}_\beta, \bm{X}}\bm{K}_{\bm{X}, \bm{Z}_\beta}  + \bm{K}_{\beta} )^{-1})  K_{\bm{Z}_\beta, \mathbf{x}'}\\
        &= K(\mathbf{x}, \mathbf{x}') -  K_{\mathbf{x}, \bm{Z}_\beta} \bm{K}_{\beta}^{-1} K_{\bm{Z}_\beta, \mathbf{x}'} + K_{\mathbf{x}, \bm{Z}_\beta} (\sigma^{-2}\bm{K}_{\bm{Z}_\beta, \bm{X}}\bm{K}_{\bm{X}, \bm{Z}_\beta}  + \bm{K}_{\beta} )^{-1})  K_{\bm{Z}_\beta, \mathbf{x}'}\\
        &= K(\mathbf{x}, \mathbf{x}') -  K_{\mathbf{x}, \bm{Z}_\beta} \bm{K}_{\beta}^{-1} K_{\bm{Z}_\beta, \mathbf{x}'}\\
        &\quad+ K_{\mathbf{x}, \bm{Z}_\beta} \bm{K}_{\beta}^{-1} (\bm{K}_{\beta} (\sigma^{-2}\bm{K}_{\bm{Z}_\beta, \bm{X}}\bm{K}_{\bm{X}, \bm{Z}_\beta}  + \bm{K}_{\beta} )^{-1} \bm{K}_{\beta}) \bm{K}_{\beta}^{-1}  K_{\bm{Z}_\beta, \mathbf{x}'}\,.
     \end{align}
     This expression coincides with the optimal sparse GP solution described by \citet{titsias2009variational}, 
     with optimal variational covariance \((\bm{K}_{\beta} (\sigma^{-2}\bm{K}_{\bm{Z}_\beta, \bm{X}}\bm{K}_{\bm{X}, \bm{Z}_\beta}  + \bm{K}_{\beta} )^{-1} \bm{K}_{\beta})\). As a result, the optimal solution for VaLLA coincides with the optimal solution for standard sparse GPs. Let us now compute the optimal value of \(\bm{A}\) given the optimal value of \(\hat{\bm{A}}\). First, notice that using Woodbury Matrix identity
     \begin{equation}
        (\bm{I} + \Phi_{\bm{Z}_\beta} \bm{A} \Phi_{\bm{Z}_\beta}^T)^{-1} = \bm{I} - \Phi_{\bm{Z}_\beta} (\bm{A} +  \bm{K}_{\beta})^{-1} \Phi_{\bm{Z}_\beta}^T \,.
    \end{equation}
    Therefore, the relation between \(\bm{A}\) and \(\hat{\bm{A}}\) is \(\hat{\bm{A}} = -(\bm{A} +  \bm{K}_{\beta})^{-1}\). Meaning that
    \begin{equation}
    \label{eq:optimal_A}
        \hat{\bm{A}} = -\bm{K}_{\beta}^{-1} + (\sigma^{-2}\bm{K}_{\bm{Z}_\beta, \bm{X}}\bm{K}_{\bm{X}, \bm{Z}_\beta}  + \bm{K}_{\beta} )^{-1} \implies \bm{A} = \frac{1}{\sigma^2} \bm{K}_{\beta}^{-1} \bm{K}_{\bm{Z}_\beta, \bm{X}}\bm{K}_{\bm{X}, \bm{Z}_\beta} \bm{K}_{\beta}^{-1}\,.
    \end{equation}

    \paragraph{Global optimum}
    
    To complete the proof of the solution in Eq. (\ref{eq:optimal_A}) 
    being not only optimal but also a maximum of the ELBO, we must test the behavior of the second derivative w.r.t. $\bm{A}$. Let us reuse the previous results, where we found that 
    \begin{equation}
        \frac{\partial \text{KL}(q(f)\mid p(f|y)) }{\partial \hat{a}_{i,j}} = \frac{1}{2} \delta_j^T (\Phi_{\bm{Z}_\beta}^T \bm{B}^{-1} \Phi_{\bm{Z}_\beta} - \Phi_{\bm{Z}_\beta}^T \bm{M}^{-1}\Phi_{\bm{Z}_\beta})\delta_i\,.
    \end{equation}
    Taking a second derivative w.r.t. another location \(\hat{a}_{u,v}\) yields
    \begin{equation}
        \frac{\partial}{\partial \hat{a}_{u,v}}\frac{\partial \text{KL}(q(f)\mid p(f|y)) }{\partial \hat{a}_{i,j}} = \frac{1}{2}\frac{\partial}{\partial \hat{a}_{u,v}} \delta_j^T (\Phi_{\bm{Z}_\beta}^T \bm{B}^{-1} \Phi_{\bm{Z}_\beta} - \Phi_{\bm{Z}_\beta}^T \bm{M}^{-1}\Phi_{\bm{Z}_\beta})\delta_i\,.
    \end{equation}
    Considering that \(\bm{B}\) does not depend on \(\hat{\bm{A}}\), the first term drops from the derivative. Thus
    \begin{equation}
        \frac{\partial}{\partial \hat{a}_{u,v}}\frac{\partial \text{KL}(q(f)\mid p(f|y)) }{\partial \hat{a}_{i,j}} = -\frac{1}{2}\frac{\partial}{\partial \hat{a}_{u,v}} \delta_j^T (\Phi_{\bm{Z}_\beta}^T \bm{M}^{-1}\Phi_{\bm{Z}_\beta})\delta_i\,.
    \end{equation}
    Here we aim to use the chain rule for matrices,
    \begin{equation}
        \frac{\partial g(U)}{\partial X_{ij}} = \text{tr}\left( \frac{\partial g(U)}{\partial U}^T \frac{\partial U}{\partial X_{ij}} \right)\,,
    \end{equation}
    Then, consider that
    \begin{equation}
        \frac{\partial\  \delta_j^T (\Phi_{\bm{Z}_\beta}^T \bm{M}^{-1}\Phi_{\bm{Z}_\beta})\delta_i}{\partial \bm{M}} =  -\bm{M}^{-1}\Phi_{\bm{Z}_\beta} \delta_j\delta_i^T \Phi_{\bm{Z}_\beta}^T \bm{M}^{-1}\,.
    \end{equation}
    Then,
    \begin{equation}
        \frac{\partial}{\partial \hat{a}_{u,v}} \delta_j^T (\Phi_{\bm{Z}_\beta}^T \bm{M}^{-1}\Phi_{\bm{Z}_\beta})\delta_i = \text{tr}\Bigg( \Big(-\bm{M}^{-1}\Phi_{\bm{Z}_\beta} \delta_j\delta_i^T \Phi_{\bm{Z}_\beta}^T \bm{M}^{-1}\Big)^T  \Phi_{\bm{Z}_\beta} \delta_u ( \Phi_{\bm{Z}_\beta}\delta_v)^T \Bigg)\,.
    \end{equation}
    We can work on this trace to simplify the expression as
    \begin{align}
        \frac{\partial}{\partial \hat{a}_{u,v}} \delta_j^T (\Phi_{\bm{Z}_\beta}^T \bm{M}^{-1}\Phi_{\bm{Z}_\beta})\delta_i &= \text{tr}\Bigg( \Big(-\bm{M}^{-1}\Phi_{\bm{Z}_\beta} \delta_j\delta_i^T \Phi_{\bm{Z}_\beta}^T \bm{M}^{-1}\Big)^T  \Phi_{\bm{Z}_\beta} \delta_u ( \Phi_{\bm{Z}_\beta}\delta_v)^T \Bigg)\\
        &= -\text{tr}\Bigg( \delta_v^T\Phi_{\bm{Z}_\beta}^T \Big(\bm{M}^{-1}\Phi_{\bm{Z}_\beta} \delta_j\delta_i^T \Phi_{\bm{Z}_\beta}^T \bm{M}^{-1}\Big)^T  \Phi_{\bm{Z}_\beta} \delta_u \Bigg)\\
        &= -\delta_v^T\Phi_{\bm{Z}_\beta}^T\Big(\bm{M}^{-1}\Phi_{\bm{Z}_\beta} \delta_j\delta_i^T \Phi_{\bm{Z}_\beta}^T \bm{M}^{-1}\Big)^T  \Phi_{\bm{Z}_\beta} \delta_u \\
        &= -\delta_v^T\Phi_{\bm{Z}_\beta}^T \bm{M}^{-1}\Phi_{\bm{Z}_\beta} \delta_j\delta_i^T \Phi_{\bm{Z}_\beta}^T \bm{M}^{-1}  \Phi_{\bm{Z}_\beta} \delta_u\,. \\
    \end{align}
    Naming \(\bm{Q} = \Phi_{\bm{Z}_\beta}^T \bm{M}^{-1}  \Phi_{\bm{Z}_\beta}\), we got
    \begin{equation}
        \frac{\partial}{\partial \hat{a}_{u,v}}\frac{\partial \text{KL}(q(f)\mid p(f|y)) }{\partial \hat{a}_{i,j}} = \frac{1}{2} \bm{Q}_{v,j} \cdot \bm{Q}_{i,u} = \frac{1}{2}\bm{Q}_{j, v} \cdot \bm{Q}_{i,u}\,,
    \end{equation}
    where in the last equality we used that \(\bm{Q}\) is symmetric. Using the definition of \(\bm{M}\) we know that
    \begin{equation}
        \bm{M}^{-1} = \bm{I} - \Phi_{\bm{Z}_\beta} (\bm{A} +  \bm{K}_{\beta})^{-1} \Phi_{\bm{Z}_\beta}^T \implies \bm{Q} = \bm{K}_{\beta} - \bm{K}_{\beta}(\bm{A} + \bm{K}_{\beta})^{-1}\bm{K}_{\beta}\,.
    \end{equation}
    This shows that \(\bm{Q}\) is the posterior covariance of a GP with prior covariances \(\bm{K}_{\beta}\) and noise covariances \(\bm{A}\).
    \begin{equation}
        \frac{\partial}{\partial \hat{\bm{A}}}\frac{\partial \text{KL}(q(f)\mid p(f|y)) }{\partial \hat{\bm{A}}} = \frac{1}{2} \bm{Q} \otimes \bm{Q}\,,
    \end{equation}
    with \(\otimes\) denoting the Kronecker product and \(\bm{Q}\) a definite positive matrix. As the Kronecker product of two definite positive matrices is definite positive, the optimal is a minimum of the KL. Given that this second derivative is positive-definite, the value found is the global minimum of the ELBO objective.

\end{proof}

\section{Pseudocode}

\begin{algorithm}[t]
\caption{VaLLA's training loop with \(\alpha = 1\)}\label{alg:cap}
\begin{algorithmic}
\REQUIRE Pre-trained MAP solution $f$, input batch $\mathbf{X}, \mathbf{y}$, \(m\) number of inducing points and \(T\) iterations.
\STATE $\mathbf{Z} \gets kmeans(\mathbf{X}, m)$ \hfill \COMMENT{Initialize inducing points}
\STATE $\bm{L} \gets \bm{I}$ \hfill \COMMENT{Initialize Cholesky decomposition of \(\bm{A}\)}
\FOR{$i \in \{0, \dots, T-1\}$}
\STATE $(\mathbf{X}_b, \mathbf{y}_b) \gets get\_batch()$ \hfill \COMMENT{Get mini-batch of data}
\STATE 
\STATE $\bm{J_x} \gets compute\_jacobian(\mathbf{X}_b)$ \hfill \COMMENT{Compute Jacobians}
\STATE $\bm{J_z} \gets compute\_jacobian(\mathbf{Z})$ 
\STATE
\STATE $\bm{K_x} \gets \sigma_0^2 \bm{J_x}\bm{J_x}^T$ \hfill \COMMENT{Compute Kernels}
\STATE $\bm{K_{xz}} \gets \sigma_0^2 \bm{J_x}\bm{J_z}^T$ 
\STATE $\bm{K_z} \gets \sigma_0^2 \bm{J_z}\bm{J_z}^T$ 
\STATE $\bm{A} \gets \bm{LL}^T$\hfill \COMMENT{Compute Variational Matrix}
\STATE 
\STATE $Q\_mean \gets f(\mathbf{X}_b)$\hfill \COMMENT{Compute posterior mean}
\STATE $Q\_var \gets \bm{K_x} - \bm{K_{xz}}(\bm{A}^{-1} + \bm{K_{z})^{-1}\bm{K_{xz}}^T}$\hfill \COMMENT{Compute posterior covariance matrix}
\STATE 
\STATE $KL \gets compute\_KL(\bm{A}, \bm{K_z})$\hfill \COMMENT{Compute Kullback-Leibler divergence}
\STATE $NLL \gets compute\_NLL(\mathbf{y}_n, Q\_mean, Q\_var)$\hfill \COMMENT{Compute Negative Log-likelihood}
\STATE $loss \gets -\frac{len(\mathbf{X})}{len(\mathbf{X}_b)}NLL + KL$
\STATE Optimize parameters by minimizing $loss$.
\ENDFOR
\end{algorithmic}
\end{algorithm}

Algorithm~\ref{alg:cap} shows the structure of VaLLA's training loop, where no Early-Stopping is considered. Using the kernels and \(\bm{A}\) it is easy to compute the KL in Eq.~\ref{eq:opt_kl}. As a result, the training loop is easy to implement, since \(q(f)\), given by \texttt{Q\_mean} and \texttt{Q\_var} are easily computable as detailed in the algorithm. 

\section{MAP solution in Hilbert Space}\label{app:map}

Whether the map solution is in the Hilbert space might be difficult (if not impossible) to know. However, there are cases where it can be theoretically shown; for example in a linear model. Let the map solution be a linear model as
\begin{equation}
    g(\mathbf{x}, (\bm{w},b)) = \bm{w}^T\mathbf{x} + b\,.
\end{equation}
Then, the features (Jacobians) are
\begin{equation}
    \phi(\mathbf{x}) = (\mathbf{x}^T, 1)^T\,.
\end{equation}
With a single inducing point \(\mathbf{z}\) and scalar value \(a \in \mathbb{R}\), the mean function would be
\begin{equation}
    m(\mathbf{x}) = a \phi(\mathbf{x})^T\phi(\mathbf{z}) = a (\mathbf{x}^T\mathbf{z} + 1)\,.
\end{equation}
This recovers the MAP solution if \(a = b\) and \(\mathbf{z} = \bm{w}/a\).

However, if the model is not linear but has a linear last layer as
\begin{equation}
    g(\mathbf{x}, (\bm{w},b ,\bm{\theta})) = \bm{w}^T h_{\bm \theta}(\mathbf{x}) + b\,,
\end{equation}
where \(h_{\bm \theta}\) is a non-linear function that depends on parameters \(\bm \theta\). Then, the features (Jacobians) are
\begin{equation}
    \phi(\mathbf{x}) = \Big(h_{\bm \theta}(\mathbf{x})^T, 1, (\nabla_{\bm \theta} h_{\bm \theta}(\mathbf{x}))^T\bm{w} \Big)^T\,.
\end{equation}
Here it might be difficult to check if there exists a combination that yields the map solution as the mean function.

\section{Over-fitting and Early Stopping}\label{app:early}

\begin{figure*}[t]
	\begin{center}
	\begin{tabular}{ccc}
    {\scriptsize VaLLA Val \(M=5\)} & {\scriptsize VaLLA No-Val \(M=5\)} & {\scriptsize LLA}\\
	\includegraphics[width=0.30\textwidth]{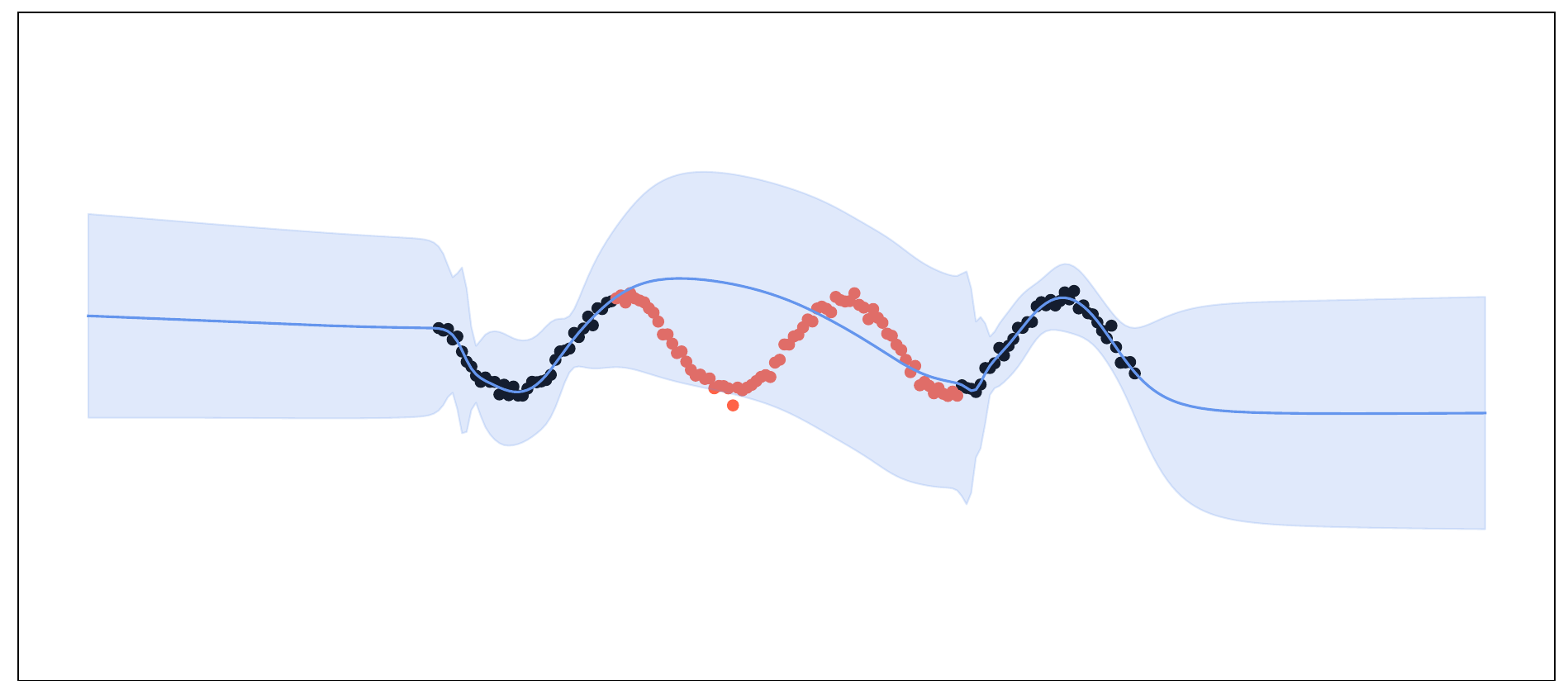} & \includegraphics[width=0.30\textwidth]{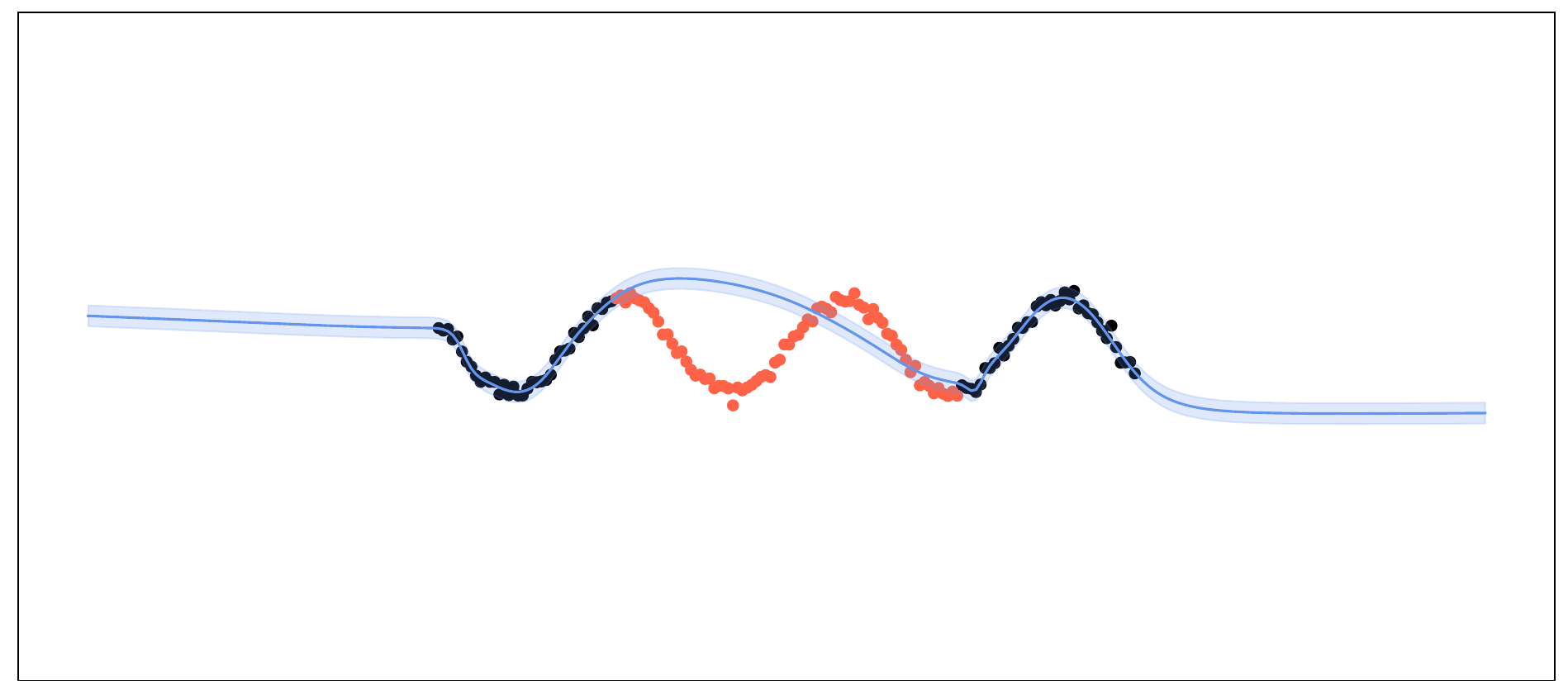} & \includegraphics[width=0.30\textwidth]{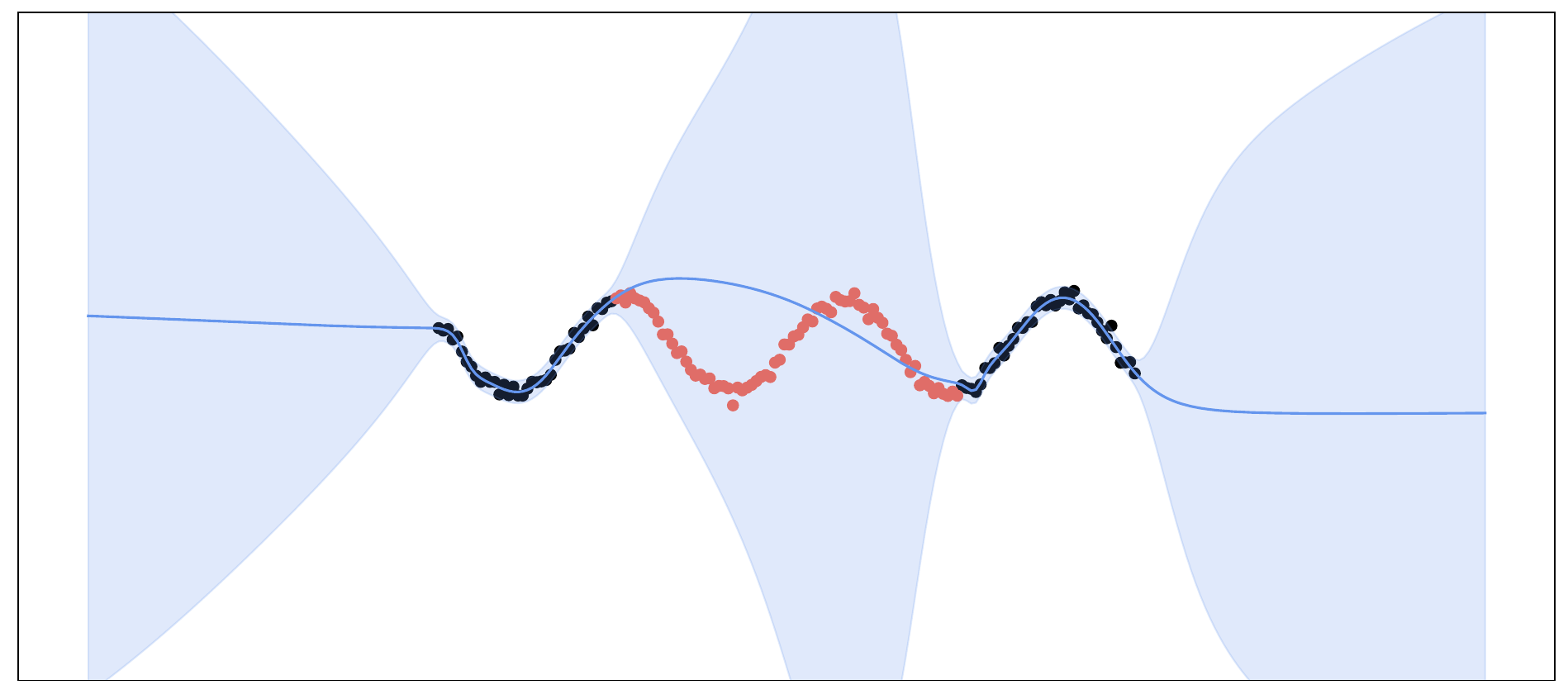}\\
    {\scriptsize VaLLA Val \(M=10\)} & {\scriptsize VaLLA No-Val \(M=10\)} & {\scriptsize ELLA \(M=10\)}\\
	\includegraphics[width=0.30\textwidth]{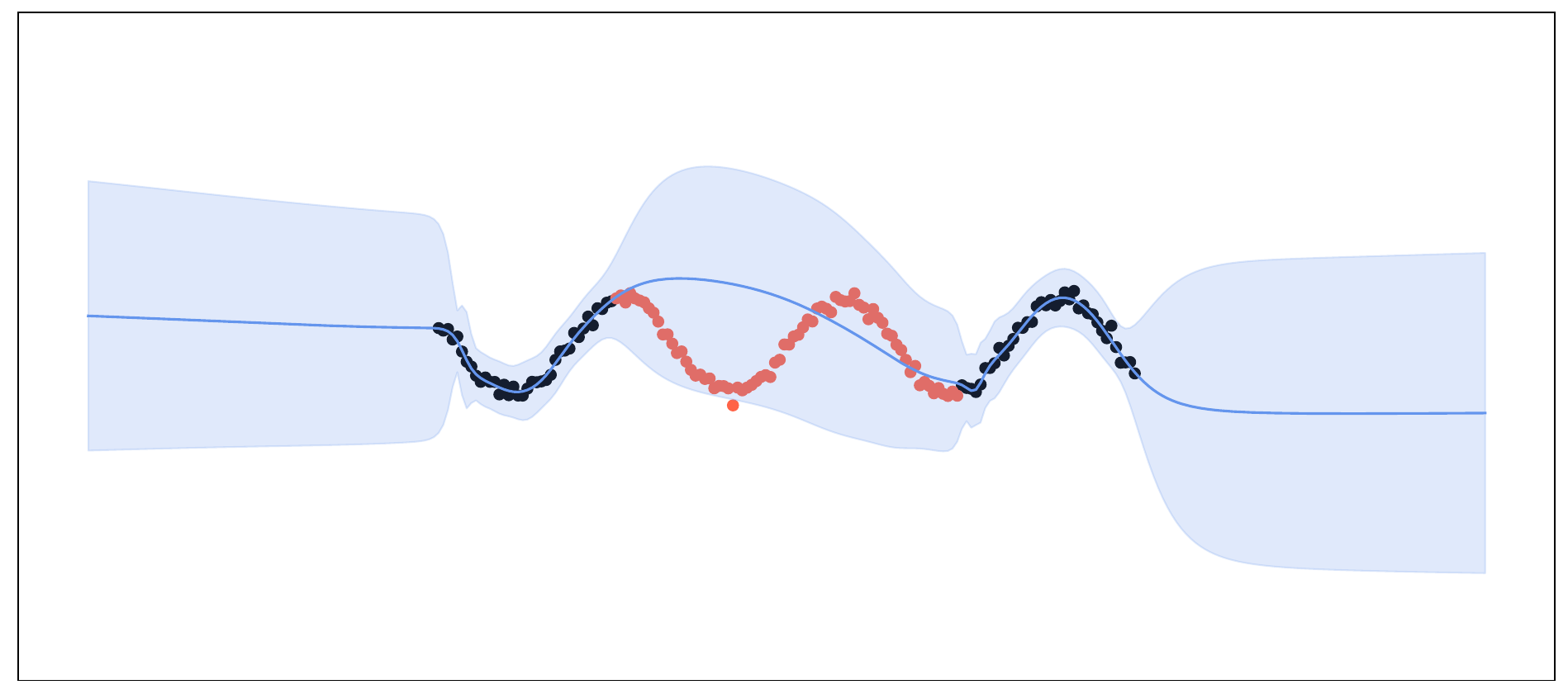} & \includegraphics[width=0.30\textwidth]{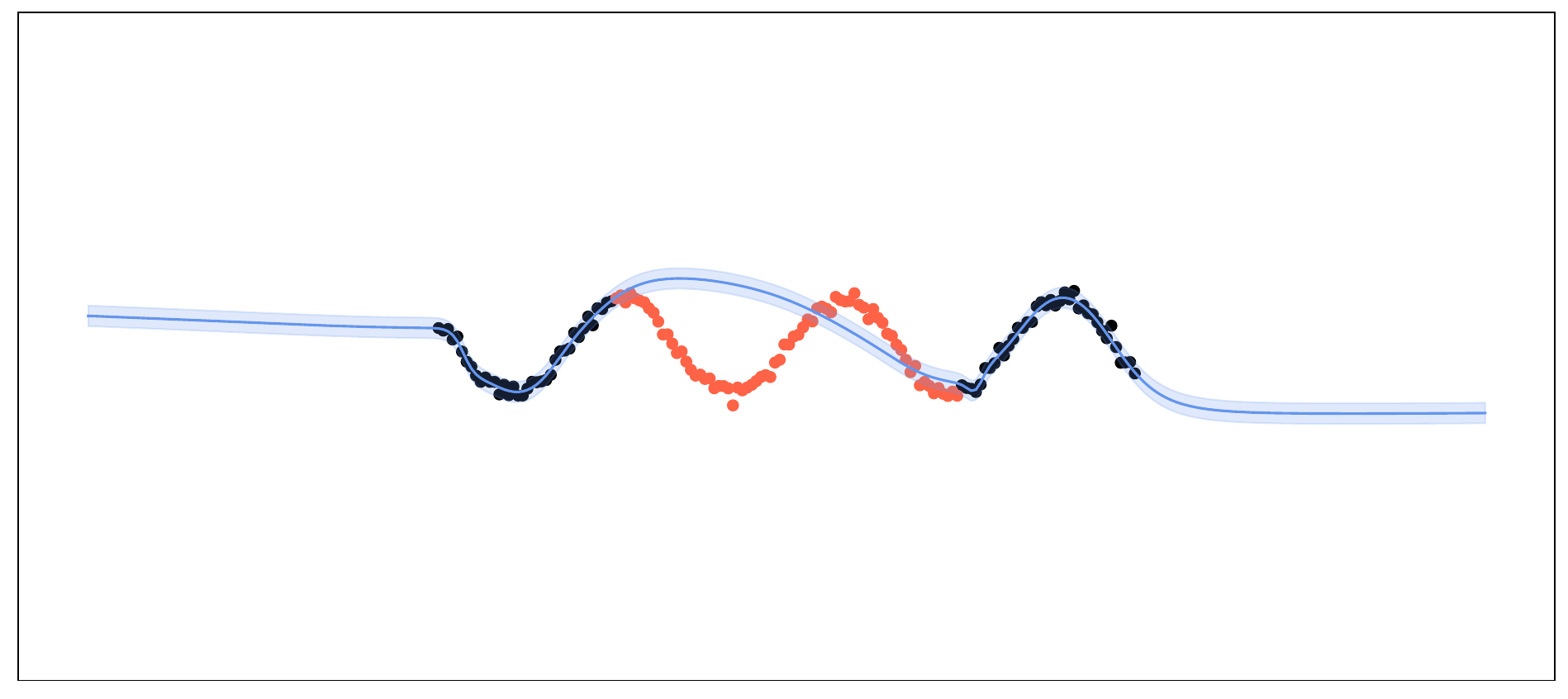} & \includegraphics[width=0.30\textwidth]{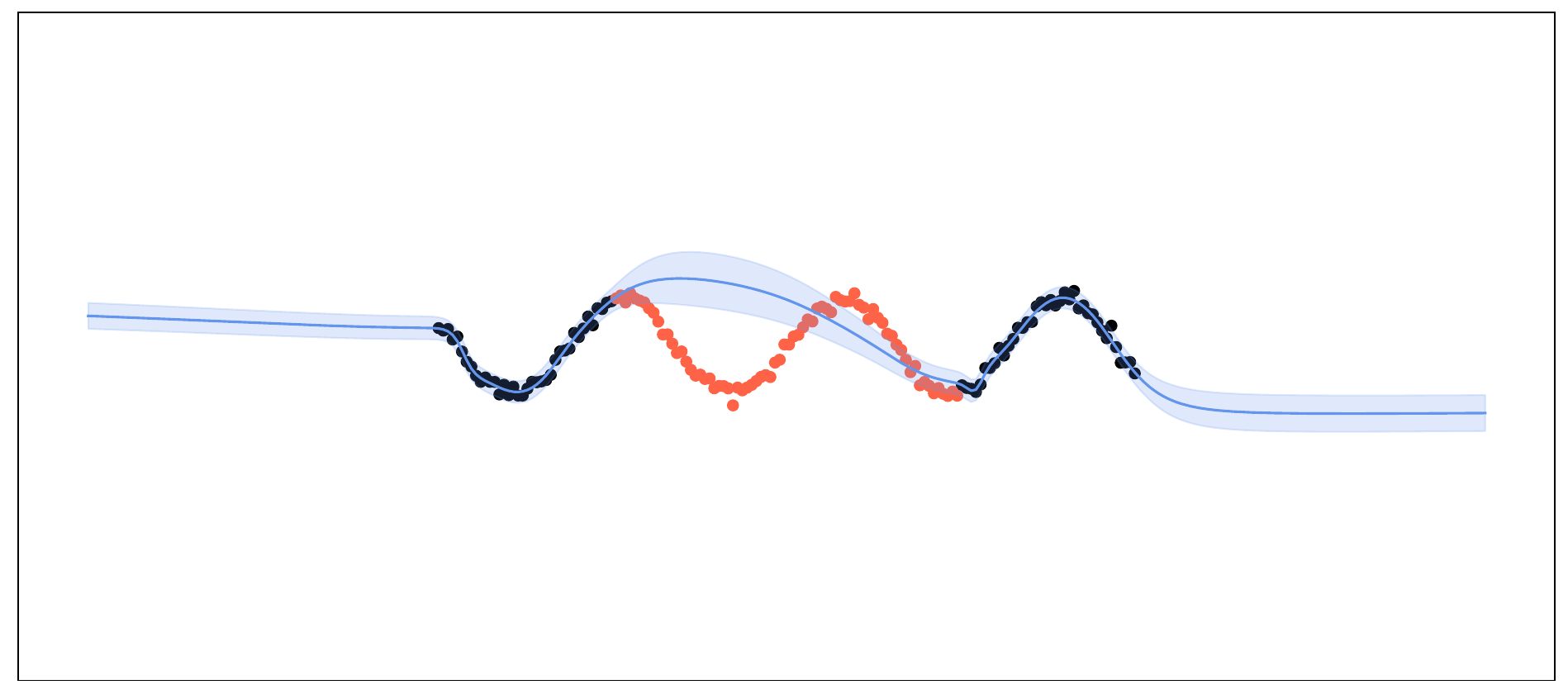}\\
	\end{tabular}
	\end{center}
    \caption{Predictive distribution (mean and two times the standard deviation) on a toy 1D regression dataset with a 2 hidden layer MLP with \(50\) units. Training points are shown in black and the validation set is shown in orange. The first column shows the obtained predictive distribution using early-stopping and a validation set, with \(5\) and \(10\) inducing points, respectively. 
	The second column shows the results obtained without early-stopping. LLA and ELLA are shown in the last column.}
    \label{fig:val}
\end{figure*}

In Section~\ref{sec:alpha} we discussed the fact that the standard maximization of the ELBO does not allow the optimization of the prior variance. To summarize that section, the optimal value for the prior variance is infinite as a result of the mean being fixed to the optimal MAP solution. As discussed, we circumvent this by applying \(\alpha\)-divergences, which are not ill-defined in this learning setup; allowing the optimization of the prior. However, the use of this optimization objective is not perfect and we faced the fact that it tends to over-fit the prior variance to the training data. The middle column of Figure~\ref{fig:val} (middle) shows the obtained predictive distribution (two times standard deviation) learned from VaLLA using the black points as training data. The MAP solution is obtained using a 2 hidden layer MLP with \(50\) hidden units and \emph{tanh} activation, optimized to minimize the RMSE of the training data for \(10000\) iterations with Adam and learning rate \(10^{-3}\). VaLLA on the other hand is trained for \(20000\) iterations. As one may see in the image, the prior variance is fitted to the data to the point where the uncertainty does not increase in the middle gap of the data. 

In this experiment, VaLLA optimizes hyper-parameters along with the variational objective. LLA optimizes the prior variance and likelihood variance by maximizing the marginal log likelihood and ELLA uses LLA's optimal hyper-parameters.

In this situation there are two simple courses of action: we could return to the original ELBO and choose the prior variance by cross-validation; or we could perform early-stopping with the \(\alpha\)-divergences objective, using a validation set to stop training before over-fitting the prior variance parameter. This last approach may not work as it assumes that there is a point during training where the prior variance truly explains the underlying data without over-fitting. However, as the prior variance is set to a relatively large value compared to the optimal one (which is small and leads to over-fitting), this method resulted in great performance for VaLLA, while avoiding using a costly cross-validation approach. The left column of Figure~\ref{fig:val} shows the obtained predictive distribution (two times standard deviation) learned from VaLLA, in this case, using the black points as training data and the orange points as the validation set. In the experiments, we computed the NLL of the validation set every \(100\) training iterations and stopped training when it worsens. This also allowed us to save computational time.

\section{Increasing Inducing Points}\label{app:inducing}

Using the optimal covariance in Proposition~\ref{prop:optimal}, if the set of inducing points equals the training points \(\mathbf{Z} = \mathbf{X}\), the posterior distribution of VaLLA equals that of the exact LLA Gaussian Process. This suggest that increasing the number of inducing points would lead to better uncertainty estimations. In this section, we aim to show how close is the predictive distribution of VaLLA to that of LLA when  we increase the number of inducing points in \(\mathbf{Z}\).

\begin{figure*}[t]
	\begin{center}
	\begin{tabular}{ccc}
    {\scriptsize \(M=5\)} & {\scriptsize \(M=10\)} & {\scriptsize \(M=20\)} \\
	\includegraphics[width=0.30\textwidth]{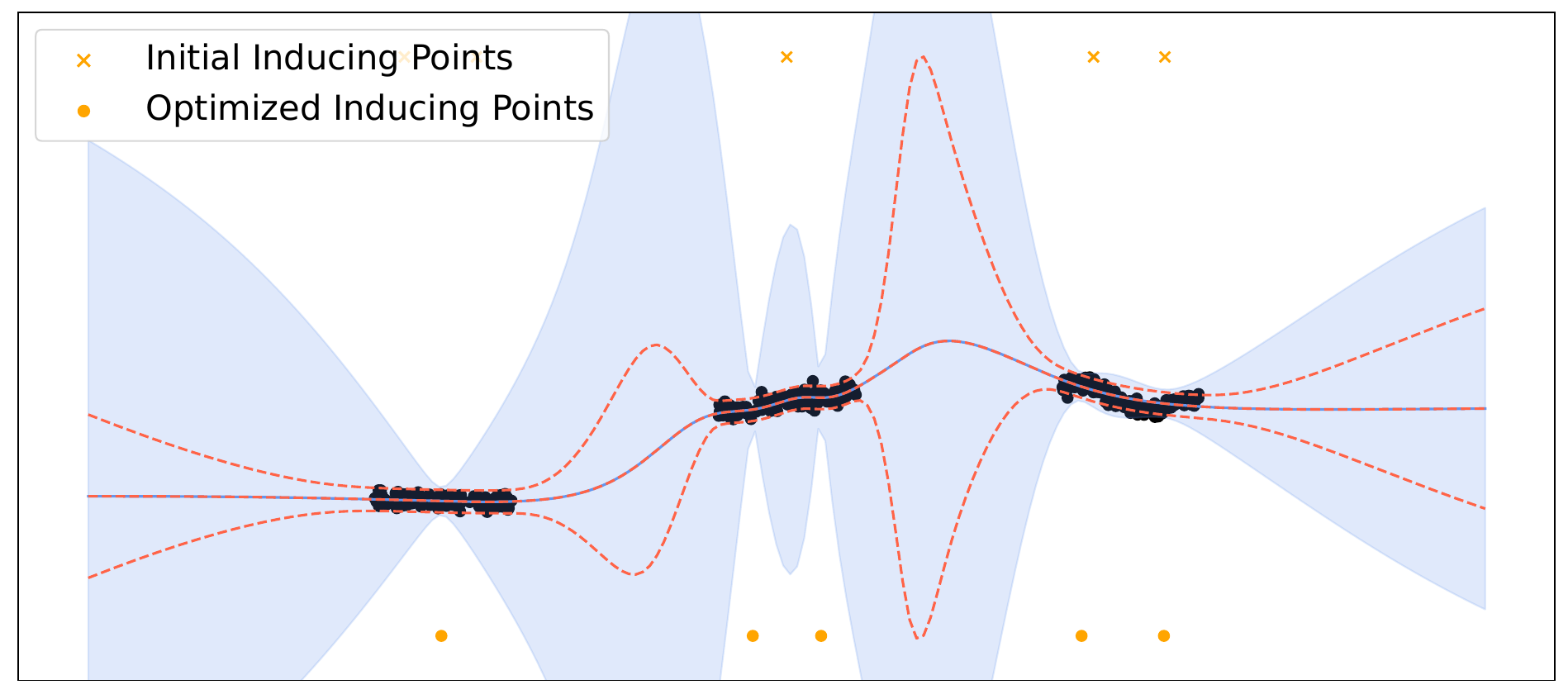} & \includegraphics[width=0.30\textwidth]{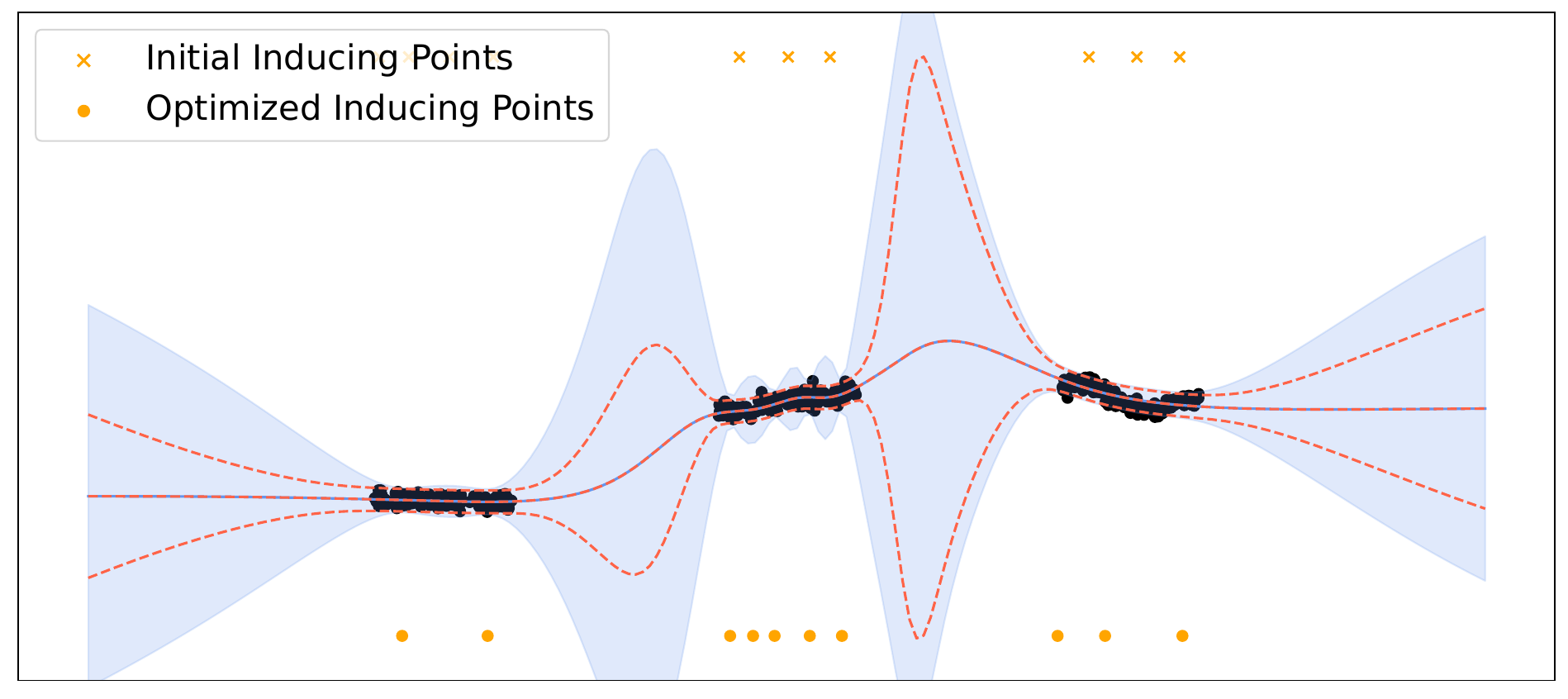} & \includegraphics[width=0.30\textwidth]{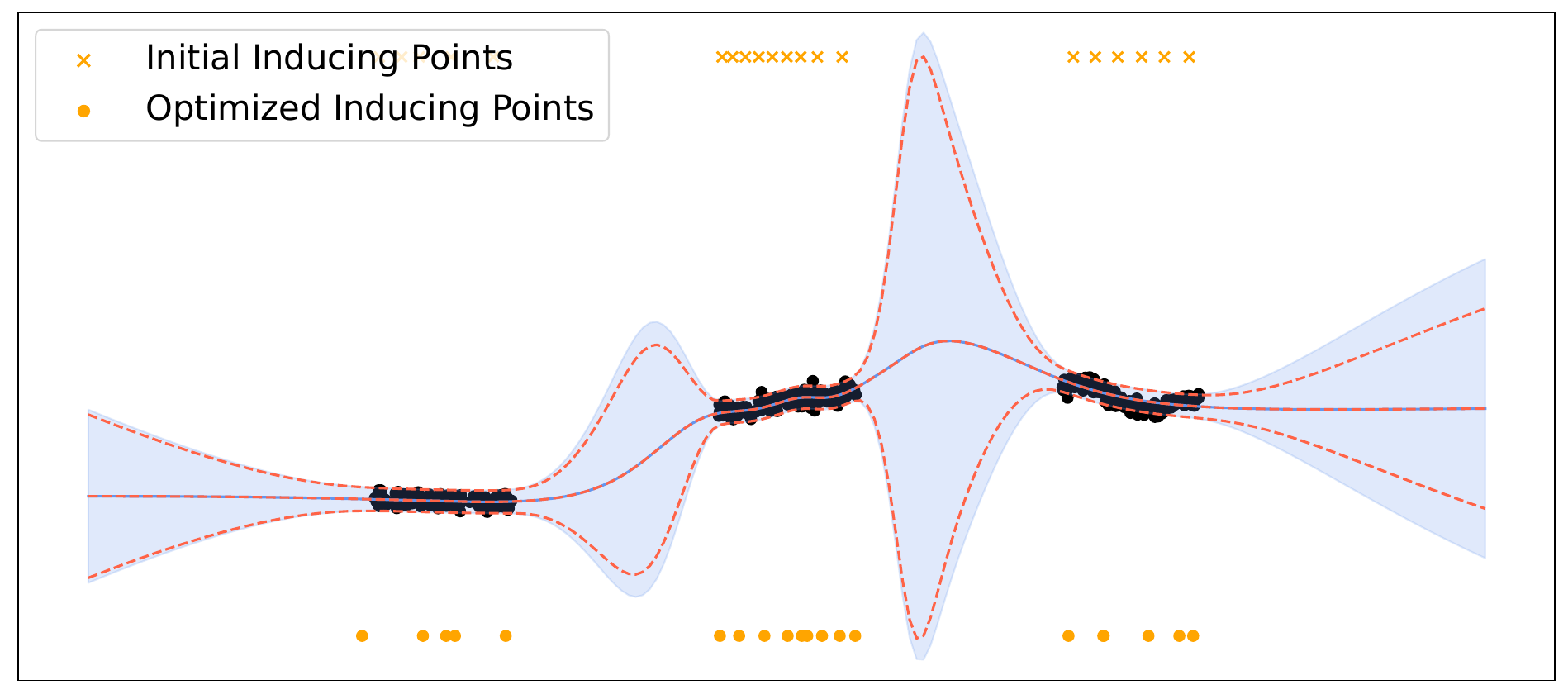}\\
    \includegraphics[width=0.30\textwidth]{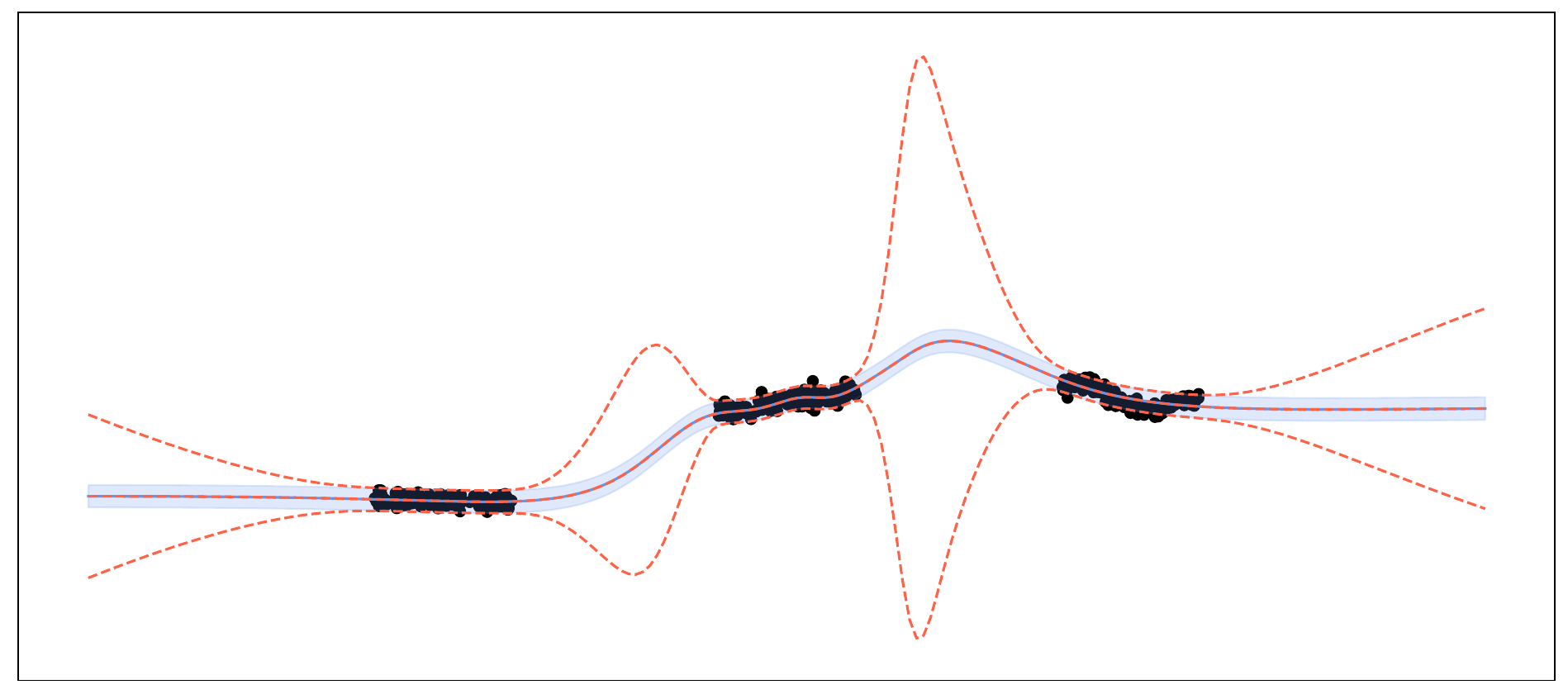} & \includegraphics[width=0.30\textwidth]{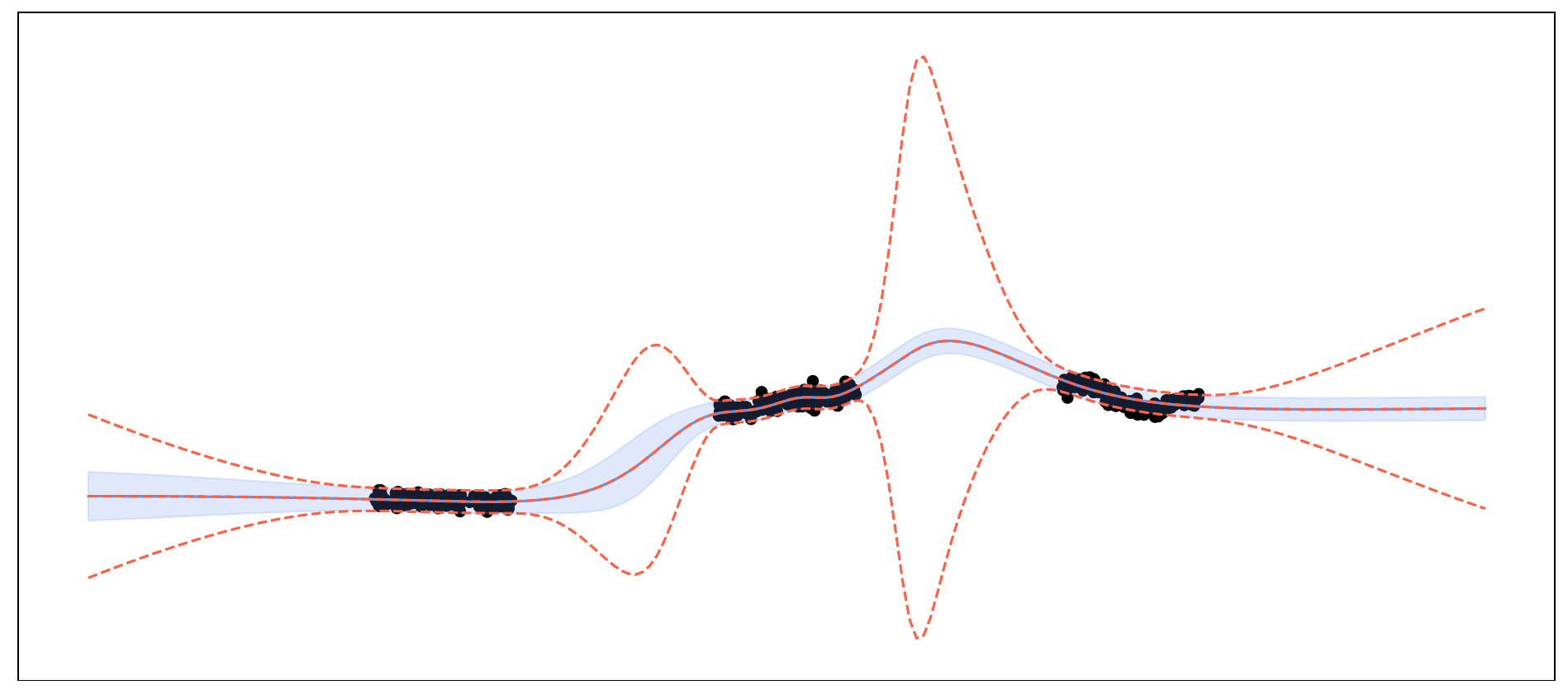} & \includegraphics[width=0.30\textwidth]{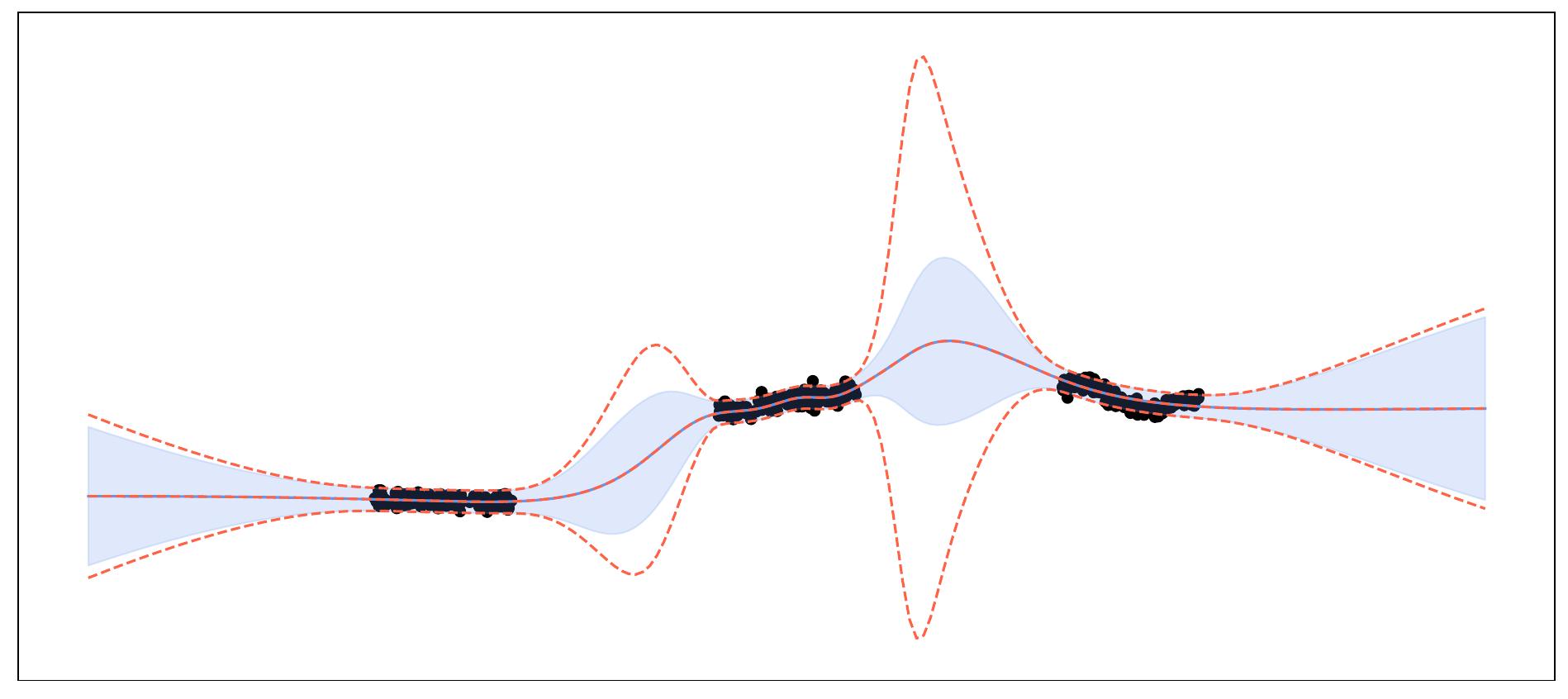}\\
	\end{tabular}
	\end{center}
    \caption{Predictive distribution (two times the standard deviation) on a toy 1D regression dataset with a 2 hidden layer MLP with \(50\) units. The obtained results for \(5\), \(10\) and \(20\) inducing points for VaLLA are shown in the first row. ELLA's predictive distribution with the same amount of samples from the training data is shown in the second row. LLA's predictive distribution is shown in dotted orange.}
    \label{fig:inducing}
\end{figure*}

Figure~\ref{fig:inducing} shows the obtained predictive distribution of VaLLA (first row) and ELLA (second row) for \(M=5\), \(M=10\) and \(M=20\) inducing points/samples. The initial and final locations of the inducing points are also shown for VaLLA. The posterior distribution obtained by LLA is shown in dotted orange. The MAP solution is obtained using a 2 hidden layer MLP with \(50\) hidden units and \emph{tanh} activation, optimized to minimize the RMSE of the training data for \(12000\) iterations with Adam and learning rate \(10^{-3}\). VaLLA on the other hand is trained for \(30000\) iterations.  For this experiment, VaLLA and ELLA use the optimal prior variance and likelihood variance obtained by optimizing LLA's marginal log likelihood. As one may see in the image, it is clear that one of the main differences between the two methods is that VaLLA tends to over-estimate the variance whereas ELLA tends to infra-estimate it, compared to LLA. Furthermore, the value of \(M\) for which the model is closer to the LLA posterior is lower for VaLLA than for ELLA. As we increase $M$, VaLLA's predictive distribution becomes closer and closer to that of LLA.

The initial and final position of the inducing locations is also shown in the figure. For this experiments, the initial values are computed using K-Means. It can be seen how VaLLA is capable of tuning the inducing locations and move them from one cluster of points to another as needed. This is one of the main advantages of this method compared to ELLA.

\section{Experimental Details}\label{app:details}

Source code for the conducted experiments can be accessed in the following repository: \url{https://github.com/Ludvins/Variational-LLA}.

\subsection{MAP solutions}

For regression problems (Year, Airline and Taxi datasets), a 3-layer fully connected NN was used with \(200\) units in each layer. The optimal weights are obtained by minimizing the RMSE using \(20000\) iterations of batch size \(100\) and Adam optimizer \citep{kingma2014adam} with learning rate \(10^{-2}\) and weight decay \(10^{-2}\).

For MNIST and FMNIST experiments, a 2-layer fully connected NN was used with \(200\) units in each layer. The optimal weights are obtained by minimizing the NLL using \(20000\) iterations of batch size \(100\) and Adam optimizer \citep{kingma2014adam} with learning rate \(10^{-3}\) and weight decay \(10^{-3}\).

\subsection{Laplace Library}

The Laplace library \citep{daxberger2021laplace} was used to perform last-layer, KFAC and diagonal approximations of the LLA method and optimize the prior variance on each case. The latter is done by optimizing the log marginal likelihood of the data using the library's \texttt{log\_marginal\_likelihood} method for \(40.000\) iterations with the Adam optimizer and learning rate \(10^{-3}\).

\subsection{Efficient Kernel Computation for MLP}\label{app:ntk}

In this section we discuss an efficient implementation for computing the Neural Tangent Kernel \(\kappa(\mathbf x, \mathbf x')\). First of all, take into account that the computation of the kernel can be reduced to a summation on the number of parameters of the model:
\begin{equation}
    \kappa(\mathbf x, \mathbf x') = \sigma_0^2 J_{\hat{\bm{\theta}}}(\mathbf x)^T J_{\hat{\bm{\theta}}}(\mathbf x') = \sigma_0^2\sum_{\theta_s \in \hat{\bm{\theta}}} \frac{\partial}{\partial \theta_s} g(\mathbf{x}, \hat{\bm{\theta}}) \frac{\partial}{\partial \theta_s} g(\mathbf{x}', \hat{\bm{\theta}}).
\end{equation}
One of the limitations of computing the kernel is storing \(J_{\hat{\bm{\theta}}}(\mathbf x)\) in memory, which is a 3 dimensional tensor of (batch size, number of classes, number of parameters). Computing the kernel as a sum allows to simplify the required computations significantly (we no longer have to store in memory the Jacobians). Consider now a MLP as
\begin{equation}
    g(\mathbf{x}, \hat{\bm{\theta}}) = h_L \circ a \circ H_{L-1} \circ \cdots \circ a \circ h_1(\mathbf{x})\,,
\end{equation}
where each function \(a\) is a non-linear activation function and each function \(h\) is a linear function of the form
\begin{equation}
    h_l(\mathbf x) = \bm{W}_l^T \mathbf x + \bm{b}_l\,.
\end{equation}
With this, \(g\) is supposed to be a fully-connected neural network of \(L\) layers. Each of the partial derivatives of the neural network are
\begin{equation}
    \frac{\partial}{\partial W_{l,j,i}} g(\mathbf{x}, \hat{\bm{\theta}}) \quad \text{and}\quad \frac{\partial}{\partial b_{l,j}} g(\mathbf{x}, \hat{\bm{\theta}}) \quad \forall l=1,\dots,L\,,
\end{equation}
and the kernel is computed simply by adding the product of these derivatives. 
Here, \(i\) is a sub-index denoting input $i$-th to layer $l$.
Similarly, $j$ is a sub-index denoting each component of the bias vector parameter at layer $l$, or similarly, each output of that layer.

In fact, using the structure of the model and the chain rule, the derivative of the \(o^{th}\) output of the network w.r.t. the \(j^{th},i^{th}\) weight parameter of the \(l^{th}\) layer is:
\begin{equation}
    \frac{\partial}{\partial W_{l,j,i}} g_o(\mathbf{x}, \hat{\bm{\theta}}) = \left({\color{teal}\frac{\partial}{\partial h_l} g_o(\mathbf{x}, \hat{\bm{\theta}})}\right) ^T \left({\color{red}\frac{\partial}{\partial W_{l,j,i}} h_l}\right)\,,
	\label{eq:output_derivative}
\end{equation}
where each of the two vectors in the r.h.s. has length equal to the number of units in the layer \(l\). In fact
\begin{equation}
    {\color{red}\frac{\partial}{\partial W_{l,j,i}} h_l} = \bm{1}_l \cdot {\color{purple}a}({\color{blue}h_{l-1}})_i\,,
	\label{mlp:structured_derivative}
\end{equation}
where ${\color{purple}a}({\color{blue}h_{l-1}})_i$ corresponds to the inputs of the \(l^{th}\) layer. Moreover, \(\frac{\partial}{\partial h_l} g_o(\mathbf{x}, \hat{\bm{\theta}})\) can also be computed using the chain rule:
\begin{equation}
	{\color{teal}\frac{\partial}{\partial h_l} g_o(\mathbf{x}, \hat{\bm{\theta}})} = \frac{\partial}{\partial h_{l+1}} g_o(\mathbf{x}, \hat{\bm{\theta}})\frac{\partial}{\partial h_{l}} h_{l+1} = {\color{teal}\frac{\partial}{\partial h_{l+1}} g_o(\mathbf{x}, \hat{\bm{\theta}})} {\color{orange}\bm{W}_l}^T \text{diag}({\color{cyan}a'}({\color{blue}h_{l}}))\,,
\end{equation}
which can be easily computed by back-propagating the derivatives. The same derivations can be easily done for the biases of each layer \(b_{l,j}\). 
As a result, the derivatives only depend on a back-propagating term \({\color{teal}\frac{\partial}{\partial h_l} g_o(\mathbf{x}, \hat{\bm{\theta}})}\) for each layer, the value of the parameters \({\color{orange}\bm{W}_l}, {\color{orange}\bm{b}_l}\) and the propagated outputs at each layer \({\color{blue}h_1},\dots,{\color{blue}h_{L-1}}\) evaluated at the non-linear activation \( {\color{purple}a}(\cdot)\) and its derivative \( {\color{cyan}a'}(\cdot)\). This means that, if we store the intermediate outputs of each layer (\({\color{blue}h_1},\dots,{\color{blue}h_{L-1}}\)) on the forward pass of the model, by using a single backward pass, we can compute \({\color{teal}\frac{\partial}{\partial h_l} g_o(\mathbf{x}, \hat{\bm{\theta}})}\) for each layer. 

Critically, given each ${\color{teal}\frac{\partial}{\partial h_l} g_o(\mathbf{x}, \hat{\bm{\theta}})}$, we can add the contribution of each layer to the kernel, using (\ref{eq:output_derivative}). In this process, we can sped-up the computations by using structure in the derivatives. For example, in (\ref{mlp:structured_derivative}) we observe that the derivative has a simple form which is a vector of ones times a scalar. Furthermore,  there is no dependence on $j$, the output unit corresponding to the weight $W_{l,j,i}$. Therefore, for two instances $\mathbf{x}$ and $\mathbf{x}'$, the kernel contribution (ignoring the prior variance parameter) corresponding to outputs $o$ and $o'$ is:
\begin{align}
\frac{\partial}{\partial W_{l,j,i}} g_o(\mathbf{x}, \hat{\bm{\theta}}) \frac{\partial}{\partial W_{l,j,i}} g_o'(\mathbf{x}', \hat{\bm{\theta}}) 
	&= 
\left({\color{teal}\frac{\partial}{\partial h_l} g_o(\mathbf{x}, \hat{\bm{\theta}})}\right) ^T \left({\color{red}\frac{\partial}{\partial W_{l,j,i}} h_l}\right)
\left({\color{teal}\frac{\partial}{\partial h_l} g_o(\mathbf{x}', \hat{\bm{\theta}})}\right) ^T \left({\color{red}\frac{\partial}{\partial W_{l,j,i}} h_l}\right)	\\
	&= 
\left({\color{teal}\frac{\partial}{\partial h_l} g_o(\mathbf{x}, \hat{\bm{\theta}})}\right) ^T \left({\color{red}\frac{\partial}{\partial W_{l,j,i}} h_l}\right)
\left({\color{red}\frac{\partial}{\partial W_{l,j,i}} h_l}\right)^T \left({\color{teal}\frac{\partial}{\partial h_l} g_o'(\mathbf{x}', \hat{\bm{\theta}})}\right) \\
	&= 
\left({\color{teal}\frac{\partial}{\partial h_l} g_o(\mathbf{x}, \hat{\bm{\theta}})}\right) ^T 
	\bm{1}_l \cdot {\color{purple}a}({\color{blue}h_{l-1}(\mathbf{x})})_i
	\cdot {\color{purple}a}({\color{blue}h_{l-1}(\mathbf{x}')})_i \bm{1}_l^T
	\left({\color{teal}\frac{\partial}{\partial h_l} g_o'(\mathbf{x}', \hat{\bm{\theta}})}\right) \\
	&= 
	s_{o,\mathbf{x}}^l {\color{purple}a}({\color{blue}h_{l-1}(\mathbf{x})})_i
	\cdot {\color{purple}a}({\color{blue}h_{l-1}(\mathbf{x}')})_i s_{o',\mathbf{x}'}^l \\
	&= 
	s_{o,\mathbf{x}}^l \mathbf{s}_{o',\mathbf{x}'}^l
	{\color{purple}a}({\color{blue}h_{l-1}(\mathbf{x})})_i
	{\color{purple}a}({\color{blue}h_{l-1}(\mathbf{x}')})_i 
	\,,
\end{align}
with $s_{o,\mathbf{x}}^l = \left({\color{teal}\frac{\partial}{\partial h_l} g_o(\mathbf{x}, \hat{\bm{\theta}})}\right) ^T \bm{1}_l$ a scalar. Similar simplifications occur in the case of, \emph{e.g.}, a convolutional layer.

Summing up, by using this method, all the required kernel matrices can be easily and efficiently computed, for a mini-batch of data points and a set of inducing points, 
with a similar cost as that of letting the mini-batch or the inducing points go through the DNN.
A disadvantage is, however, that the described computations will have to be manually coded for each different DNN architecture. This becomes tedious in the case of very big DNN with complicated layers, as
described in Section \ref{sec:limitations}.

\section{Further Analysis of the Quantile Metric}\label{app:Q}

\begin{figure*}
	\begin{center}
	\begin{tabular}{ccc}
    {\scriptsize Year} & {\scriptsize Airline} & {\scriptsize Taxi} \\
	\includegraphics[width=0.30\textwidth]{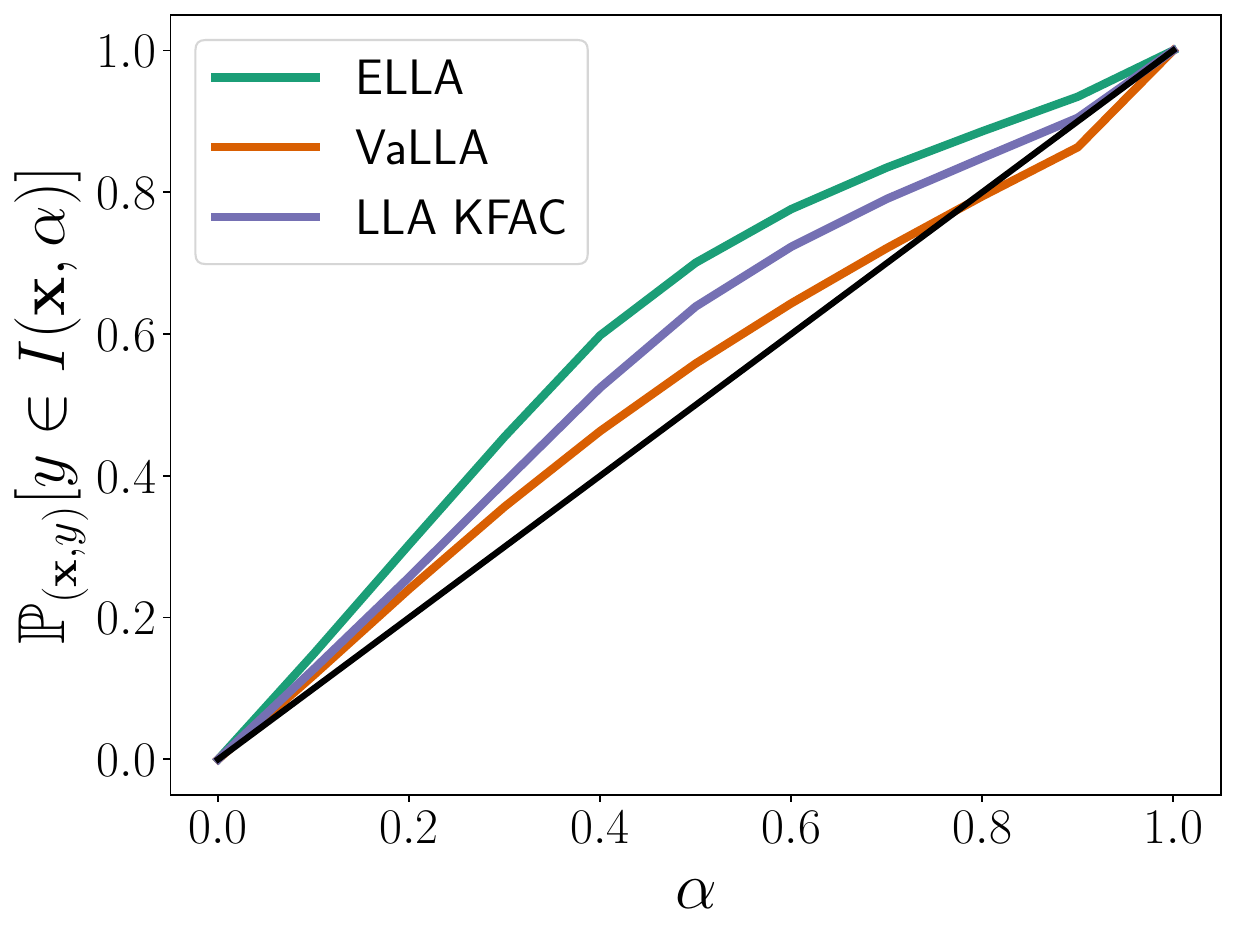} & \includegraphics[width=0.30\textwidth]{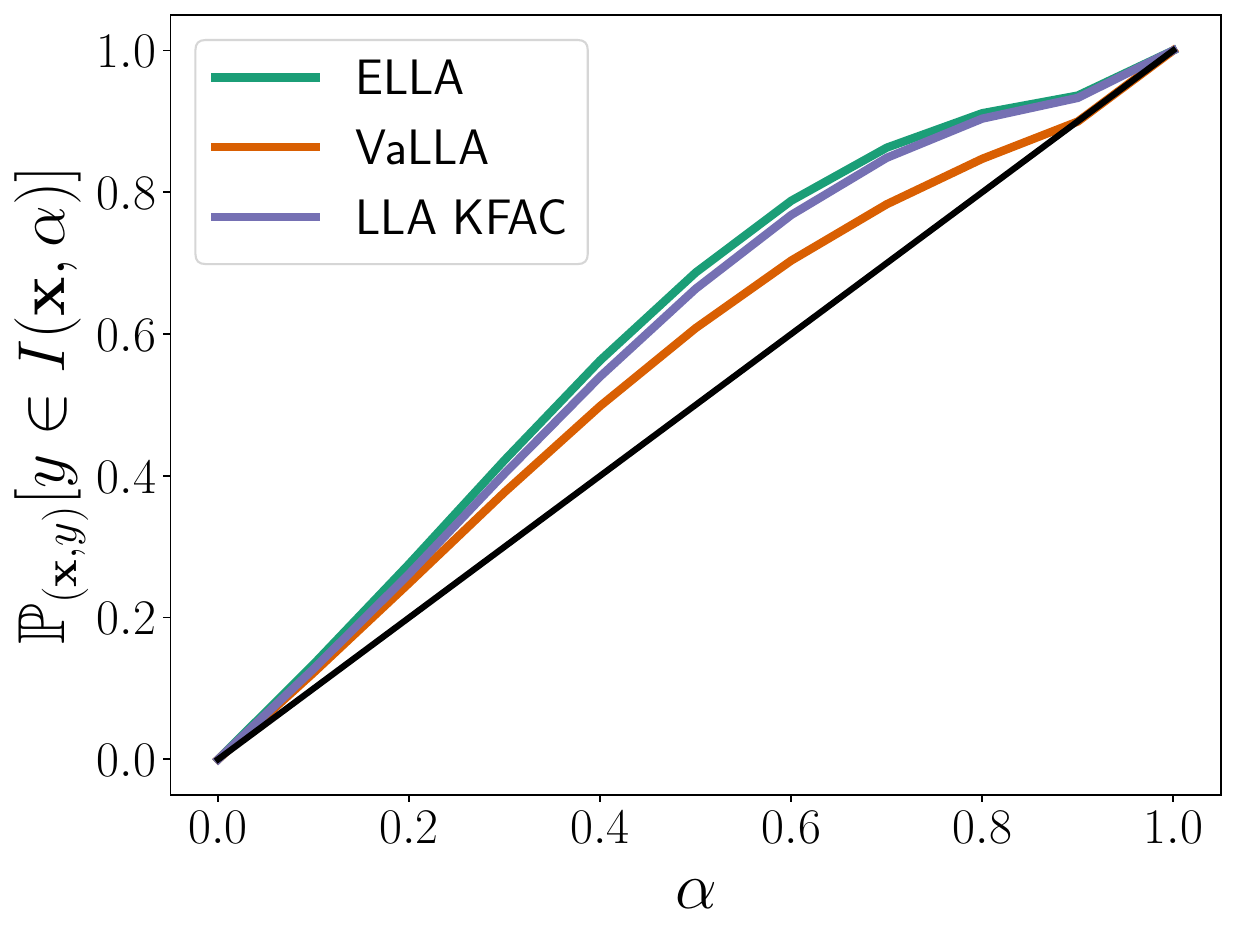} & \includegraphics[width=0.30\textwidth]{imgs/Q_Taxi_plot.pdf}\\
	\end{tabular}
	\end{center}
    \caption{Illustration of CQM, for each method, on the regression datasets Year (left), Airline (middle) and Taxi (right).  
	The MAP solution is given by a fully connected network with 3 hidden layers of \(200\) units and \emph{tanh} activations.}
    \label{fig:app:QCM}
\end{figure*}

As stated in Section~\ref{sec:exp}, we proposed a new metric for regression problems that, in a way, extends ECE to regression problems with Gaussian predictive distributions \emph{with the same mean}. This kind of metric is desirable for LLA methods as all of them rely on keeping the optimal MAP solution as the predictive mean of the model. They only differ in the predictive variance. Formally, CQM computes 
for each \(\alpha \in (0, 1)\) the probability that points fall into the predictive centered interval of probability \(\alpha\). The underlying reasoning is that, if the model explains the data well enough, \(\alpha \cdot 100\%\) of the points will fall inside the  \(\alpha \cdot 100\%\) centered quantile interval. Thus, the metric defined as 
\begin{equation}\label{eq:app:CQM}
	\text{CQM} = \int_0^1 \ \Big|\mathbb{P}_{(\mathbf{x}^\star, y^\star)}\left[ y^\star \in I(\mathbf{x}^\star, \alpha) \right] - \alpha\Big| \ d\alpha\,,
\end{equation}
should be roughly \(0\) when the model predictive distribution is similar to the actual one, given by the observed data. 

Figure~\ref{fig:app:QCM} shows the evolution of \(\mathbb{P}_{(\mathbf{x}^\star, y^\star)}\left[ y^\star \in I(\mathbf{x}^\star, \alpha) \right]\) w.r.t. \(\alpha\) for the best performing models in the regression problems. CQM corresponds to the area between the shown curve and \(y = x\) (black line). This figure allows to argue that (in general) all methods are over-estimating the predictive variance as they are giving values above the diagonal. That is, for a specific value of \(\alpha \in (0, 1)\), the reported probabilities are higher than \(\alpha\), meaning that, on average, there are more points in \(I(\mathbf{x}, \alpha)\) than they should. That is, the predicted interval is larger than it should, which can only mean that the variance is over-estimated. From a geometrical perspective, it is clear that CQM is always greater than \(0\) and lower than \(0.5\); independently of the model and dataset used.

In fact, this figure allow to visually study the level of over/infra-estimation of the prediction uncertainty, for each degree of confidence \(\alpha\). For example, in the Year 
dataset (Figure~\ref{fig:app:QCM}) we see that VaLLA slightly over-estimates the uncertainty for \(\alpha \in (0, 0.7)\) while it infra-estimates it for larger values of \(\alpha\).

\end{document}